\documentclass{article}
\usepackage{PRIMEarxiv}

\usepackage[T1]{fontenc}
\usepackage[utf8]{inputenc}
\usepackage{bm}
\usepackage{amsmath}
\usepackage{amsfonts}
\usepackage{amssymb}

\usepackage{graphicx}
\usepackage{graphbox} 
\usepackage{caption}
\DeclareCaptionLabelSeparator{sixteenpt}{\kern16pt}
\usepackage{subcaption} 
\usepackage{multirow}
\usepackage{listings}
\usepackage[most]{tcolorbox}
\tcbset{
    base/.style={
        boxrule=2pt,
        arc=0mm,
        breakable,
        colback=white!97!black,
        colframe=black!80,
        fonttitle=\small\bfseries,
        left=2mm,
        right=2mm,
        top=2mm,
        bottom=2mm
    }
}
\usepackage{hhline}
\usepackage{lscape}

\usepackage{algorithm}
\usepackage{algorithmic}
\usepackage{float}

\usepackage{booktabs}
\usepackage{array}
\usepackage{threeparttable}

\usepackage{bbm}
\usepackage{anyfontsize}
\usepackage{pifont}

\usepackage{appendix}

\usepackage{natbib}
 \bibpunct[, ]{(}{)}{,}{a}{}{,}%
\usepackage{url}
\usepackage[dvipsnames,svgnames,table]{xcolor}
\usepackage[letterpaper=true,colorlinks=true,pdfpagemode=none,citecolor=OliveGreen,linkcolor=BrickRed,urlcolor=BrickRed,pdfstartview=FitH]{hyperref}

\newcommand{\sym}[1]{\textsuperscript{\kern-0.1em\scriptsize #1}}

\newcommand{\fignote}[1]{%
  \par\vspace{3pt}%
  {\footnotesize\itshape\raggedright\setlength{\parindent}{0pt}\textbf{Note.}\enspace #1\par}%
}
\let\tabnote\fignote

\begin{document}

\pagestyle{fancy}
\fancyhf{}
\fancyhead[L]{\textit{Wisdom of LLM Crowds}}
\fancyhead[R]{\textit{Fang et al.}}
\cfoot{\thepage}
\thispagestyle{empty}

\title{Harnessing the Wisdom of LLM Crowds through Complementarity-Driven Iterative Collaboration}

\author{
  Yanbin Fang \\
  Business School \\
  Chinese University of Hong Kong, Shanghai, China \\
  \texttt{yibofang@link.cuhk.edu.hk}
  \And
  Xuan Wei \\
  Antai College of Economics and Management \\
  Shanghai Jiao Tong University, Shanghai, China \\
  \texttt{weix@sjtu.edu.cn}
  \And
  Wei Chen \\
  School of Business \\
  University of Connecticut, Stamford, CT, US \\
  \texttt{weichen@uconn.edu}
}

\maketitle

\begin{abstract}
Large language models (LLMs) are increasingly deployed in enterprise contexts to support complex problem-solving tasks. Yet individual LLMs remain bounded by model-specific capability limitations. These heterogeneous capability boundaries pose a deployment challenge, but they also create an opportunity: strategically coordinating multiple LLMs may unlock collective intelligence that exceeds the performance of any single model.
Existing approaches fix how models are combined in advance, overlooking the dynamic and state-dependent role of complementarity in complex LLM problem solving.
Drawing on the wisdom-of-crowds paradigm, we reconceptualize collective LLM intelligence as relay-style complementarity: a sequential coordination process in which each successor model is selected to address the specific bottleneck identified in its predecessor's output.
To operationalize this relay-style complementarity, we propose WILC (Wisdom Integration of LLM Crowds), a framework grounded in two design principles. First, iterative reflection-and-refinement establishes a state-preserving workflow through which models diagnose and refine prior outputs. Second, complementarity-driven model selection governs model transitions through a dual-gate mechanism: prospective complementarity fit (PCF) identifies the worker most suited to the current bottleneck, while posterior complementarity gain (PCG) evaluates whether the selected transition improves the evolving solution. Together, these design principles support strategic model transitions that refine answers through capability complementarity.
Extensive experiments across four diverse benchmarks show that WILC achieves superior performance compared to existing approaches, including single-model self-refinement, ensemble methods, and dedicated query-routing methods.
Under standardized pricing assumptions, WILC achieves comparable average benchmark performance to GPT-5.2 at approximately 7$\times$ lower estimated per-query cost, while facilitating data sovereignty through self-hosted deployment.
This study contributes to IS research by extending wisdom-of-crowds theory from static aggregation to sequential AI complementarity and by providing transferable design principles for multi-AI coordination.
\end{abstract}


\section{Introduction}\label{sec:Intro}
Large language models (LLMs) have demonstrated remarkable performance across a wide range of tasks and have been increasingly adopted in enterprise contexts. For instance, firms leverage LLMs internally to support employees in resolving operational or analytical challenges, and deploy them externally to deliver LLM-driven services to users through conversational or interactive applications~\citep{haki2025integrating, raza2025industrial}.
Such deployments, however, require organizations to choose among many available models that differ markedly in capability. This heterogeneity stems from variations in training data, model architectures, and training techniques~\citep{shnitzer2023large, lu2024routing}. Even as frontier models continue to advance, their relative strengths remain uneven across task types, and the model best suited to a given problem is rarely known in advance. This issue is particularly salient among open-source models---a preferred choice for private enterprise deployment given strategic considerations such as data privacy, cost efficiency, and customization flexibility~\citep{aimultiple_enterprise_genai_2024}. Yet this very heterogeneity also presents a promising opportunity to enhance LLM deployments without escalating model parameter scales. Specifically, integrating the complementary advantages of multiple models may unlock collective wisdom and improve problem-solving performance beyond what individual models achieve alone.

The aim of integrating multiple LLMs (hereafter, \textit{the wisdom of LLM crowds}) aligns with the established ``wisdom of crowds'' paradigm, which focuses on aggregating the judgments of large groups to achieve superior collective performance compared to any single member and even expert individuals~\citep{surowiecki2005wisdom, wei2022combining}. However, the nature of the problem differs fundamentally when the crowd members shift from humans to LLMs.
\textit{First}, regarding the outputs, LLMs produce unstructured outputs, such as free-form text, code, and reasoning chains, rather than the commensurable judgments commonly studied in traditional crowd settings. This limits the direct application of conventional statistical aggregation~\citep{ atanasov2017distilling,wei2025human}, although voting remains possible when outputs can be reduced to standardized final answers.
\textit{Second}, the inference paradigm differs. While traditional wisdom of crowds relies on aggregating independent and single-shot judgments~\citep{becker2022crowd, thomas2021model}, LLMs typically operate via autoregressive decoding, transforming problem solving into a dynamic process of inference-time computation~\citep{wei2022chain}. This sequential generation makes it natural to allocate additional inference-time resources to revise an initial draft; outputs can often be improved through iterative refinement mechanisms, including self-correction and reflective critique~\citep{madaan2023self, shinn2023reflexion}.
Consequently, LLM crowds create an opportunity to harness collective intelligence not only by aggregating completed outputs, but also by deliberately coordinating models during the evolving problem-solving process.

Current approaches for the wisdom of LLM crowds can be broadly divided into two categories: (a) LLM ensemble and (b) multi-agent systems (MASs).
LLM ensembles typically operate by either routing queries to the most suitable model based on models' capabilities~\citep{shnitzer2023large, lu2024routing} or fusing outputs from multiple models~\citep{wang2022self, jiang2023llm}.
Conversely, MAS frameworks orchestrate collaboration through the role-playing of agents and the use of either predefined or automatically generated team workflows~\citep{li2024survey, guo2024large}.
Despite leveraging multiple, often heterogeneous, models, existing designs in both categories still fall short of achieving deep and synergistic capability complementarity.
In ensembles, routing and fusion usually determine the collaboration point either \textit{before} generation through model selection or \textit{after} generation through output fusion, giving them limited access to evolving intermediate states. MASs support richer interactions among models, but their roles and workflows are typically predefined or reused across queries within the same task family, rather than tailored to the individual query. Neither line of work adjusts its coordination as an individual solution actually develops,
even though reflective mechanisms~\citep{shinn2023reflexion, madaan2023self} can already
expose where a solution falls short.

\begin{figure}[!htb]
  \centering
  \caption{Motivating Example: Static Aggregation vs. Relay-Style Complementarity.}
  \includegraphics[width=\textwidth]{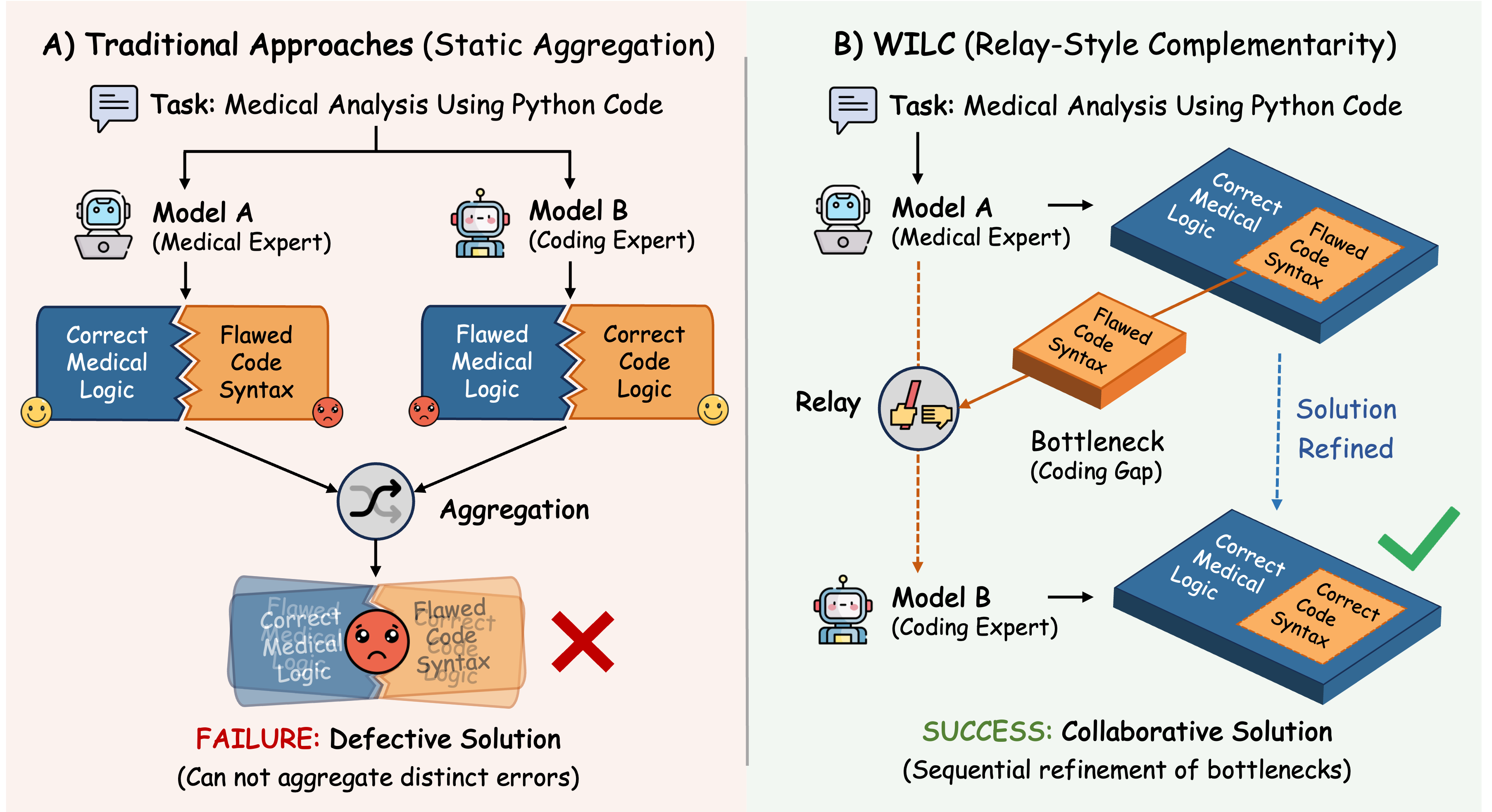}
  \fignote{The task requires both medical knowledge and coding skills. \textbf{Left:} Static aggregation combines completed responses after generation, which can retain component-specific errors when no individual response resolves all task requirements. \textbf{Right:} WILC achieves relay-style complementarity by first assigning the task to model A and then using the bottleneck identified in its intermediate solution (the syntax error) to guide the transition to model B, which corrects the solution.}
\label{fig:motivation-example}
\end{figure}

To transcend these capability boundaries and unlock the latent potential of heterogeneous LLM crowds, coordination must move \textit{within} the generation process, which requires clarifying how complementarity is realized once problem solving unfolds as a sequence rather than a set of parallel contributions. Complementarity is generally defined as ``the quality of being different but useful''~\citep{cambridge2025complementarity}, yet its realization mechanism differs between static judgment tasks and dynamic sequential reasoning tasks. Traditional wisdom-of-crowds settings commonly combine parallel judgments through static aggregation~\citep{surowiecki2005wisdom,yin2021learning,wei2025human}. In contrast, complex LLM problem solving is path-dependent: an impasse reached by a model reflects a specific capability bottleneck in the evolving solution. We conceptualize the use of such intermediate bottlenecks to coordinate models during generation as \textit{relay-style complementarity}. Rooted in capability asymmetries across heterogeneous models~\citep{hemmer2025complementarity}, relay-style complementarity arises when a successor model can specifically address the predecessor's bottleneck and deliver non-negative marginal improvement (i.e., complementary team performance)~\citep{bansal2021does} to the evolving solution. This distinction between static aggregation and relay-style complementarity is illustrated through the example in Figure~\ref{fig:motivation-example}, which contrasts how each mechanism addresses a hybrid task: analyzing a patient's clinical notes to calculate a personalized drug dosage using Python code.

To operationalize this relay-style complementarity, we follow the design science paradigm~\citep{hevner2004design,abbasi2024pathways,fang2025computational} and propose WILC (Wisdom Integration of LLM Crowds), a novel collaborative framework that conceptualizes problem solving as a dynamic, relay-style iterative reflection-and-refinement process. Drawing on the insight that capability heterogeneity allows one model to compensate for the capability limitations of another~\citep{kamoi2024can,he2025makes}, WILC introduces a strategic model switching mechanism. When an incumbent model falters, the system identifies the specific nature of the issues (hereafter, \textit{bottlenecks}) and transfers the evolving problem-solving context to a successor model. The core rationale is to maximize \textit{model-bottleneck fitness}, selecting a successor not merely because it is better on average, but because its specific capability spectrum aligns with the current bottlenecks. In this way, each model transition is conditioned on the current solution state rather than determined solely by the initial query or a predefined workflow.

To promote the realization of such complementarity at each relay step, WILC employs a complementarity verification mechanism to govern the iteration. It operates along two dimensions: (a) \textit{prospective complementarity fit} (PCF), which identifies the model with the highest potential to address the current bottleneck by adapting a contextual multi-armed bandit algorithm~\citep{li2010contextual}, and (b) \textit{posterior complementarity gain} (PCG), which evaluates whether the selected model preserves or improves the evolving solution. By combining prospective selection with posterior verification, WILC constrains unproductive model transitions and sustains collaborative refinement until the crowd's complementary potential is exhausted or a satisfactory answer is attained.


To validate the effectiveness of WILC, we conduct extensive experiments with diverse LLM crowds across four benchmarks: code generation, mathematical reasoning, general knowledge reasoning, and data visualization. Our results demonstrate that WILC effectively coordinates the complementary capabilities of the LLM crowd and outperforms existing methods. Importantly, WILC mitigates the performance degradation risks that organizations face when making suboptimal model selection decisions amid multiple available models.
Ablation studies validate the contributions of key components within the WILC framework, notably, WILC outperforms single-model self-reflection across every benchmark-crowd combination, indicating that its gains cannot be reduced to iterative refinement alone. 
Furthermore, we analyze the LLM invocation overhead of WILC to evaluate its cost-effectiveness, demonstrating that, under the standardized pricing assumptions used in our analysis, WILC achieves performance comparable to GPT-5.2 at roughly 7$\times$ lower estimated per-query cost by coordinating moderate-scale open-source models, while simultaneously facilitating data sovereignty through self-hosted deployment.

The rest of this paper is organized as follows: The following section reviews the relevant literature. Then, we detail the proposed WILC framework in Section \ref{sec:WILC} and evaluate the proposed framework by describing the experimental design and showing the empirical results in Section \ref{sec:Experiments}. Finally, we conclude the paper by presenting the contributions, implications, limitations, and future research directions in Section \ref{sec:Conclusion}.

\section{Related Work}\label{sec:RelatedWork}
We review three streams of work: capability complementarity (Section \ref{sec:Complementarity}); approaches that leverage the wisdom of multiple LLMs through ensembles and multi-agent systems (Sections \ref{sec:LLMEnsemble} and \ref{sec:MAS}); and the contextual multi-armed bandit problem (Section \ref{sec:MAB}), which provides the technical foundation of our design.

\subsection{Complementarity in Collaboration}\label{sec:Complementarity}
Complementarity, generally defined as ``quality of being different but useful when combined''~\citep{cambridge2025complementarity}, serves as a cornerstone for understanding how diverse entities can outperform any single one acting alone~\citep{hemmer2025complementarity}. 
Existing literature has explored complementarity through the lens of human-algorithm collaboration, systematically investigating the conceptualization and methodologies to leverage the complementarity~\citep{bansal2021does, fugener2022cognitive, donahue2022human}. 

For the conceptualization, existing literature consistently posits that complementarity is realized when the combined system yields higher performance (or equivalently, lower loss) than either the human or the algorithm acting in isolation~\citep{donahue2022human}. It is theoretically impossible if the performances of the human and the algorithm are constant across different contexts, or if one entity consistently dominates the other~\citep{bansal2021does}. This implies that for complementarity to exist, the capabilities of the collaborators must be heterogeneous across different contexts, such that neither entity is universally superior.
Drawing on the conceptualization, researchers have designed approaches to realize complementary team performance in tasks like image classification and crowdsourcing. 
Specifically, \citet{steyvers2022bayesian} proposed a Bayesian modeling approach to combine the classifications and confidence scores from humans and machine classifiers in the image classification task. Extending this Bayesian perspective, \citet{wei2025human} proposed a human-algorithm collaborative framework for crowdsourcing tasks, which leverages a hybrid complementarity score to dynamically adjust the integration of human labels and algorithmic predictions.

In this study, we draw on the insights from existing studies to design a collaborative framework for LLM crowds from the perspective of complementarity. In the context of LLM crowds, inherent variations in training data distributions, model architectures, and training techniques engender capability heterogeneity across different LLMs, thereby underpinning the existence of complementarities. Building upon prior research, we adapt the conceptualization of complementarity to the proposed relay-style collaboration, thus constructing a novel artifact for achieving complementarity in LLM crowds.

\subsection{Large Language Model Ensemble}\label{sec:LLMEnsemble}
Large language model (LLM) ensemble, which stems from ensemble learning~\citep{dietterich2000ensemble,sagi2018ensemble}, has emerged as a promising research direction that aims to harness the collective wisdom of LLMs to achieve superior performance. 
Depending on whether the models used for ensemble are identical, current LLM ensemble methods can be categorized into self-ensemble and heterogeneous ensemble.

Self-ensemble involves generating multiple responses to a single query using the same LLM and aggregating them to derive a final solution. A representative work is Chain-of-Thought with Self-Consistency (CoT-SC)~\citep{wang2022self}, which utilizes diverse chain-of-thought prompts to elicit multiple reasoning paths from the model, ultimately selecting the final output via majority voting. Building on CoT-SC, subsequent studies have introduced various enhancements, such as incorporating more complex reasoning chains~\citep{fu2023complexity}, implementing answer verification mechanisms to screen out implausible results~\citep{li2023making}, and adopting self-agreement strategies to improve answer reliability~\citep{lin2024just}. 

Heterogeneous ensemble methods leverage the heterogeneous strengths of different models primarily through two approaches: (1) routing queries to the LLM that is best suited for a particular task based on its capabilities; and (2) fusing outputs from multiple models to synthesize a superior response. It is predicated on the assumption that different LLMs possess heterogeneous capabilities, and therefore it is possible to leverage their complementary strengths and mitigate individual weaknesses~\citep{shnitzer2023large,jiang2023llm,lu2024routing}. For model routing methods, an external model router is typically trained to assign each query to the most suitable LLM~\citep{shnitzer2023large,lu2024routing}. For instance, \citet{shnitzer2023large} learn a routing function for each model based on their performance on benchmark datasets by supervised learning. Given a new query, the best model is selected by comparing the output of each routing function, which represents the probability of the model providing the correct answer. 
For answer fusion methods, studies typically learn a fusion model to synthesize different models' responses. For instance, the LLM-Blender framework first evaluates the responses of $N$ LLMs and then employs a fusion model to merge the outputs of the top $K$. This fusion model is fine-tuned from a pre-trained LLM that takes the input query and $K$ responses and produces an improved output as the final response~\citep{jiang2023llm}.

In summary, self-ensemble improves the reliability of a single LLM but cannot draw on the strengths of other models. Heterogeneous ensembles do combine multiple models, yet they fix the collaboration point either before or after generation: routing remains bounded by the capability of the model it selects, while fusion depends on the fusion model and can be misled when weaker members produce correlated errors. In both cases, the models never interact while the solution is still taking shape.

\subsection{Multi-Agent Systems (MASs)}\label{sec:MAS}
Inspired by human teamwork, multi-agent systems (MASs) have been developed to foster the collaboration between LLM-based agents with diverse expertise~\citep{li2024survey}. Compared to a single agent, MASs provide enhanced capabilities by two core design elements: (a) role-playing, i.e., assigning diverse roles to agents, and (b) team workflow, i.e., facilitating interactions among agents to harness the power of teamwork~\citep{guo2024large}. Based on the designer's identity, existing works can be divided into two categories: (a) manually designed MAS by human experts and (b) automatically designed MAS by LLMs.

For manually designed MAS, human experts predefine clear role divisions and collaboration workflows based on specific task properties and domain knowledge~\citep{xiong2023examining, hong2024metagpt, qian2024chatdev}. Such systems are typically tailored for specific tasks (like software development~\citep{xiong2023examining}, scientific experiments~\citep{zheng2023chatgpt}, scientific debates~\citep{tang2024medagents, du2023improving}, social simulations~\citep{park2022social}, etc.) and remain static once established. However, manually designed MASs have clear limitations: (1) they require significant expert effort and domain knowledge to establish, resulting in high design costs; (2) their fixed roles and workflows lack flexibility and are difficult to adapt to changing tasks; (3) scalability is poor, as adding new roles or restructuring requires manual intervention.

To address the limitations of manual design in terms of cost, flexibility, and scalability, recent research has proposed a new paradigm of LLM-designed MAS. Building on the remarkable understanding and planning capabilities of LLMs, these systems can automatically infer and generate adaptive agent roles and collaboration workflows based on input queries~\citep{chen2023autoagents}. Although LLM-designed MAS offers adaptability and flexibility, existing approaches still face key issues: (1) they often design architectures based on task types rather than specific queries, resulting in similar structures that fail to optimize performance for individual queries; (2) the automatically generated roles and workflows can be redundant or imprecise, lacking rigorous control and potentially undermining effectiveness in complex situations.

In our research, we propose a framework that synthesizes the strengths of both paradigms while mitigating their respective drawbacks. First, we establish a clear collaboration structure that applies to arbitrary queries, ensuring process rigor and avoiding the redundancy often found in automatically generated workflows. Second, within this structure, collaboration is tailored to each specific query, so that capability complementarity is realized at a fine-grained level rather than at the level of task types. WILC is itself a multi-agent system, but its agents are distinct LLMs with intrinsically different capability profiles rather than one backbone model differentiated by role prompts.

\subsection{Contextual Multi-Armed Bandit}\label{sec:MAB}
The multi-armed bandit (MAB) problem models sequential decision-making under uncertainty~\citep{robbins1952some,sutton1998reinforcement}. At each step, the decision-maker selects one arm and observes a reward, facing the classic exploration--exploitation trade-off: exploit the arm with the best observed return or explore uncertain arms that may perform better. A classic solution is the upper confidence bound (UCB) algorithm~\citep{auer2002finite}, which scores each arm $a$ at time $t$ by
$
    UCB_a(t) = \hat{\mu}_a(t) + \alpha\sqrt{\frac{2\ln t}{n_a(t)}},
$
where $\hat{\mu}_a(t)$ is the empirical mean reward, $n_a(t)$ is the number of times arm $a$ has been selected, the second term is an exploration bonus, which decreases as the arm is chosen more frequently. Here, $\alpha$ is a hyperparameter that controls the degree of exploration. MAB assumes a context-free scenario where the decision remains identical regardless of who the decision-maker is. This limitation becomes apparent in real-world applications such as recommendation systems, where the best option depends on user-specific context.

To address this limitation, researchers have proposed a variety of effective algorithms for the contextual multi-armed bandit (CMAB) problem, among which the LinUCB algorithm~\citep{li2010contextual} is one of the most classic. LinUCB extends UCB to handle contextual information by modeling the expected reward of arm $a$ as a linear function of its $d$-dimensional context $\mathbf{x}_{t,a}$ with a parameter vector $\boldsymbol{\theta}_a$. Namely, at time $t$, the expected reward of arm $a$ is given by:
$\mathbb{E}[R_{t,a}|\mathbf{x}_{t,a}] = \mathbf{x}_{t,a}^\top \boldsymbol{\theta}_a.$
Each time an arm is chosen and its reward observed, the parameter vector $\boldsymbol{\theta}_a$ is updated for the next arm-selection.

For each arm $a$, LinUCB maintains $\boldsymbol{A}_a = \boldsymbol{D}_a^\top \boldsymbol{D}_a + \boldsymbol{I}$ and $\boldsymbol{b}_a = \boldsymbol{D}_a^\top \boldsymbol{c}_a$, where $\boldsymbol{D}_a \in \mathbb{R}^{m \times d}$ is the design matrix whose rows are the $m$ contexts previously observed for arm $a$, and $\boldsymbol{c}_a \in \mathbb{R}^{m}$ is the corresponding reward vector. Ridge regression then yields the estimated parameter vector $\boldsymbol{\hat{\theta}}_a = \boldsymbol{A}_a^{-1} \boldsymbol{b}_a$, and each arm is scored by
\begin{equation}
UCB_a(t) = \mathbf{x}_{t,a}^\top\boldsymbol{\hat{\theta}}_a + \alpha\sqrt{\mathbf{x}_{t,a}^\top \boldsymbol{A}_a^{-1}\mathbf{x}_{t,a}},
\label{eq:linucb_ucb_standard}
\end{equation}
where the second term is a context-dependent exploration bonus and $\alpha$ controls its weight. A self-contained introduction is provided in Appendix~\ref{sec:linucb}.
After choosing an arm $a_t$ for the new context $\mathbf{x}_{t,a_t}$, the reward $R_{t,a_t}$ of this choice is received and thus a new observation ($a_t, \mathbf{x}_{t,a_t}, R_{t,a_t}$) is obtained. The algorithm then updates $\boldsymbol{A}_{a_t}$, $\boldsymbol{b}_{a_t}$, and the capability vector $\boldsymbol{\hat{\theta}}_{a_t}$ for the next arm-selection by:
\begin{equation}
  \boldsymbol{A}_{a_t} \leftarrow \boldsymbol{A}_{a_t} + \mathbf{x}_{t,a_t} \mathbf{x}_{t,a_t}^\top,\quad
  \boldsymbol{b}_{a_t} \leftarrow \boldsymbol{b}_{a_t} + R_{t,a_t} \mathbf{x}_{t,a_t},\quad
  \boldsymbol{\hat{\theta}}_{a_t} \leftarrow \boldsymbol{A}_{a_t}^{-1} \boldsymbol{b}_{a_t}.
\label{eq:update_A_b_standard} 
\end{equation}

In summary, the contextual bandit formulation aligns well with our setting: we select the most suitable model (i.e., arm) for an evolving query that encodes the latest problem-solving state (i.e., context). As an online learning algorithm, it further allows the capability estimates to be updated instantly from feedback obtained during problem solving, which reduces reliance on pre-collected training data.

\section{WILC: A Complementarity-Driven Framework}\label{sec:WILC}

This section presents our proposed framework for the wisdom of LLM crowds, \textit{Wisdom Integration of LLM Crowds (WILC)}. Given a crowd of $K$ LLMs, $\mathcal{W} = \{w_1, w_2, ..., w_K\}$, where each $w_k$ represents a unique LLM, our goal is to harness their collective intelligence such that the collaboration among these models achieves superior problem-solving performance compared to any single LLM operating alone. In the following, we begin by outlining the key design principles that facilitate complementarity in Section~\ref{sec:design-points}, followed by a detailed introduction of WILC. For convenience, a summary of major notation is provided in Appendix~\ref{sec:notation}.

\subsection{From Complementarity to Design Principles}\label{sec:design-points}

To translate the concept of \textit{relay-style complementarity} into a practical and actionable artifact, we distill two core design principles (DPs) that serve as the guiding tenets for the development of WILC.

\textbf{DP1: Iterative Reflection-and-Refinement.} Since complementarity is realized through a ``relay'' process, the system must support a \textit{multi-round and state-preserving workflow} in which models reflect on and refine prior drafts. As illustrated in the right panel of Figure~\ref{fig:motivation-example}, each handover requires a reflection that identifies the limitations of the current answer, together with a context that carries the initial task, the answer of the current model, and the identified bottleneck forward to the successor model.

\textbf{DP2: Complementarity-Driven Model Selection.} The effectiveness of relay-style collaboration hinges on the successor model, at each relay step, possessing capabilities that complement the bottleneck of the current model. Following prior studies on complementarity, we adapt its realization condition to the sequential relay process. Specifically, the successor model must appear well-suited to the identified bottleneck and, crucially, its contribution must be verified to yield a non-negative improvement to the existing answer. Hence, a \textit{complementarity verification mechanism} is necessary to ensure the collaborative process continues refining the answer until complementarity is exhausted or a satisfactory result is attained.

Together, these two principles translate relay-style complementarity into an operational design: DP1 establishes the state-preserving collaboration process, while DP2 governs model transitions by requiring successor workers to both fit the identified bottleneck and improve the evolving solution.

\subsection{Framework Formulation}\label{sec:framework-formulation}

\begin{figure}[!htb]
  \centering
  \caption{Illustration of the Iterative Reflection-and-Refinement Process.}
  \includegraphics[width=\textwidth]{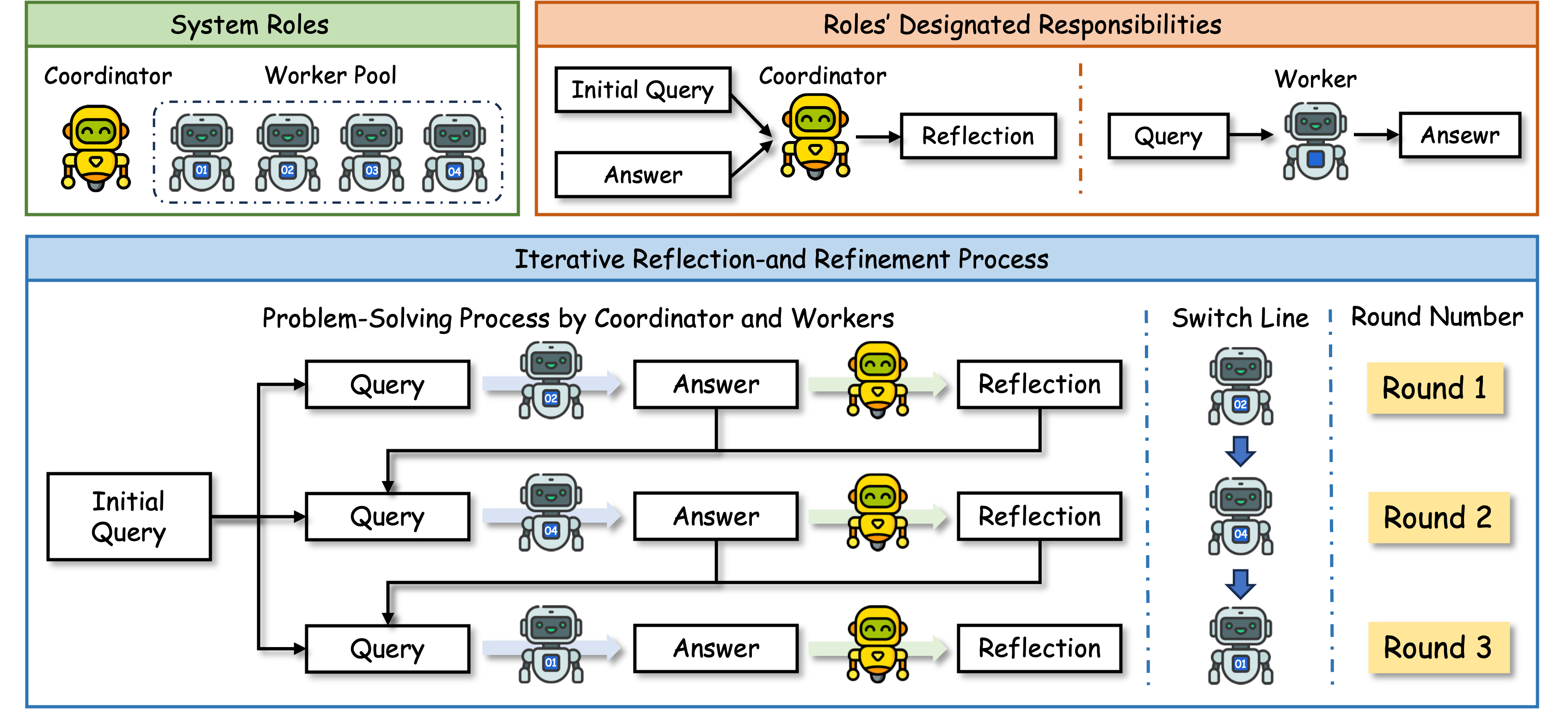}
  \label{fig:framework}
  \vspace{-4pt}
\end{figure}

Following DP1, our framework WILC conceptualizes the problem-solving process of LLM collaboration as an iterative reflection-and-refinement process. This process is designed to leverage the inherent reflective capability of LLMs, and iteratively refine the answer in a collaborative manner by combining the complementary advantages of different models. Figure \ref{fig:framework} provides a visual walkthrough of our framework, illustrating the system roles, role responsibilities, and the iterative reflection-and-refinement process. 

\subsubsection{System Roles and Their Responsibilities}
Our framework establishes two distinct roles in the problem-solving process: (a) a \textit{coordinator} and (b) a pool of \textit{workers}. 
Serving as executors, the worker pool comprises a heterogeneous set of LLMs responsible for generating solutions, leveraging their specialized capabilities to produce initial answers or refine prior drafts based on specific query contexts.
Complementing this, the coordinator, typically designated from the worker pool, acts as the orchestrator tasked with reflective supervision. By critically evaluating current solutions against the initial query, the coordinator diagnoses reasoning bottlenecks and generates diagnostic insights that serve as the ``baton'' for the relay, i.e., providing explicit directives that guide the selection of the next worker and the focus of subsequent refinement.
In practice, coordinator selection proceeds as follows: for an initial small batch of queries from the target task (we use 10 in our experiments), an arbitrary worker serves as the interim coordinator to generate reflections and compute proxy rewards. Based on the burn-in results from this batch, the worker that most frequently achieves the highest proxy reward is then designated as the permanent coordinator for subsequent queries.\footnote{This selection relies solely on the proxy reward derived from model-generated reflections (i.e., the count of identified issues), without access to ground-truth labels, and therefore introduces no data leakage from the test set.} Section~\ref{sec:robustness-checks} empirically validates that WILC's performance is insensitive to the choice of coordinator (strongest vs.\ weakest differs by only 0.66 pp), confirming that this arbitrary initialization does not degrade results.
This hierarchical division between execution and coordination ensures that the diverse generative strengths of the worker pool are effectively channeled through the coordinator's diagnostic oversight, thereby establishing a rigorous feedback loop for iterative refinement.


\subsubsection{Iterative Reflection-and-Refinement Process}

The iterative reflection-and-refinement process is initiated when a user submits an initial query, and proceeds through multiple rounds in which the answer is continually refined via collaborative efforts. As illustrated in the lower panel of Figure \ref{fig:framework}, we take three rounds as an example to demonstrate this multi-round process.

In round 1, the query is the initial query, and we select the most suitable worker (i.e., \textit{Worker 02} in the figure) to provide an initial answer to the initial query. The coordinator then reflects on this answer, rigorously evaluating its correctness and identifying any potential weaknesses (i.e., the \textit{bottleneck} in the current round). Based on this reflection, the query for the next round is constructed as the concatenation of the initial query, the current answer, and the reflection. Formally, the queries in different rounds are given by:
\begin{equation}
q_{r} = 
\begin{cases}
    q_0, & r=1 \\
    q_{r-1} + a_{r-1} + f_{r-1}, & r \geq 2,
\end{cases}
\end{equation}
where $r$ denotes the round number; $q_r$ is the query for round $r$; $q_0$ is the initial query; $q_{r-1}$, $a_{r-1}$, and $f_{r-1}$ refer to the query, answer, and reflection for round $r-1$, respectively.

In round 2, the updated query, enriched with the previous answer and reflection, is now used to guide the worker selection, aiming to identify the worker that is most suitable to address the bottleneck identified in the previous reflection. This iterative process is repeated, with each cycle involving worker selection, answer generation, and reflection generation. The worker selection step is guided by the complementarity verification mechanism, which will be detailed in Section \ref{sec:verification-mechanism}. In this three-round example, the resulting relay-style collaboration follows the sequence \textit{Worker 02 $\rightarrow$ Worker 04 $\rightarrow$ Worker 01}, as illustrated by the ``Switch Line'' in Figure \ref{fig:framework}.
Through this iterative reflection-and-refinement process, our framework enables the sequential collaboration among heterogeneous workers, whereby each worker is strategically selected to contribute its capabilities to address the bottleneck identified in preceding iterations. This dynamic allocation of workers' capabilities ensures that the collective wisdom of the LLM crowd is effectively harnessed, leading to increasingly refined answers in a principled and systematic manner.

\subsection{Complementarity Verification Mechanism} \label{sec:verification-mechanism}

The effectiveness of the iterative process described in Section \ref{sec:framework-formulation} hinges on worker selection at rounds $r \geq 2$: the selected worker must be able to address the bottleneck identified in the preceding answer. Following DP2, we therefore design a complementarity verification mechanism, evaluating whether a candidate worker can complement the current problem-solving state and deliver non-negative value in the next round.

\subsubsection{Definition: Complementarity of Workers}

Building upon the established definitions of complementarity in the literature~\citep{bansal2021does, donahue2022human}, we define the \textit{complementarity of workers} in our sequential collaboration scenario. After completing round $r-1$, let worker $A$ be the generator of answer $a_{r-1}$, and let worker $B$ be a candidate for round $r$. We regard worker $B$ as complementary to worker $A$ under the current state if: (a) worker $B$ appears well-suited to address the specific bottleneck $f_{r-1}$ based on its capabilities, and (b) after worker $B$ contributes its answer, there is a non-negative improvement to the quality of the answer relative to $a_{r-1}$. Note that worker $A$ and worker $B$ can be the same model in situations where a model possesses diverse abilities to address the bottleneck identified in its previous answer. This generalizes our framework to allow for both inter-model and intra-model complementarity, enabling repeated self-improvement by a single strong worker as well as collaboration among different models.

Aligned with the two facets of this definition, we design a dual-gate complementarity verification mechanism. The \textit{prospective complementarity fit (PCF)} gate models the fitness between each worker and the query containing the current problem-solving state (i.e., initial query, current answer, and identified bottleneck), and selects the most promising worker (Section~\ref{sec:PCF}). The \textit{posterior complementarity gain (PCG)} gate then checks whether the selected worker degrades the answer relative to the previous round, as assessed by the coordinator rather than against ground truth (Section~\ref{sec:PCG}).

\subsubsection{Prospective Complementarity Fit (PCF)}\label{sec:PCF}
Our scenario for worker selection aligns naturally with the contextual multi-armed bandit (CMAB) problem (as introduced in Section~\ref{sec:MAB}). Formally, let $\mathcal{W} = \{w_1, \dots, w_K\}$ denote the set of workers (arms). We model the problem-solving state at round $r$---whether an initial query\footnote{In the initial round where no prior answer exists, the initial query itself is conceptualized as the bottleneck state requiring resolution.} or a bottleneck identified after reflection---as a context vector $\mathbf{x}_{q_r} \in \mathbb{R}^d$, derived from the text embedding of the query. We assume the expected reward (i.e., answer quality) for choosing worker $w_k$ is linear with respect to the context: $\mathbb{E}[R_{r, w_k} | \mathbf{x}_{q_r}] = \mathbf{x}_{q_r}^\top \boldsymbol{\theta}_{w_k}$, where $\boldsymbol{\theta}_{w_k} \in \mathbb{R}^d$ is an unknown parameter vector representing the worker's capability profile. Building on this formulation, we adapt the LinUCB to learn $\boldsymbol{\theta}_{w_k}$ and capture the model-bottleneck fitness~\citep{li2010contextual}. Our adaptation proceeds in two distinct phases:

\textbf{Cold-Start Phase.} The few rounds of worker selection within a single problem-solving process yield too few observations to estimate worker capabilities reliably. We therefore introduce a cold-start phase that establishes a preliminary capability estimate before the framework is deployed.
Specifically, we construct an auxiliary dataset $\mathcal{Q}$ covering queries from diverse problem domains. To mirror the real context, the dataset contains two types of queries: (a) initial queries and (b) bottleneck-contextualized queries (i.e., queries containing the current bottleneck to be resolved, analogous to the inputs in rounds $r \geq 2$). We deliberately assign a larger proportion to the latter type, as our primary objective is to model each worker's capability in addressing specific bottlenecks (i.e., \textit{model-bottleneck fitness}).
Each worker is tasked with responding to the queries in this auxiliary dataset, and their answers are evaluated against the corresponding ground-truth answers. This evaluation produces a reward for each pair of worker and query, thereby providing the data for the initial estimation of every worker's capabilities.

Following the LinUCB formulation, the learning process of the parameter vector $\boldsymbol{\theta}_{w_k}$ of each worker $w_k$ proceeds as follows. Let $\boldsymbol{D}_{w_k}$ denote the design matrix with dimension $m \times d$, where each row corresponds to a context vector previously observed for worker $w_k$ across $m$ historical trials. Here, $m$ equals the number of queries in the auxiliary dataset. Let $\boldsymbol{c}_{w_k} \in \mathbb{R}^m$ denote the corresponding reward vector. Based on LinUCB, the estimated parameter vector $\boldsymbol{\hat{\theta}}_{w_k}^{\text{cold}}$ is given by:
\begin{equation}
\label{eq:linucb_theta_est_wilc}
\boldsymbol{\hat{\theta}}_{w_k}^{\text{cold}} = (\boldsymbol{D}_{w_k}^\top \boldsymbol{D}_{w_k} + \boldsymbol{I})^{-1} \boldsymbol{D}_{w_k}^\top \boldsymbol{c}_{w_k} = \boldsymbol{A}_{w_k}^{-1} \boldsymbol{b}_{w_k},
\end{equation}
where $\boldsymbol{I}$ denotes the $d \times d$ identity matrix, and for simplicity, let $\boldsymbol{A}_{w_k} = \boldsymbol{D}_{w_k}^\top \boldsymbol{D}_{w_k} + \boldsymbol{I}$ and $\boldsymbol{b}_{w_k} = \boldsymbol{D}_{w_k}^\top \boldsymbol{c}_{w_k}$.

\textbf{Problem-Solving Phase.} 
During the problem-solving phase, the LinUCB algorithm leverages the learned capability parameters to select the worker with the highest fitness for the query at each round. 
Given the query $q_r$ at round $r$, we first transform it into a $d$-dimensional context vector $\mathbf{x}_{q_r} \in \mathbb{R}^d$ (utilizing an embedding model), and then select the worker with the highest upper confidence bound (UCB) score by the following equation, passing it to the PCG gate for further verification:
\begin{equation}
\label{eq:linucb_decision_wilc}
w_r = \arg\max_{w_k \in \mathcal{W}} \mathrm{UCB}_{w_k}(r) = \arg\max_{w_k \in \mathcal{W}} \left( \mathbf{x}_{q_r}^\top \boldsymbol{\hat{\theta}}_{w_k} + \alpha \sqrt{\mathbf{x}_{q_r}^\top \boldsymbol{A}_{w_k}^{-1} \mathbf{x}_{q_r}} \right),
\end{equation}
where $\alpha$ is a hyperparameter controlling the exploration-exploitation trade-off.

After a worker is finally chosen and its reward observed, LinUCB further refines its capability estimates by incorporating this new observation, continually updating its belief to inform future selections.
However, during the multi-round process of solving an initial query, the true reward for each intermediate answer is unavailable, as the ground-truth answer is unknown. To address this challenge, we introduce two key domain adaptations~\citep{abbasi2024pathways} in the implementation of LinUCB during this phase.

First, we introduce a \textit{proxy reward mechanism} that converts the coordinator's reflection into a scalar reward signal. During reflection generation, the coordinator is required to enumerate distinct issues in a structured format, and we assign a penalty value of $\gamma$ (e.g., 0.1) to each identified issue. Given a reflection with $s$ issues, the proxy reward $R_{r,w_k}$ is computed as:
\begin{equation}
  R_{r,w_k} =\max\{
  1 - s\gamma, 0
  \}
\end{equation}
which yields a reward in the interval $[0, 1]$. The proxy reward thus supplies a reward signal at every round without access to ground truth. When there is no issue identified (i.e., $R_{r,w_k} = 1$), we consider the answer to be satisfactory and the iteration process can be terminated.

Second, we introduce a \textit{weighted updating mechanism} that balances the general capability estimates learned from true rewards in the cold-start phase against the more immediate but potentially biased proxy feedback from the current query. Formally, at round $r$, after selecting worker $w_k$ and observing reward $R_{r,w_k}$, we obtain a new observation tuple $(w_k, \mathbf{x}_{q_r}, R_{r,w_k})$. We adapt the standard LinUCB update in Equation~(\ref{eq:update_A_b_standard}) by incorporating a weighting parameter $\beta \in [0,1]$:
\begin{equation}
\hat{\boldsymbol{\theta}}_{w_k} \leftarrow \beta\boldsymbol{\hat{\theta}}_{w_k}^{\text{cold}} + (1-\beta)\boldsymbol{A}_{w_k}^{-1}\boldsymbol{b}_{w_k}\text{,}
\label{eq:update_theta_weighted}
\end{equation}
where $\boldsymbol{\hat{\theta}}_{w_k}^{\text{cold}}$ denotes the capability estimate of worker $w_k$ learned in the cold-start phase. A larger $\beta$ places greater weight on the more reliable and broadly learned prior estimate, whereas a smaller $\beta$ allows the update to respond more strongly to query-specific feedback, which is more responsive to the focal query but also noisier. Thus, $\beta$ governs a trade-off between exploiting generalizable prior knowledge and adapting to local query-specific signals, thereby mitigating proxy-reward bias and improving the robustness of worker selection. The adaptive selection of $\beta$ will be discussed in Section~\ref{subsubsec:burnin}.

Through iteratively selecting promising workers and continually refining capability estimates, PCF ensures a rigorous and adaptive approach to select the most promising worker who fits best with the current query.

\subsubsection{Posterior Complementarity Gain (PCG)}\label{sec:PCG}

The PCG gate serves as an additional safeguard mechanism for validating the worker selected by the PCF gate, checking that its contribution does not degrade the assessed quality of the answer.
Its verification procedure is straightforward: PCG first records the proxy reward of the previous answer, then evaluates the proxy reward for the current worker's submission. If the current worker's proxy reward is no lower than that of the preceding answer (i.e., $R_{r} \geq R_{r-1}$), the worker is deemed to have provided a non-negative improvement and the iteration continues. In contrast, if the current worker's proxy reward is strictly lower ($R_{r} < R_{r-1}$), the worker fails to pass the complementarity verification, and the previous answer is retained. If even the most promising worker is unable to maintain or improve the current proxy reward, we take this as evidence that the remaining complementarity in the worker pool is unlikely to be realized under the current assessment, and the iteration terminates.
It is worth noting that this termination mechanism endows WILC with an appealing adaptive property: when the PCG gate halts the collaboration after the burn-in round (i.e., a single round), WILC operates as an evaluate-then-select mechanism---all workers produce responses during the burn-in, and the best-performing worker's output is directly returned based on proxy evaluation---investing additional rounds only when iterative refinement is expected to yield meaningful gains. This adaptive depth control also brings a practical cost advantage: simpler queries that are resolved in a single round incur substantially fewer API calls, while the full multi-round overhead is reserved for genuinely difficult queries where the additional investment is justified by measurable performance gains. We provide a detailed cost-effectiveness analysis in Section~\ref{sec:cost-analysis}.

\subsection{Designs for Robustness Enhancement}\label{sec:robust-designs}
Although the core WILC framework enables principled complementarity-driven collaboration, its performance may still be challenged by query-level distributional shifts and selection uncertainty in practice. To address these issues, we further introduce two robustness-oriented designs: a burn-in initialization strategy and a one-step forward search (OSFS) mechanism.
\subsubsection{Burn-In Initialization Strategy}
\label{subsubsec:burnin}
To mitigate potential distributional shifts between the cold-start training data and the novel queries encountered during actual problem solving, we introduce a \textit{burn-in initialization} strategy at the outset of the process. 
Operationally, at the first round for a given query, all workers are required to generate an answer to the initial query. The corresponding proxy rewards for each worker's answer (i.e., $R_{1,w_k}$ for worker $w_k$) are then calculated, and the adapted LinUCB algorithm for the problem-solving phase is applied to update the capability parameters of all workers using these newly obtained observations. At the same time, for the first round of worker selection, the worker with the highest proxy reward is directly chosen to proceed as the initial executor for the problem-solving process.

The burn-in strategy improves the effectiveness and robustness of WILC through two aspects. \textit{First}, it enables the framework to quickly adapt its capability estimation to novel queries that may deviate significantly from the cold-start training data, which matters in real-world deployments where incoming queries span a wide variety of types and complexities. This immediate recalibration updates worker parameters to better match the current query, rather than relying only on less relevant prior knowledge. \textit{Second}, it enables more informed initial worker selection. By collecting actual response and performance on the query at hand before making any exclusionary decisions, WILC reduces the risk of sub-optimal initial selections, thereby providing a stronger starting point that facilitates a smoother refinement process in subsequent rounds.

Furthermore, the proxy-reward signal in the burn-in phase provides a principled way for dynamically setting the weighting parameter $\beta$ in Equation (\ref{eq:update_theta_weighted}). 
The central idea is to compare the ranking of workers' UCB scores, derived from LinUCB parameters trained in the cold-start phase, with the proxy reward rankings for new queries. If the two rankings are highly consistent, the cold-start estimates remain credible for the current query and we assign a larger $\beta$ to rely more heavily on the cold-start parameters; conversely, if the rankings diverge, we reduce $\beta$ and place more weight on the newly observed feedback. Concretely, we compute the Spearman's rank correlation coefficient~\citep{sedgwick2014spearman}, denoted as $\rho$, between the UCB score list and the proxy reward list, and set $\beta=(1+\rho)/2$. Because $\rho \in [-1,1]$, this transformation ensures $\beta \in [0,1]$.

\subsubsection{One-Step Forward Search (OSFS)}

To further enhance the robustness of worker selection within the framework, we introduce a one-step forward search design. Specifically, at round $r$, after the worker $w_r$ is selected based on PCF and PCG, and their answer $a_r$, reflection $f_r$, and proxy reward $R_{r,w_r}$ are obtained, we add an additional forward simulation if the current selected worker differs from the previous one at round $r-1$ (i.e., $w_r \neq w_{r-1}$). In this simulation, the previous worker $w_{r-1}$ is also tasked with generating an answer to the current query $q_r$, yielding a simulated answer $a'_{r}$, reflection $f'_{r}$, and proxy reward $R'_{r,w_{r-1}}$. If the simulated proxy reward exceeds that of the current worker, i.e., $R'_{r,w_{r-1}} > R_{r,w_r}$, we thus use the simulated results to overwrite the current round's outputs, i.e.,
\begin{equation}
(a_r, f_r, R_{r,w_r}) \leftarrow (a'_{r}, f'_{r}, R'_{r,w_{r-1}}).
\end{equation}

This design addresses two distinct sources of risk that arise when the framework switches workers. \textit{First}, it mitigates selection uncertainty. The worker selection within the PCF gate is based on the LinUCB algorithm, whose capability estimates are themselves updated from proxy rewards rather than true rewards, so a recommended switch may turn out to be suboptimal. The one-step forward search introduces a fallback mechanism that enables the system to revert to the previous worker when empirical evidence suggests superior performance, thereby reducing the risk of acting on a misleading capability estimate. \textit{Second}, it promotes reflection-response alignment. The reflection $f_{r-1}$ that drives round $r$ is generated specifically for the answer produced by worker $w_{r-1}$, and is subsequently incorporated into the query $q_r$. There are therefore cases where, despite the LinUCB algorithm recommending a new worker based on updated capability estimates, $w_{r-1}$ possesses a more concrete understanding of the bottleneck raised in its own reflection and can address it more effectively. By evaluating both candidates on the current query, the forward search allows the framework to retain whichever contribution the proxy reward actually favors, rather than committing to the switch in advance.

\subsection{WILC Framework Overview}\label{sec:wilc-overview}
Algorithm~\ref{alg:wilc} provides a unified, end-to-end specification of the WILC framework, integrating all components introduced in Sections~\ref{sec:framework-formulation}--\ref{sec:robust-designs}. Notably, the operational ordering in Algorithm~\ref{alg:wilc} differs from the conceptual organization of the preceding subsections, and we clarify this distinction here. Sections~\ref{sec:verification-mechanism} and~\ref{sec:robust-designs} are organized by \textit{design purpose}: the complementarity verification mechanism (PCF and PCG gates) and the robustness enhancement designs (burn-in and one-step forward search) are treated as separate, self-contained components, each elaborated in full before the next is introduced. Algorithm~\ref{alg:wilc}, in contrast, presents these components in their \textit{execution order}. In the problem-solving phase, the burn-in initialization strategy (Section~\ref{subsubsec:burnin}) governs round $r=1$, during which all workers respond to the initial query and capability estimates are updated for all. From round $r \geq 2$ onward, the PCF gate (Section~\ref{sec:PCF}), one-step forward search (Section~\ref{sec:robust-designs}), PCG gate (Section~\ref{sec:PCG}), and weighted capability update are applied in sequence within each iteration. This interleaving of components across rounds reflects how WILC operates in practice, and Algorithm~\ref{alg:wilc} serves as the definitive reference for implementing the full framework.

\begin{algorithm}[!htb]
  \caption{Wisdom Integration of LLM Crowds (WILC)}
  \label{alg:wilc}
  \small
  \renewcommand{\baselinestretch}{1.0}\selectfont
  \renewcommand{\algorithmicrequire}{\textbf{Input:}}
  \renewcommand{\algorithmicensure}{\textbf{Output:}}
  \begin{algorithmic}[1]
  \REQUIRE Worker pool $\mathcal{W} = \{w_1, \dots, w_K\}$; coordinator; auxiliary dataset $\mathcal{Q}$; exploration parameter $\alpha$; issue penalty $\gamma$; maximum rounds $r_{\max}$
  \ENSURE Final answer $a^*$ for each query $q_0$ (i.e., the initial query)
  \STATE \textbf{--- Cold-Start ---}
  \FOR{each worker $w_k \in \mathcal{W}$}
      \STATE Let $w_k$ answer all queries in $\mathcal{Q}$; evaluate against ground truth to obtain rewards
      \STATE Estimate capability vector: $\boldsymbol{\hat{\theta}}_{w_k}^{\text{cold}}$ with Eq.~(\ref{eq:linucb_theta_est_wilc})
  \ENDFOR
  \STATE \textbf{--- Problem-Solving ---}
  \STATE \textit{// Round $r=1$: Burn-In Initialization}
  \STATE Construct query: $q_1 \leftarrow q_0$; Embed $q_1$ to obtain context vector $\mathbf{x}_{q_1} \in \mathbb{R}^d$
  \FOR{each worker $w_k \in \mathcal{W}$}
      \STATE Obtain $(a_1, f_1, R_{1,w_k}) \leftarrow$ 
      worker $w_k$ generates answer $a_1$ to $q_1$; coordinator reflects to obtain $f_1$ enumerating $s$ issues; compute proxy reward $R_{1,w_k} \leftarrow \max\{1 - s\gamma,\; 0\}$
  \ENDFOR
  \STATE Compute Spearman's $\rho$ between UCB scores and $\{R_{1,w_k}\}$; set $\beta 
  \leftarrow (1+\rho)/2$
  \FOR{each worker $w_k \in \mathcal{W}$}
      \STATE Update: $\boldsymbol{\hat{\theta}}_{w_k}$ with Eq.~(\ref{eq:update_theta_weighted})
  \ENDFOR
  \STATE Select initial worker: $w_1 \leftarrow \arg\max_{w_k \in \mathcal{W}} R_{1,w_k}$; record $(a_1, f_1, R_{1,w_1})$
  \IF{$R_{1,w_1} = 1$}
      \RETURN $a^* \leftarrow a_1$ \hfill \textit{\{Satisfactory at round 1\}}
  \ENDIF
  \STATE \textit{// Rounds $r \geq 2$: Iterative Reflection-and-Refinement}
  \FOR{$r = 2, \dots, r_{\max}$}
      \STATE Construct query: $q_r \leftarrow \{q_{r-1}, a_{r-1}, f_{r-1}\}$; Embed $q_r$ to obtain context vector $\mathbf{x}_{q_r} \in \mathbb{R}^d$
      \STATE Select worker: $w_r$ with Eq.~(\ref{eq:linucb_decision_wilc}) \hfill \textit{\{Prospective Complementarity Fit (PCF)\}}
      \STATE Obtain $(a_r, f_r, R_{r,w_r}) \leftarrow$ worker $w_r$ generates answer $a_r$ to $q_r$; coordinator reflects to obtain $f_r$ enumerating $s$ issues; compute proxy reward: $R_{r,w_r} \leftarrow \max\{1 - s\gamma,\; 0\}$
      \STATE \textit{// One-step Forward Search}
      \IF{$w_r \neq w_{r-1}$}
          \STATE Obtain $(a'_r, f'_r, R'_{r,w_{r-1}}) \leftarrow$ worker $w_{r-1}$ generates answer $a'_r$ to $q_r$; coordinator reflects to obtain $f'_r$ enumerating $s$ issues; compute proxy reward: $R'_{r,w_{r-1}} \leftarrow \max\{1 - s\gamma,\; 0\}$
          \IF{$R'_{r,w_{r-1}} > R_{r,w_r}$}
              \STATE $(a_r, f_r, R_{r,w_r}) \leftarrow (a'_r, f'_r, R'_{r,w_{r-1}})$
              \STATE $w_r \leftarrow w_{r-1}$
          \ENDIF
      \ENDIF
      \IF{$R_{r,w_r} < R_{r-1,w_{r-1}}$} 
      \RETURN $a_{r-1}$ (complementarity exhausted) \hfill \textit{\{Posterior Complementarity Gain (PCG)\}}
      \ENDIF
      \STATE Update: $\boldsymbol{\hat{\theta}}_{w_r}$ with Eq.~(\ref{eq:update_theta_weighted})
      \IF{$R_{r,w_r} = 1$}
          \RETURN $a^* \leftarrow a_r$ \hfill \textit{\{Satisfactory, no issues identified\}}
      \ENDIF
  \ENDFOR
  \RETURN $a^* \leftarrow a_r$
  \end{algorithmic}
  \end{algorithm}

\section{Empirical Evaluations}\label{sec:Experiments}
In this section, we comprehensively evaluate the proposed WILC framework for different LLM crowds across diverse tasks. First, we describe the experimental setup (including the evaluation tasks, LLM crowds, baseline methods, and other implementation details). Then, we report and discuss the experimental results.

\subsection{Experimental Setup}

\subsubsection{Evaluation Tasks}
We evaluate the proposed WILC framework across four representative task domains that capture core forms of enterprise knowledge work: code generation, mathematical reasoning, general knowledge reasoning, and data visualization. For each domain, we select a corresponding typical benchmark dataset: HumanEval for code generation, MATH-500 for mathematical reasoning, MMLU for general knowledge reasoning, and VisEval for data visualization. The diversity of these datasets enables us to evaluate the performance of the proposed framework in a comprehensive manner. The descriptions of these benchmarks are as follows:

\begin{itemize}
    \item \textbf{HumanEval}~\citep{chen2021evaluating}: HumanEval tests the ability of LLMs to generate correct code from natural language prompts. Developed by OpenAI, it consists of 164 hand-written Python programming problems, each paired with a reference answer and test cases. The evaluation metric is Pass@k, which measures the probability that at least one of the top-k generated answers will pass the tests. Following common practice, we report Pass@1, i.e., the percentage of problems where the model generates a correct answer on the first attempt.
    \item \textbf{MATH-500}~\citep{hendrycks2021measuring}: MATH-500 evaluates the mathematical reasoning and problem-solving proficiency of LLMs. It contains 500 challenging problems spanning five core mathematical domains (algebra, combinatorics, geometry, number theory, and precalculus), derived from high-level high school competitions such as the American Mathematics Competitions (AMC) and the American Invitational Mathematics Examination (AIME). Rather than testing simple arithmetic, these problems require complex, multi-step reasoning and abstract problem-solving skills. The benchmark is evaluated using \textit{exact match} on the final numerical answer, demanding precision in both the reasoning process and the final calculation.
    \item \textbf{MMLU}~\citep{hendryckstest2021}: Massive multitask language understanding (MMLU) is a comprehensive benchmark for evaluating models' multitask understanding across 57 diverse subjects, ranging from elementary mathematics and U.S. history to computer science, and spanning difficulty levels from high school to professional domains. It comprises 15,908 multiple-choice questions and is designed to measure a model's ability to generalize knowledge without extensive task-specific training. Given the extensive scale of the dataset and the computational costs associated with LLM inference, we randomly selected 50 questions from each subject, resulting in a total of 2,850 questions. The evaluation metric is \textit{accuracy}, i.e., the percentage of questions answered correctly.
    \item \textbf{VisEval}~\citep{chen2024viseval}: VisEval is a large-scale benchmark for natural language-to-visualization generation (NL2VIS), containing 1,150 unique visualizations and 2,524 $\langle$NL, VIS$\rangle$ pairs drawn from 146 databases. It evaluates LLMs' ability to generate accurate Python visualization code from natural-language instructions given one or more data files, and assesses each generated visualization along three dimensions: \textit{validity}, i.e., whether the code successfully renders a visualization; \textit{legality}, i.e., whether the visualization satisfies the query requirements; and \textit{readability}, i.e., how effectively it presents the underlying information. The benchmark's original \textit{pass rate} aggregates results by visualization; we instead evaluate performance at the level of individual queries and define \textit{accuracy} as the proportion of queries whose generated result passes both the validity and legality checks.
\end{itemize}

\subsubsection{LLMs} 
We evaluate WILC on two distinct LLM crowds, using open-source models with parameter scales of approximately 14 billion (14B) and 30 billion (30B), as detailed in Table \ref{tab:llms}.
Building on the deployment considerations noted in Section~\ref{sec:Intro}, we situate our evaluation in a setting where organizations privately deploy open-source models, motivated by data privacy, cost efficiency, and customization flexibility. This setting instantiates the capability heterogeneity discussed earlier in a concrete form: the models we deploy come from different developers and training pipelines~\citep{shnitzer2023large, lu2024routing}, so their relative strengths vary across task types and no single member of the crowd dominates the others. This provides a natural foundation for the relay-style complementarity that WILC is designed to exploit.
Our selection of the 14B and 30B parameter scales is further motivated by practical hardware constraints. Models in this range can typically fit within the memory envelope of widely available commercial GPUs (e.g., 24GB or 48GB VRAM), allowing a crowd of such models to be served via horizontal scaling with multiple standard cards in parallel. In contrast, deploying ultra-large models (e.g., 70B or larger) often necessitates vertical scaling with specialized high-memory hardware or complex multi-GPU tensor parallelism, creating a significantly higher barrier to entry.\footnote{In addition to private enterprise deployment, on-device and edge deployment is increasingly viewed as a promising application scenario for LLMs, where resource constraints make medium-sized or smaller models particularly relevant.}
Our setup therefore simulates a pragmatic enterprise scenario built on accessible, heterogeneous models rather than a single costly hardware configuration.
\begin{table}[!htp]
  \caption{Overview of LLMs Used in Our Experiments}
  \centering
  \small
  \begin{tabular*}{0.9\textwidth}{@{\extracolsep{\fill}}ccc||ccc@{}}
    \toprule
    \multicolumn{3}{c||}{\textbf{14B Scale}} & \multicolumn{3}{c}{\textbf{30B Scale}} \\
    \midrule
    \textbf{Model Name} & \textbf{\# Params} & \textbf{Affiliation} & \textbf{Model Name} & \textbf{\# Params} & \textbf{Affiliation} \\
    \midrule
    Qwen2.5: 14B & 14.8B & Alibaba & GLM4: 32B & 32B & Zhipu AI \\
    Phi4: 14B & 14.7B & Microsoft & Qwen2.5: 32B & 32.8B & Alibaba \\
    DeepSeek-R1: 14B & 14.8B & DeepSeek & Gemma2: 27B & 27.2B & Google \\
    \bottomrule
  \end{tabular*}
  \label{tab:llms}
\end{table}

\subsubsection{Baseline Methods}
To comprehensively evaluate the effectiveness of WILC, we compare it with several carefully selected baselines that represent different approaches to improving LLM performance.

\begin{itemize}
    \item \textbf{Single Execution}: As a fundamental baseline, we first evaluate each worker model individually through single execution, which establishes the performance benchmark of standalone models and provides insights into their inherent capabilities. See Appendix \ref{sec:single-execution} for the single execution prompts for each task. 
    \item \textbf{ReAct}~\citep{yao2022react}: ReAct is a framework for task-solving agents that combines reactive reasoning with action execution. It allows the model to reason about the current state of the task and take actions to progress towards the goal. Since ReAct is originally designed for LLM-based agents, we modify it for our setting by introducing a tailored ReAct prompt template, as detailed in Appendix \ref{sec:react}.
    \item \textbf{Reflexion}~\citep{shinn2023reflexion}: Reflexion is a framework for language-based agents that enables them to improve through ``verbal reinforcement'': after an agent produces a result, it generates a linguistic self-reflection on its performance, and uses it as context to guide future attempts. This baseline is particularly relevant to our research as it shares our framework's emphasis on iterative reflection-and-refinement, while relying on a single model rather than leveraging the collective wisdom of multiple LLMs.
    \item \textbf{Self-ensemble}: We implement a self-ensemble baseline in the spirit of universal self-consistency (USC)~\citep{chen2024universal}. Specifically, for each query, we invoke the same model three times to generate multiple candidate responses, and then use an additional synthesis prompt to identify the most consistent answer among them. This implementation is closely aligned with USC when the final answer is selected from the candidate set, while also allowing a lightweight synthesis-based extension in our experimental setting. See Appendix \ref{sec:self-ensemble-and-heterogeneous-ensemble} for the general response-synthesis prompt used across tasks.
    \item \textbf{Heterogeneous Ensemble}: Heterogeneous ensemble aims to combine the wisdom from heterogeneous models. We design this baseline in the spirit of LLM-Blender~\citep{jiang2023llm}, which uses a fusion model to aggregate the responses from multiple models. However, LLM-Blender needs to fine-tune a pretrained LLM as the fusion model, which is computationally expensive. For the consideration of computational cost, we adopt a prompt-based fusion strategy to merge the responses into a final answer. Specifically, the prompt for aggregation is the same as that used in the self-ensemble baseline. Also, to ensure a fair comparison with WILC, the fusion is performed by the coordinator LLM used in WILC.
\end{itemize}

\subsubsection{Implementation Details}
Unless otherwise noted, all reported results are averaged over five runs. Experiments were conducted on a high-performance GPU cluster primarily composed of NVIDIA A100 GPUs, with inference backend \texttt{llama.cpp}.\footnote{\url{https://github.com/ggml-org/llama.cpp}}
All models are quantized to 4-bit precision (i.e., Q4\_K\_M scheme) to improve memory efficiency and support practical multi-model deployment~\citep{lin2024awq}.
This setup enables the parallel execution of multiple LLM workers at both the 14B and 30B scales. The temperature is set to 0.8 for all models, following common practice in prior work.

The cold-start phase relies on a set of diverse queries $\mathcal{Q}$ with standard answers, through which each worker $w_k$ answers these queries and evaluates its performance (i.e., real reward), thereby obtaining an initial understanding of the worker's capabilities. For the initial queries in $\mathcal{Q}$, we use an uncontaminated benchmark dataset called LiveBench \citep{white2024livebench}, the data of which is carefully filtered with the aim of avoiding test contamination, namely, queries that appear in the pre-training data typically used by LLMs are excluded. Specifically, the queries in LiveBench are from recently released math competitions, arXiv papers, news articles, and datasets, and include the clean versions of Big-Bench Hard, AMPS, and IFEval benchmarks. 
It contains 1436 queries from six distinct domains with the following numbers of questions: 200 for reasoning, 150 for data analysis, 128 for coding, 400 for instruction following, 368 for math, and 190 for language. These diverse and comprehensive queries provide a solid foundation for capturing the initial capabilities of each worker in the cold-start phase. Next, we further construct queries with bottlenecks by prompting each worker to self-reflect on these initial queries for two rounds, ultimately generating an additional set of queries with bottlenecks that is twice the number of the initial queries. 

For the embedding model that converts textual queries into embeddings, we use a high-performing open model called \texttt{nomic-embed-text:v1.5}~\citep{nussbaum2025nomic}. The default embedding dimension of the model is 768 and its key advantage over others is its large context window of up to 8,192 tokens, thereby enabling it to handle extended content. After embedding the queries, we use the principal component analysis (PCA) to reduce the dimensionality of the embeddings to 32. Finally, for the hyperparameters of the LinUCB algorithm, we set the exploration parameter $\alpha = 0.1$ for both the cold-start phase and the problem-solving phase, and dynamically set the weighting parameter $\beta$ based on Spearman's rank correlation coefficient between the UCB score list and the proxy reward list.

\subsection{Performance: WILC vs. Baselines}

In this section, we report the experimental results of WILC and the baselines on the four benchmarks, as shown in Tables \ref{tab:main_results_14b} and \ref{tab:main_results_30b}.\footnote{Notably, the 14B crowd achieves higher absolute performance than the 30B crowd on HumanEval, MATH-500, and MMLU. This is primarily because the 14B crowd includes DeepSeek-R1: 14B, a reasoning-specialized model distilled from a much larger teacher, which exhibits exceptionally strong performance on these well-established reasoning and knowledge benchmarks. On VisEval, however, the task requires generating complete visualization code from natural language specifications (NL2VIS), a less standard task type that demands broad code generation capabilities (data manipulation, visualization library APIs, layout design) beyond chain-of-thought reasoning. Here, the distillation advantage of DeepSeek-R1: 14B diminishes, and the 30B crowd's larger model capacity leads to substantially better performance.} Our empirical evaluation reveals several key findings that demonstrate the effectiveness of the proposed approach. \textit{First}, WILC delivers the strongest overall performance across both the 14B and 30B model scales, with a clear advantage over competing methods. Across the eight experimental configurations (four evaluation benchmarks $\times$ two model scales), WILC achieves the best performance in seven cases, generally by a meaningful margin. For example, in the 14B-scale setting, WILC achieves the highest scores on HumanEval (95.48\%), MATH-500 (89.98\%), MMLU (88.92\%), and VisEval (76.72\%). This consistent superiority supports the effectiveness of our complementarity-based collaboration mechanism in harnessing collective intelligence from heterogeneous LLM crowds.

\textit{Second}, our method consistently surpasses both self-improvement baselines (ReAct and Reflexion) and ensemble approaches (self-ensemble and heterogeneous ensemble), exhibiting greater reliability and stability across diverse experimental configurations. Although a few baselines achieve competitive results in certain experiments, their performance may fall short when evaluated on other models or datasets. For example, while ReAct using GLM4: 32B achieves competitive performance on HumanEval at the 30B scale (92.10\%), it falls short on MATH-500 (83.96\% vs. 88.56\%) and significantly underperforms on VisEval (73.38\% vs. 84.06\%). Similarly, the heterogeneous ensemble performs comparably to WILC on HumanEval (94.34\%), but shows inconsistent results across other benchmarks. In contrast, WILC demonstrates consistently robust and substantial performance improvements across all evaluated benchmarks and model scales. On average, WILC surpasses the strongest baseline by 3.45 percentage points at the 14B scale and by 2.83 percentage points at the 30B scale, highlighting its effectiveness across diverse tasks and models.

\definecolor{macaronPink}{RGB}{255,192,203}
\definecolor{macaronGreen}{RGB}{189,236,182}
\definecolor{macaronYellow}{RGB}{255,244,179}
\definecolor{macaronBlue}{RGB}{181,218,255}
\definecolor{macaronPurple}{RGB}{222,205,255}
\definecolor{macaronOrange}{RGB}{255,224,178}
\definecolor{macaronCyan}{RGB}{178,235,242}

\begin{table}[h!]
    \caption{Performance Comparison on Four Benchmarks: WILC vs. Baselines (14B Scale)}
    \centering
    \small
    \setlength{\tabcolsep}{9pt}
    \begin{tabular}{l|lllll}
      \toprule
      \textbf{Method} & \textbf{HumanEval} & \textbf{MATH-500} & \textbf{MMLU} & \textbf{VisEval} & \textbf{AVG.} \\
      \midrule
      \cellcolor{macaronPink!25}Single Execution (Qwen2.5: 14B) & \cellcolor{macaronPink!25}75.98\sym{***} & \cellcolor{macaronPink!25}72.68\sym{***} & \cellcolor{macaronPink!25}79.08\sym{***} & \cellcolor{macaronPink!25}59.46\sym{***} & \cellcolor{macaronPink!25}71.80\sym{***} \\
      \cellcolor{macaronPink!25}Single Execution (Phi4: 14B) & \cellcolor{macaronPink!25}81.70\sym{***} & \cellcolor{macaronPink!25}76.28\sym{***} & \cellcolor{macaronPink!25}78.72\sym{***} & \cellcolor{macaronPink!25}60.90\sym{***} & \cellcolor{macaronPink!25}74.41\sym{***} \\
      \cellcolor{macaronPink!25}Single Execution (DeepSeek-R1: 14B) & \cellcolor{macaronPink!25}85.58\sym{***} & \cellcolor{macaronPink!25}88.96\sym{**} & \cellcolor{macaronPink!25}81.32\sym{***} & \cellcolor{macaronPink!25}52.66\sym{***} & \cellcolor{macaronPink!25}77.13\sym{***} \\
      \\[-0.8em]
      \cellcolor{macaronGreen!25}ReAct (Qwen2.5: 14B) & \cellcolor{macaronGreen!25}81.20\sym{***} & \cellcolor{macaronGreen!25}67.10\sym{***} & \cellcolor{macaronGreen!25}81.28\sym{***} & \cellcolor{macaronGreen!25}63.20\sym{***} & \cellcolor{macaronGreen!25}73.20\sym{***} \\
      \cellcolor{macaronGreen!25}ReAct (Phi4: 14B) & \cellcolor{macaronGreen!25}88.16\sym{***} & \cellcolor{macaronGreen!25}76.12\sym{***} & \cellcolor{macaronGreen!25}85.70\sym{***} & \cellcolor{macaronGreen!25}71.74\sym{**} & \cellcolor{macaronGreen!25}80.43\sym{***} \\
      \cellcolor{macaronGreen!25}ReAct (DeepSeek-R1: 14B) & \cellcolor{macaronGreen!25}93.22\sym{**} & \cellcolor{macaronGreen!25}89.32 & \cellcolor{macaronGreen!25}85.00\sym{***} & \cellcolor{macaronGreen!25}52.08\sym{***} & \cellcolor{macaronGreen!25}79.91\sym{***} \\
      \\[-0.8em]
      \cellcolor{macaronYellow!25}Reflexion (Qwen2.5: 14B) & \cellcolor{macaronYellow!25}74.28\sym{***} & \cellcolor{macaronYellow!25}74.24\sym{***} & \cellcolor{macaronYellow!25}78.34\sym{***} & \cellcolor{macaronYellow!25}58.06\sym{***} & \cellcolor{macaronYellow!25}71.23\sym{***} \\
      \cellcolor{macaronYellow!25}Reflexion (Phi4: 14B) & \cellcolor{macaronYellow!25}81.46\sym{***} & \cellcolor{macaronYellow!25}75.96\sym{***} & \cellcolor{macaronYellow!25}78.40\sym{***} & \cellcolor{macaronYellow!25}59.88\sym{***} & \cellcolor{macaronYellow!25}73.92\sym{***} \\
      \cellcolor{macaronYellow!25}Reflexion (DeepSeek-R1: 14B) & \cellcolor{macaronYellow!25}86.20\sym{***} & \cellcolor{macaronYellow!25}88.96\sym{**} & \cellcolor{macaronYellow!25}81.36\sym{***} & \cellcolor{macaronYellow!25}52.66\sym{***} & \cellcolor{macaronYellow!25}77.30\sym{***} \\
      \\[-0.8em]
      \cellcolor{macaronBlue!25}Self-ensemble (Qwen2.5: 14B) & \cellcolor{macaronBlue!25}81.20\sym{***} & \cellcolor{macaronBlue!25}75.44\sym{***} & \cellcolor{macaronBlue!25}82.60\sym{***} & \cellcolor{macaronBlue!25}42.98\sym{***} & \cellcolor{macaronBlue!25}70.56\sym{***} \\
      \cellcolor{macaronBlue!25}Self-ensemble (Phi4: 14B) & \cellcolor{macaronBlue!25}84.76\sym{***} & \cellcolor{macaronBlue!25}79.04\sym{***} & \cellcolor{macaronBlue!25}82.34\sym{***} & \cellcolor{macaronBlue!25}\underline{73.74}\sym{**} & \cellcolor{macaronBlue!25}79.97\sym{***} \\
      \cellcolor{macaronBlue!25}Self-ensemble (DeepSeek-R1: 14B) & \cellcolor{macaronBlue!25}92.00\sym{***} & \cellcolor{macaronBlue!25}\underline{89.88} & \cellcolor{macaronBlue!25}85.30\sym{***} & \cellcolor{macaronBlue!25}59.32\sym{***} & \cellcolor{macaronBlue!25}81.62\sym{***} \\
      \\[-0.8em]
      \cellcolor{macaronPurple!25}Heterogeneous Ensemble & \cellcolor{macaronPurple!25}\underline{94.34}\sym{**} & \cellcolor{macaronPurple!25}84.00\sym{***} & \cellcolor{macaronPurple!25}\underline{87.72}\sym{*} & \cellcolor{macaronPurple!25}71.26\sym{***} & \cellcolor{macaronPurple!25}\underline{84.33}\sym{***} \\
      \\[-0.8em]
      \cellcolor{macaronOrange}\textbf{WILC} & \cellcolor{macaronOrange}\textbf{95.48} & \cellcolor{macaronOrange}\textbf{89.98} & \cellcolor{macaronOrange}\textbf{88.92} & \cellcolor{macaronOrange}\textbf{76.72} & \cellcolor{macaronOrange}\textbf{87.78} \\
      \bottomrule
    \end{tabular}
    \label{tab:main_results_14b}
  \fignote{The coordinator is selected following the procedure described in Section~\ref{sec:framework-formulation}; accordingly, DeepSeek-R1: 14B serves as the coordinator for HumanEval, MATH-500, and MMLU, while Qwen2.5: 14B coordinates VisEval. Significance levels $^{*}p<0.1$, $^{**}p<0.05$, $^{***}p<0.01$ are from paired $t$-tests comparing each baseline with WILC across five runs. The best result in each column is shown in \textbf{bold} and the second best is \underline{underlined}. AVG. is the unweighted average computed from unrounded values.}
  
\end{table}

\begin{table}[h!]
    \caption{Performance Comparison on Four Benchmarks: WILC vs. Baselines (30B Scale)}
    \centering
    \small
    \setlength{\tabcolsep}{10.5pt}
    \begin{tabular}{l|lllll}
      \toprule
      \textbf{Method} & \textbf{HumanEval} & \textbf{MATH-500} & \textbf{MMLU} & \textbf{VisEval} & \textbf{AVG.} \\
      \midrule
      \cellcolor{macaronPink!25}Single Execution (Qwen2.5: 32B) & \cellcolor{macaronPink!25}86.70\sym{***} & \cellcolor{macaronPink!25}71.52\sym{***} & \cellcolor{macaronPink!25}84.08\sym{***} & \cellcolor{macaronPink!25}72.14\sym{***} & \cellcolor{macaronPink!25}78.60\sym{***} \\
      \cellcolor{macaronPink!25}Single Execution (GLM4: 32B) & \cellcolor{macaronPink!25}85.00\sym{***} & \cellcolor{macaronPink!25}76.72\sym{***} & \cellcolor{macaronPink!25}84.22\sym{***} & \cellcolor{macaronPink!25}70.58\sym{***} & \cellcolor{macaronPink!25}79.10\sym{***} \\
      \cellcolor{macaronPink!25}Single Execution (Gemma2: 27B) & \cellcolor{macaronPink!25}75.20\sym{***} & \cellcolor{macaronPink!25}50.36\sym{***} & \cellcolor{macaronPink!25}74.50\sym{***} & \cellcolor{macaronPink!25}64.44\sym{***} & \cellcolor{macaronPink!25}66.10\sym{***} \\
      \\[-0.8em]
      \cellcolor{macaronGreen!25}ReAct (Qwen2.5: 32B) & \cellcolor{macaronGreen!25}88.20\sym{**} & \cellcolor{macaronGreen!25}73.20\sym{***} & \cellcolor{macaronGreen!25}83.90\sym{***} & \cellcolor{macaronGreen!25}73.70\sym{***} & \cellcolor{macaronGreen!25}79.80\sym{***} \\
      \cellcolor{macaronGreen!25}ReAct (GLM4: 32B) & \cellcolor{macaronGreen!25}\textbf{92.10} & \cellcolor{macaronGreen!25}83.96\sym{***} & \cellcolor{macaronGreen!25}86.08\sym{***} & \cellcolor{macaronGreen!25}73.38\sym{***} & \cellcolor{macaronGreen!25}83.88\sym{***} \\
      \cellcolor{macaronGreen!25}ReAct (Gemma2: 27B) & \cellcolor{macaronGreen!25}74.04\sym{***} & \cellcolor{macaronGreen!25}52.36\sym{***} & \cellcolor{macaronGreen!25}78.20\sym{***} & \cellcolor{macaronGreen!25}57.04\sym{***} & \cellcolor{macaronGreen!25}65.41\sym{***} \\
      \\[-0.8em]
      \cellcolor{macaronYellow!25}Reflexion (Qwen2.5: 32B) & \cellcolor{macaronYellow!25}87.20\sym{***} & \cellcolor{macaronYellow!25}72.64\sym{***} & \cellcolor{macaronYellow!25}84.70\sym{***} & \cellcolor{macaronYellow!25}73.60\sym{***} & \cellcolor{macaronYellow!25}79.50\sym{***} \\
      \cellcolor{macaronYellow!25}Reflexion (GLM4: 32B) & \cellcolor{macaronYellow!25}86.20\sym{***} & \cellcolor{macaronYellow!25}80.60\sym{***} & \cellcolor{macaronYellow!25}86.88\sym{**} & \cellcolor{macaronYellow!25}72.60\sym{***} & \cellcolor{macaronYellow!25}82.47\sym{***} \\
      \cellcolor{macaronYellow!25}Reflexion (Gemma2: 27B) & \cellcolor{macaronYellow!25}73.78\sym{***} & \cellcolor{macaronYellow!25}51.00\sym{***} & \cellcolor{macaronYellow!25}72.08\sym{***} & \cellcolor{macaronYellow!25}63.20\sym{***} & \cellcolor{macaronYellow!25}65.02\sym{***} \\
      \\[-0.8em]
      \cellcolor{macaronBlue!25}Self-ensemble (Qwen2.5: 32B) & \cellcolor{macaronBlue!25}80.86\sym{***} & \cellcolor{macaronBlue!25}67.50\sym{***} & \cellcolor{macaronBlue!25}85.42\sym{***} & \cellcolor{macaronBlue!25}77.34\sym{***} & \cellcolor{macaronBlue!25}77.80\sym{***} \\
      \cellcolor{macaronBlue!25}Self-ensemble (GLM4: 32B) & \cellcolor{macaronBlue!25}84.66\sym{***} & \cellcolor{macaronBlue!25}68.92\sym{***} & \cellcolor{macaronBlue!25}86.60\sym{**} & \cellcolor{macaronBlue!25}\underline{82.28}\sym{**} & \cellcolor{macaronBlue!25}80.60\sym{***} \\
      \cellcolor{macaronBlue!25}Self-ensemble (Gemma2: 27B) & \cellcolor{macaronBlue!25}75.60\sym{***} & \cellcolor{macaronBlue!25}66.08\sym{***} & \cellcolor{macaronBlue!25}77.62\sym{***} & \cellcolor{macaronBlue!25}59.20\sym{***} & \cellcolor{macaronBlue!25}69.60\sym{***} \\
      \\[-0.8em]
      \cellcolor{macaronPurple!25}Heterogeneous Ensemble & \cellcolor{macaronPurple!25}88.86\sym{**} & \cellcolor{macaronPurple!25}\underline{87.20}\sym{*} & \cellcolor{macaronPurple!25}\underline{87.10}\sym{***} & \cellcolor{macaronPurple!25}77.00\sym{***} & \cellcolor{macaronPurple!25}\underline{85.04}\sym{***} \\
      \\[-0.8em]
      \cellcolor{macaronOrange}\textbf{WILC} & \cellcolor{macaronOrange}\underline{91.12} & \cellcolor{macaronOrange}\textbf{88.56} & \cellcolor{macaronOrange}\textbf{87.72} & \cellcolor{macaronOrange}\textbf{84.06} & \cellcolor{macaronOrange}\textbf{87.87} \\
      \bottomrule
    \end{tabular}
    \label{tab:main_results_30b}
  \tabnote{Notation conventions follow Table~\ref{tab:main_results_14b}. The coordinator is selected following the procedure in Section~\ref{sec:framework-formulation}; accordingly, Qwen2.5: 32B serves as the coordinator for HumanEval and VisEval, while GLM4: 32B coordinates MATH-500 and MMLU.}

\end{table}

\begin{figure*}[t]
  \centering
  \caption{Performance Gains from Wisdom of Crowds: WILC vs. Single Model Execution}
  \begin{subfigure}[b]{0.4\textwidth}
      \centering
      \includegraphics[width=\linewidth]{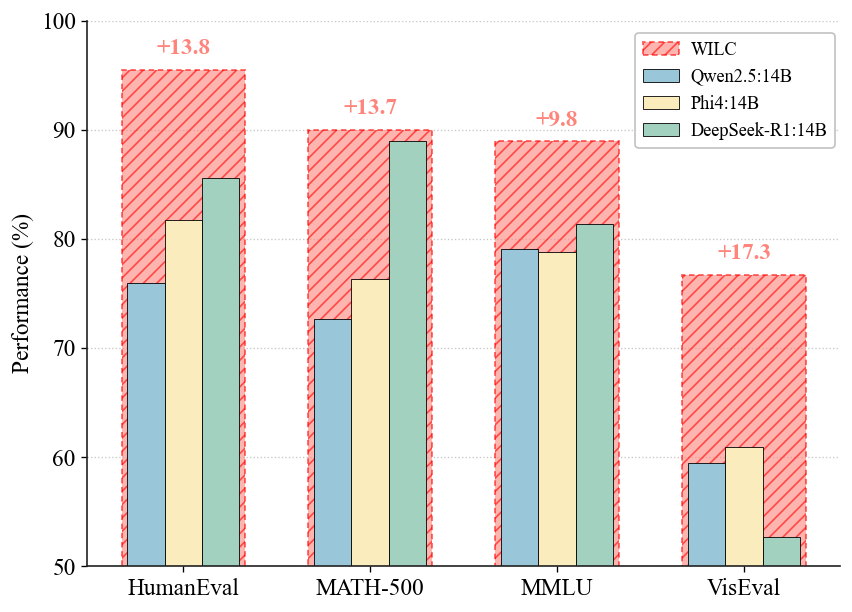}
      \caption{Model Scale: 14B}
      \label{fig:single-vs-wilc-14b}
  \end{subfigure} 
  \quad
  \begin{subfigure}[b]{0.4\textwidth}
      \centering
      \includegraphics[width=\linewidth]{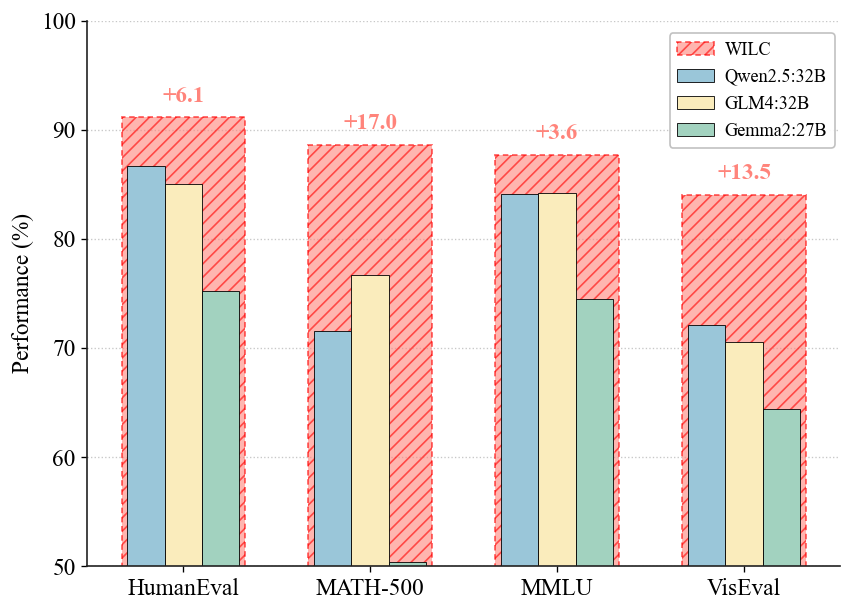}
      \caption{Model Scale: 30B}
      \label{fig:single-vs-wilc-30b}
  \end{subfigure}
  \label{fig:single-vs-wilc}
  \fignote{The increases depicted in the figure indicate the absolute difference between WILC and the median performance of individual model executions. This visualizes the incremental value added by the wisdom of crowds over individual model execution.}
  \vspace{-4pt}
\end{figure*}

To more clearly illustrate the performance gains achieved by the wisdom of LLM crowds, we compare the performance of WILC against the median single-execution performance of worker models within an LLM crowd. As shown in Figure~\ref{fig:single-vs-wilc}, this comparison provides an intuitive view of the improvements attributable to the wisdom of LLM crowds. We choose the median\footnote{Using the mean as the reference point leads to the same conclusion.} as the reference point rather than the mean or maximum because it provides a more robust baseline that is less sensitive to outliers. More importantly, in real-world applications where the best-performing model for a specific query is unknown a priori, the median better reflects the performance of a typical standalone model than the maximum does. The results reveal substantial and consistent performance improvements across all benchmarks and model scales. At the 14B scale, WILC achieves absolute performance gains ranging from +9.8 to +17.3 percentage points. Similarly, at the 30B scale, WILC achieves gains of +3.6 to +17.0 percentage points, with the most substantial improvement again observed on MATH-500.

A closer examination of the figures reveals two key patterns. \textit{First}, the magnitude of improvement correlates with task complexity: reasoning-intensive tasks like MATH-500 exhibit the largest gains, while knowledge-based tasks like MMLU show more modest improvements. This suggests that the wisdom of crowds mechanism is particularly valuable for problems requiring multi-step reasoning, where different models may excel at different reasoning stages. \textit{Second}, the performance variance among individual models within each benchmark provides insight into the critical value of the LLM crowd in mitigating selection risks while improving the performance ceiling. In practice, organizations face substantial uncertainty when selecting a single model for deployment, as the optimal model for a given task is often unknown a priori. Making a suboptimal choice can lead to significant performance degradation and business losses. Consider the MATH-500 benchmark at the 30B scale as an illustrative example: Gemma2: 27B lags approximately 26 percentage points behind the best individual model GLM4: 32B in accuracy. An organization deploying such a model would suffer considerable opportunity costs due to this performance gap. However, through the wisdom of LLM crowds, WILC not only mitigates this risk by preventing any single weak model from dominating the outcome, but remarkably achieves 88.56\% accuracy and surpasses the strongest individual model. This demonstrates that WILC can strategically leverage complementary strengths from all models in the crowd, including extracting valuable capabilities from otherwise weaker models, to deliver superior collective performance that transcends individual model limitations and maximizes organizational value.

We also compare WILC against dedicated query-routing approaches.\footnote{A fair comparison with routing methods is inherently difficult, as their performance critically depends on the quality and domain coverage of the labeled training dataset used to learn the routing function~\citep{shnitzer2023large,feng2024llmrouter}. We provide a detailed discussion and comparison in Appendix~\ref{sec:routing-comparison}.} As shown in Appendix~\ref{sec:routing-comparison}, routing methods perform competitively on benchmarks well-represented in their training data (e.g., HumanEval, MATH-500) but collapse on VisEval, a task type not directly represented in the cold-start data categories. Even restricting WILC to a single round ($r_{\max}{=}1$) performs better than all routing baselines, because WILC's burn-in evaluates all workers on the actual target query rather than predicting suitability from potentially mismatched training data.
Moreover, the full WILC framework's multi-round gains persist on VisEval because its successor selection is guided by model-bottleneck fitness---bottleneck patterns that generalize across task types regardless of training data coverage. This robustness property is practically significant, as real-world query distributions rarely conform perfectly to any pre-collected training dataset.

\subsection{Ablation Study}\label{sec:ablation-study}
To rigorously assess the contributions of key components within the WILC framework, we conduct a comprehensive and systematic ablation study involving five distinct ablation settings.
\begin{itemize}
  \item \textbf{Self-reflection:} To assess whether multi-model collaboration provides advantages over single-model iterative refinement, this ablation examines the value of cross-model collaboration by comparing WILC with a setting where each worker merely reflects and refines its own solutions.\footnote{This ablation is conceptually aligned with Reflexion~\citep{shinn2023reflexion}, as both rely on a single model iteratively refining its own output through self-generated feedback.} For simplicity, we use worker 1--3 to represent the three workers in each crowd.
  \item \textbf{Random Worker Selection:} To assess whether our capability-aware worker selection is more effective than random assignment, this ablation replaces the PCF gate with a random draw from the worker pool while retaining all other components.
  \item \textbf{Without Burn-In Strategy:} To assess the necessity of the burn-in initialization strategy for the framework, this ablation evaluates the strategy's importance by comparing WILC with the setting that removes the burn-in strategy, which instead uses LinUCB to select the worker at the first round.
  \item \textbf{Without Multi-Round Collaboration:} To assess the value of iterative multi-round refinement, this ablation restricts WILC to a single round (i.e., only the burn-in round), effectively reducing it to a one-shot model selection mechanism that selects the best worker (evaluated by proxy performance during the burn-in round) but forgoes any subsequent relay-style collaboration.
  \item \textbf{Without OSFS:} To assess the necessity of the one-step forward search (OSFS) for the proposed framework, this ablation study evaluates the setting in which the search is removed.
\end{itemize}

\begin{figure*}[!t]
  \centering
  \caption{Ablation Study Results on Four Benchmarks}
  \includegraphics[width=0.9\textwidth]{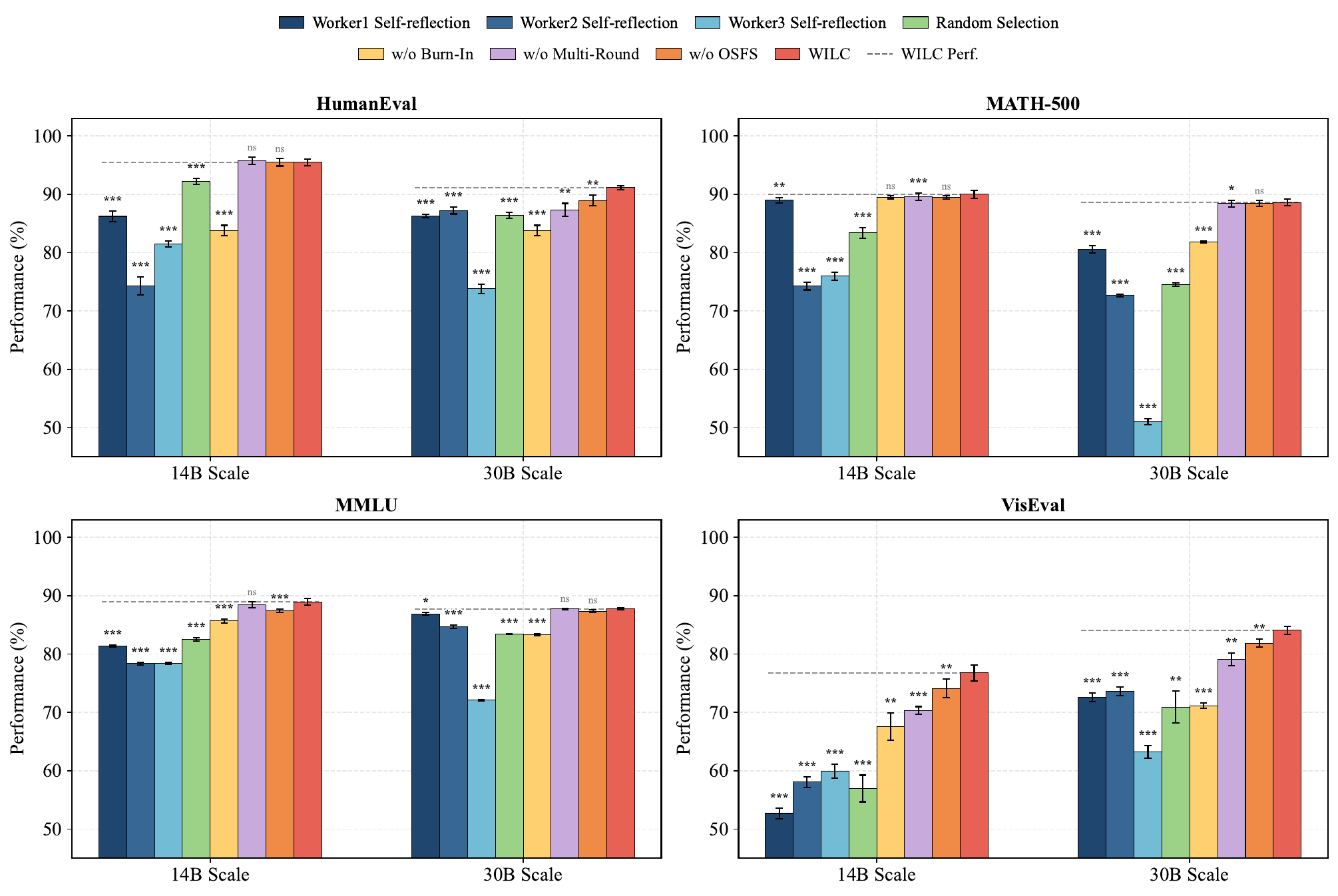}
  \fignote{Results averaged over five runs with standard errors reported. Significance levels $^{*}p<0.1$, $^{**}p<0.05$, $^{***}p<0.01$ (ns: not significant) are from paired $t$-tests against WILC. Pairing removes the run-to-run variation shared by both settings, so the markers reflect how consistently WILC leads, not by how much: on MATH-500 (14B scale), the two settings vary together across runs (run-to-run SD $\approx$ 1.5 points), yet WILC leads in all five runs by a stable margin, giving $p<0.01$ for an average gap of only 0.44 pp.
  }
  \label{fig:ablation_results}

\end{figure*}

Figure \ref{fig:ablation_results} presents the results of the ablation studies across four benchmarks, revealing several key insights into the contributions of different components within WILC. \textit{First}, WILC performs at least as well as every ablation setting in general, either significantly better or statistically comparable (the only settings without a significant advantage are those where WILC has already saturated, such as HumanEval at the 14B scale, where a single round already reaches 95.5\%), demonstrating the synergistic value of integrating all proposed components.
\textit{Second}, the self-reflection ablation highlights that WILC, by coordinating complementary capabilities across the crowd, consistently outperforms any individual model relying solely on self-refinement. This advantage holds without exception: WILC leads in all 24 comparisons (3 workers $\times$ 4 benchmarks $\times$ 2 scales) by margins of $+0.84$ to $+37.56$ percentage points, indicating that its gains cannot be attributed to iterative refinement alone. Moreover, WILC demonstrates robustness even when a weak model exists in the LLM crowd by skillfully integrating the diverse strengths of different models while mitigating individual shortcomings.
\textit{Third}, comparing WILC against the other four ablation settings (i.e., random selection, w/o burn-in, w/o multi-round, and w/o OSFS) reveals clear differences in component contributions. Because random selection retains every other component and replaces only the PCF gate with a random draw, its comparison with WILC isolates the contribution of capability-aware successor selection. This comparison yields a significant improvement in all eight benchmark--scale settings ($+3.28$ to $+19.80$ percentage points). Notably, access to a heterogeneous crowd alone is not sufficient: in six of the eight settings, random selection performs below the strongest single model under self-reflection, as an uninformed handover may pass the task to a worker ill-suited to the current bottleneck and degrade an already adequate solution. Complementarity is therefore latent in a heterogeneous crowd; realizing it depends on directing handovers toward workers whose estimated capabilities fit the diagnosed bottleneck. In our design, PCF-based selection supplies this direction, while PCG guards against proxy-assessed degradation after the handover. The w/o multi-round ablation shows that the value of additional rounds is task-dependent: it contributes 3.8 to 6.4 percentage points on the open-ended generation tasks (VisEval at both scales and HumanEval~30B), whereas on MMLU and MATH-500 the burn-in round already resolves most queries and further rounds add little. Meanwhile, the ablation settings w/o burn-in and w/o OSFS demonstrate moderate performance degradation compared to WILC, confirming that these robustness-oriented design elements contribute to enhancing the framework's stability and reliability across diverse task scenarios.

\subsection{How Capability Complementarity Emerges and Operates?}
To better understand how capability complementarity contributes to WILC's performance, we examine it from two perspectives. We first analyze the heterogeneity of worker capabilities as the basis for complementarity, and then investigate the dynamic collaboration patterns through which such complementarity is realized in the multi-round problem-solving process.
\subsubsection{Capability Heterogeneity Across Workers}
\begin{figure*}[!t]
  \centering
  \caption{Worker Heterogeneity Across Query Clusters}
  \includegraphics[width=\textwidth]{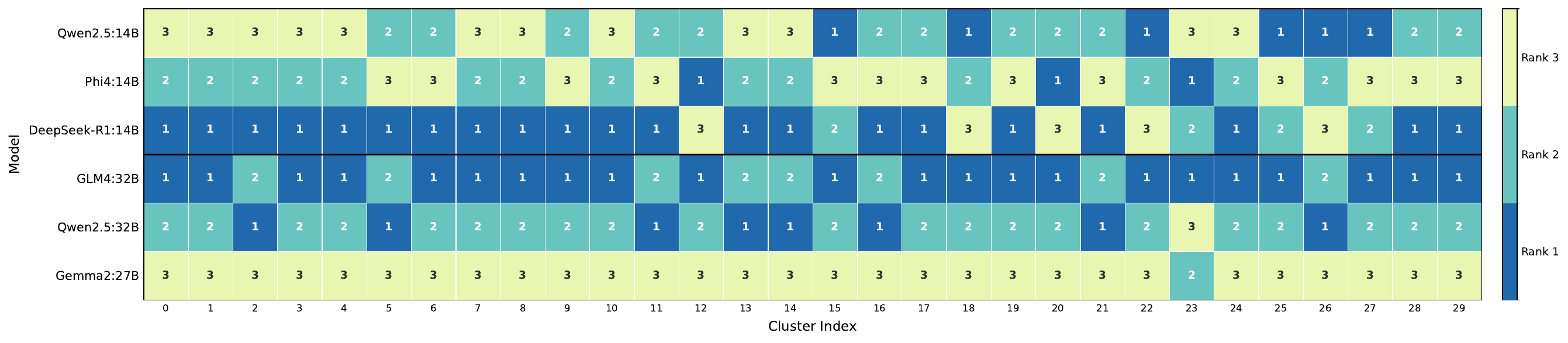}
  \label{fig:capability-spectrum}
  \vspace{-4pt}
\end{figure*}

The effectiveness of WILC depends on heterogeneous capabilities across workers, such that one worker may be better able to resolve a problem-solving bottleneck on which another worker may underperform.
To analyze this heterogeneity, we visualize rank-based capability distribution of each worker over the cold-start query set $\mathcal{Q}$. 
Specifically, we first encode each query in $\mathcal{Q}$ into its embedding representation (as introduced in Section~\ref{sec:PCF}) and then apply K-Means clustering to partition the query set into 30 clusters.\footnote{The value of $30$ is chosen for illustrative purposes; using other reasonable numbers of clusters should lead to similar conclusions.} Queries within the same cluster tend to share similar semantic and bottleneck structures, whereas queries across clusters are more distinguishable.
Next, for each worker listed in Table~\ref{tab:llms}, we compute its average accuracy within each cluster and convert these values into within-scale rankings. That is, for each cluster, workers are ranked relative to other workers in the same model scale, with Rank~1 indicating the strongest performer in that cluster. Figure~\ref{fig:capability-spectrum} visualizes the resulting ranking matrix, with cluster index on the x-axis and worker on the y-axis. Compared with absolute accuracy values, this rank-based representation more directly highlights the comparative advantage structure that is most relevant to worker selection in WILC.

Figure~\ref{fig:capability-spectrum} reveals substantial cross-worker variation in relative rankings across clusters, indicating pronounced heterogeneity in capability profiles and suggesting meaningful opportunities for complementary collaboration. At the 14B scale, DeepSeek-R1:14B ranks first in most clusters, yet Qwen2.5:14B and Phi4:14B still attain the top rank in moderate number of clusters. At the 30B scale, the top two ranks are primarily occupied by GLM4:32B and Qwen2.5:32B across different clusters. Meanwhile, in our tested 30B pool, the weaker performance of Gemma2:27B suggests that the presence of a relatively weak worker does not prevent WILC from improving overall performance. Taken together, these patterns suggest that capability complementarity in LLM crowds is not characterized by uniform superiority of a single worker, but rather by shifting relative advantages across different regions of the query space. Such heterogeneous ranking patterns provide the foundation for WILC to translate capability heterogeneity into improved collective performance through adaptive worker selection.

\subsubsection{Collaboration Dynamics in Multi-Round Problem Solving}

We analyze collaboration dynamics by examining how workers alternate across rounds. For an instance that terminates after $r$ rounds, let the worker selection sequence be $w_1 \rightarrow w_2 \rightarrow \cdots \rightarrow w_r$, where $w_i$ denotes the worker selected at round $i$. We define a \emph{switch} as a change in worker between two consecutive rounds, i.e., $w_i \neq w_{i+1}$. We then quantify such dynamics using the \emph{Switch Rate}:
\begin{equation}
\textit{Switch Rate} = \frac{\#\ \textit{switches}}{r-1},
\end{equation}
where $r$ is the total number of rounds at termination, and $\#\ \textit{switches}$ refers to the number of switches. For instance, if a dialogue terminates at round $4$ with the sequence ``Qwen2.5: 32B $\rightarrow$ GLM4: 32B $\rightarrow$ Qwen2.5: 32B $\rightarrow$ Qwen2.5: 32B'', then two switches occur across three possible transitions, yielding a switch rate of $2/3$ (i.e., 0.67). A higher switch rate typically indicates more frequent activation of different workers to facilitate complementary collaboration, whereas a lower switch rate suggests preference and reliance on a particular worker.

\begin{figure*}[!t]
  \centering
  \caption{Switch Rate Distribution on Four Benchmarks}
  \includegraphics[width=1\textwidth]{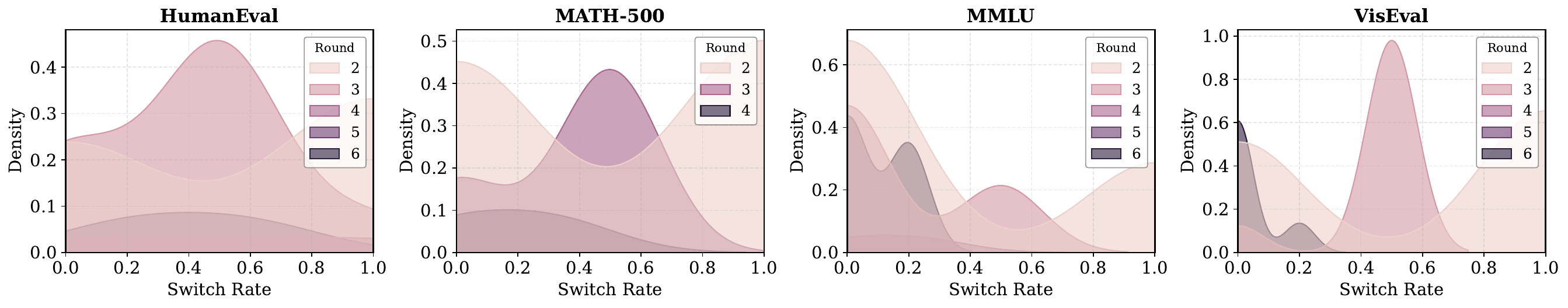}
  \label{fig:switch_rate_distribution}
  \vspace{-4pt}
\end{figure*}

Using the 30B-scale crowd as an example, Figure~\ref{fig:switch_rate_distribution} shows the distribution of switch rates on four benchmarks. The plots show that WILC does not consistently adhere to a single worker throughout the entire multi-round reasoning process. Instead, across all benchmarks, we observe substantial switching, with many dialogues exhibiting switch rates concentrated between 0.2 and 0.8. This pattern suggests that worker alternation is common and that collaboration manifests in a dynamically diverse manner. Importantly, this phenomenon is consistent with WILC's worker selection mechanism, which can perceive and capture the capability demands induced by different self-reflection rounds and proactively elicit contributions from workers whose strengths align with the evolving needs of the task. Such dynamic switching more fully exploits workers' differentiated advantages across rounds, thereby improving the overall performance of the LLM crowds. 

\begin{figure*}[!t]
    \centering
    \caption{Illustrative Multi-Round Collaboration Topologies within Three Rounds}
    \begin{subfigure}{0.35\textwidth}
      \centering
      \includegraphics[width=\linewidth]{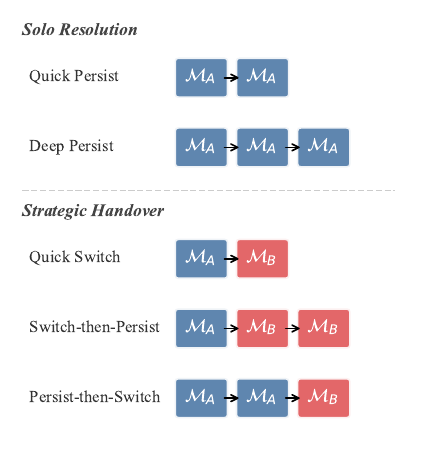}
      \caption{Topology Taxonomy} 
      \label{fig:topo1}
    \end{subfigure}
    \begin{subfigure}{0.5\textwidth}
      \centering
      \includegraphics[width=\linewidth]{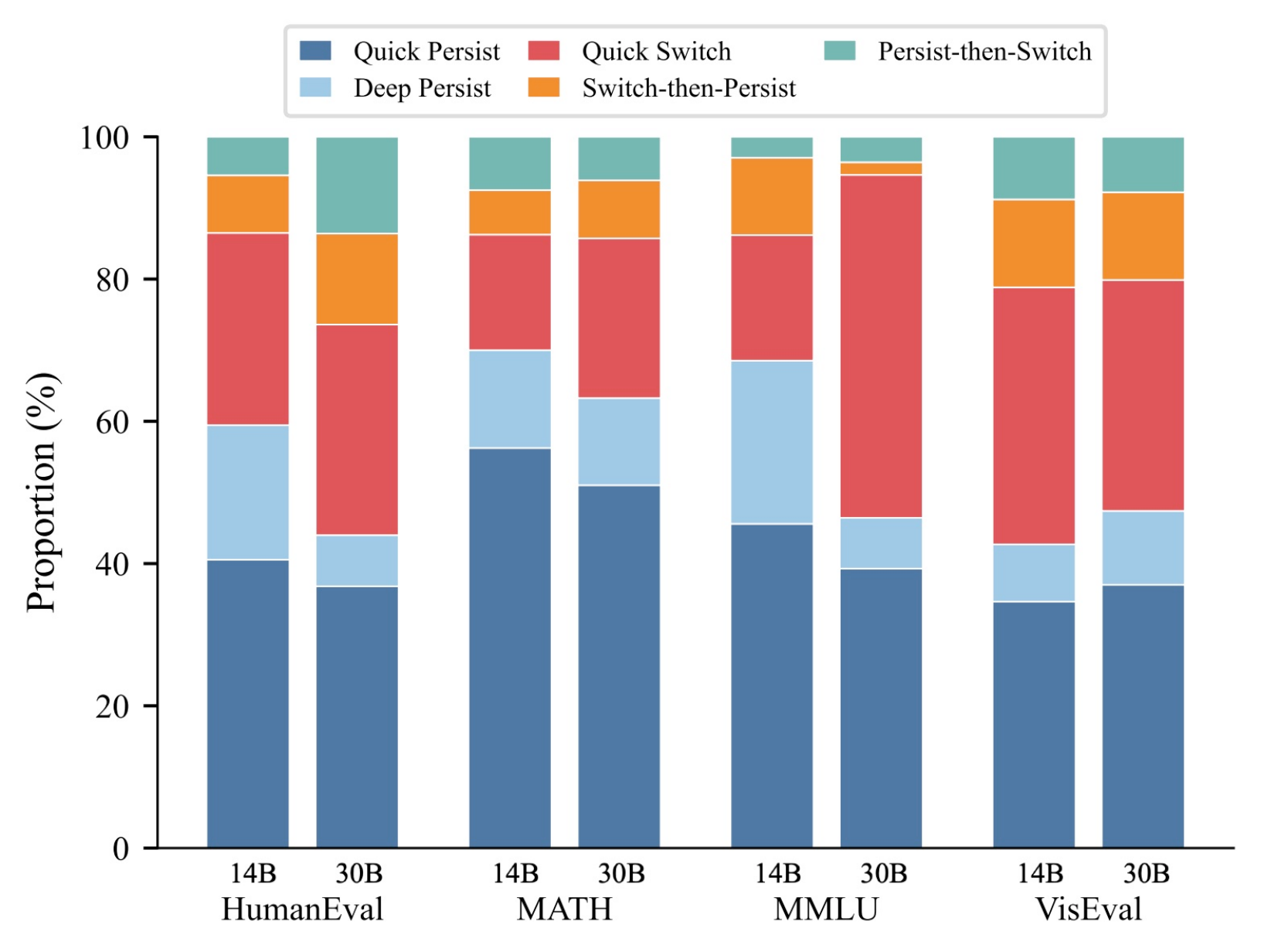}
      \caption{Topology Breakdown on Four Benchmarks} 
      \label{fig:topo2}
    \end{subfigure}
    \label{fig:collaboration_topology}
    \vspace{-4pt}
  \end{figure*}

To understand how the framework orchestrates multi-model collaboration, we examine the collaboration topologies that emerge within the first three rounds of multi-round problem-solving processes. As shown in Fig.~\ref{fig:collaboration_topology}, we identify five prototypical topologies grouped into two categories and show the distribution of these topologies across four benchmarks at two model scales. Several findings stand out. First, strategic handover accounts for a substantial share of multi-round processes across all settings, ranging from 30.0\% on MATH-14B to 57.3\% on VisEval-14B, confirming that WILC actively leverages the heterogeneity of the worker pool. Second, the topology profile varies meaningfully with task nature: VisEval and HumanEval, which involve open-ended generation, exhibit the highest handover rates (52.6--57.3\%), whereas MATH, which demands sustained chain-of-thought reasoning, favors solo resolution (63.3--70.0\%). This suggests that the topology profile is not arbitrary but rather shaped by the underlying task characteristics, indicating that WILC implicitly adapts its collaboration strategy to the nature of the problem at hand. Third, within strategic handover, quick switch is consistently the dominant subtype, indicating that the framework efficiently recognizes first-round failures and promptly redirects to a more suitable model.

\subsection{Robustness Checks}\label{sec:robustness-checks}
\subsubsection{Selection of Coordinator}
As described in Section~\ref{sec:framework-formulation}, we designate the coordinator as the worker that most frequently achieves the highest proxy reward during the burn-in phase across an initial batch of target-task queries. A natural follow-up question is whether WILC's performance is sensitive to this choice. To test this sensitivity, we compare WILC's performance at the 14B scale when the coordinator is set to the strongest versus the weakest model in the worker pool. Table~\ref{tab:coordinator_robustness} reports the results averaged over five runs. Despite the substantial gap in the coordinators' standalone capabilities, the performance differences between the two configurations are small and statistically insignificant across all four benchmarks ($p > 0.05$). On MMLU, the weakest coordinator even yields a marginally higher average score than the strongest. Overall, the mean performance gap is only 0.66 percentage points (87.78\% vs.\ 87.12\%).

This robustness stems from two aspects. \textit{First}, the coordinator's role in WILC is \textit{diagnostic} rather than \textit{generative}---it identifies bottlenecks in the current solution (e.g., logical gaps, incorrect computations, or missing steps) rather than producing the solution itself. This task demands reflective capability, which is a qualitatively different skill from problem solving. Even a model that is relatively weaker at generating correct answers can still adequately recognize and articulate the issues present in another model's output, because critiquing an existing solution is cognitively less demanding than producing a correct one from scratch. \textit{Second}, the primary driver of WILC's collaborative performance, namely the adaptive LinUCB-based worker selection mechanism, operates independently of the coordinator's identity. The PCF gate selects workers based on learned capability--bottleneck fitness, and the PCG gate validates improvement via proxy rewards; both mechanisms depend on the \textit{content} of the reflection (i.e., the number and nature of identified issues) rather than on which model produced it. Consequently, as long as the coordinator can surface salient issues with reasonable fidelity, the downstream selection and verification pipeline functions effectively regardless of the coordinator's standalone problem-solving capability.

This insensitivity also implies that extending WILC to a \textit{dynamic} coordinator selection, where the coordinator is re-designated per query based on real-time burn-in feedback, is unlikely to yield significant additional gains, since even a suboptimal coordinator produces sufficiently reliable reflections. Nonetheless, the burn-in mechanism already provides the infrastructure for such an extension if needed in future applications.
\begin{table}[h!]
    \caption{Coordinator Robustness Analysis (14B Scale)}
    \centering
    \small
    \setlength{\tabcolsep}{15.5pt}
    \begin{tabular}{l|cccc|c}
      \toprule
      \textbf{Coordinator} & \textbf{HumanEval} & \textbf{MATH-500} & \textbf{MMLU} & \textbf{VisEval} & \textbf{AVG.} \\
      \midrule
      Strongest & 95.48 $\pm$ 1.29 & 89.98 $\pm$ 1.45 & 88.92 $\pm$ 1.26 & 76.72 $\pm$ 3.09 & 87.78 \\
      Weakest  & 92.92 $\pm$ 2.22 & 89.24 $\pm$ 0.86 & 89.70 $\pm$ 1.65 & 76.62 $\pm$ 2.88 & 87.12 \\
      \midrule
      $\Delta$  & $+$2.56 & $+$0.74 & $-$0.78 & $+$0.10 & $+$0.64 \\
      \bottomrule
    \end{tabular}
    \label{tab:coordinator_robustness}
  \tabnote{Performance is reported as mean $\pm$ standard deviation over five random seeds. ``Strongest'' and ``Weakest'' refer to the model with the highest and lowest single-execution accuracy on each benchmark, respectively. $\Delta$ = Strongest $-$ Weakest.}
  
\end{table}

\subsubsection{Exploration Parameter}
We further examine the role of the exploration parameter $\alpha$ in the LinUCB algorithm, which governs the exploration--exploitation trade-off. A larger $\alpha$ encourages the system to explore less-tried workers, while $\alpha=0$ reduces LinUCB to pure exploitation with no exploration at all. Table~\ref{tab:alpha_robustness} reports the performance at the 14B scale for $\alpha \in \{0, 0.01, 0.05, 0.1, 0.5\}$.

Two observations emerge. First, the $\alpha=0$ row serves as an ablation that removes exploration entirely. Compared with the non-zero $\alpha$ settings, pure exploitation achieves a comparable but slightly lower overall average: 86.86\% versus 86.97\%--87.78\% when $\alpha>0$.This confirms that exploration, even in modest amounts, is beneficial, as it allows the system to discover complementary workers that pure exploitation would overlook. Second, across all non-zero $\alpha$ values spanning a 50-fold range (0.01--0.5), performance remains highly stable, with the average varying by only 0.81 percentage points (86.97\%--87.78\%). This demonstrates that WILC is robust to the specific choice of $\alpha$ and requires no elaborate tuning in practice.

\begin{table}[h!]
    \caption{Ablation and Sensitivity on the Exploration Parameter $\alpha$ (14B Scale, Single Run)}
    \centering
    \small
    \setlength{\tabcolsep}{14pt}
    \begin{tabular}{c|llll|c}
      \toprule
      $\alpha$ & \textbf{HumanEval} & \textbf{MATH-500} & \textbf{MMLU} & \textbf{VisEval} & \textbf{AVG.} \\
      \midrule
      0 (no exploration) & 95.85 $\pm$ 1.46 & 91.42 $\pm$ 1.44 & 88.71 $\pm$ 1.09 & 71.47 $\pm$ 3.46 & 86.86 \\
      \midrule
      0.01 & 95.85 $\pm$ 1.46 & 90.52 $\pm$ 0.27 & 88.29 $\pm$ 0.29 & 73.20 $\pm$ 3.23 & 86.97 \\
      0.05 & 95.85 $\pm$ 1.46 & 90.56 $\pm$ 0.33 & 88.28 $\pm$ 0.27 & 73.40 $\pm$ 3.59 & 87.02 \\
      0.1 (default)  & 95.48 $\pm$ 1.29 & 89.98 $\pm$ 1.45 & 88.92 $\pm$ 1.26 & 76.72 $\pm$ 3.09 & 87.78 \\
      0.5  & 95.85 $\pm$ 1.46 & 90.60 $\pm$ 0.32 & 88.28 $\pm$ 0.27 & 73.47 $\pm$ 3.86 & 87.05 \\
      \bottomrule
    \end{tabular}
    \label{tab:alpha_robustness}
\end{table}

\subsection{Additional Analyses}
\subsubsection{Value of Learning Model-Bottleneck Fitness}
\label{subsubsec:value-of-learning-model-bottleneck-fitness}
As articulated in DP2 (see Section~\ref{sec:design-points}), the effectiveness of relay-style collaboration depends on selecting a successor worker whose capabilities specifically complement the bottleneck identified in the current problem-solving state. Accordingly, WILC relies on learning \textit{model-bottleneck fitness}, rather than generic query routing based solely on the initial query or broad task category. Specifically, this section examines whether incorporating bottleneck-contextualized queries (i.e., queries containing the current bottleneck to be resolved) into the cold-start phase improves the learning of model-bottleneck fitness.

As described in Section~\ref{sec:PCF}, the cold-start dataset comprises two types of queries: (a) initial queries and (b) bottleneck-contextualized queries. We posit that the capability to resolve bottlenecks (e.g., rectifying logical inconsistencies or debugging code errors) is largely a scenario-agnostic trait inherent to a model, rather than being strictly tied to a particular problem domain. In other words, a specific type of bottleneck, such as a deficiency in long-chain reasoning, may exhibit shared structural patterns across diverse scenarios. Consequently, incorporating bottleneck-contextualized queries in the cold-start stage allows the LinUCB model to learn more generalizable model-bottleneck fitness patterns that go beyond surface-level domain features, thereby enabling the system to better identify the worker most capable of addressing a bottleneck even in new and previously unseen scenarios.

\begin{figure*}[!t]
  \centering
  \caption{Comparison of Performance with and without Bottleneck-Contextualized Queries}
  \begin{subfigure}[b]{0.48\textwidth}
      \centering
      \includegraphics[width=\linewidth]{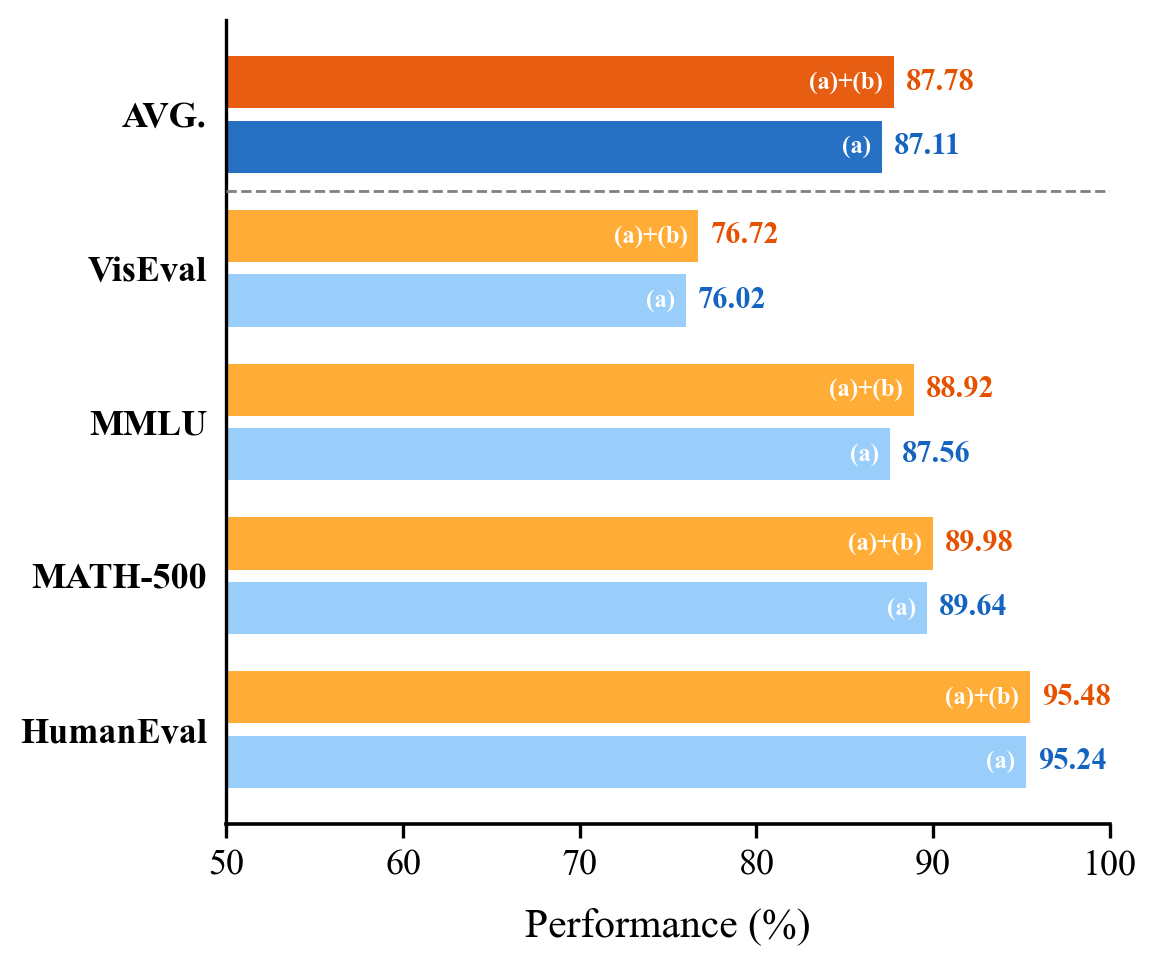}
      \caption{Model Scale: 14B}
      \label{fig:ablation-bottleneck-14b}
  \end{subfigure} 
  \quad
  \begin{subfigure}[b]{0.48\textwidth}
      \centering
      \includegraphics[width=\linewidth]{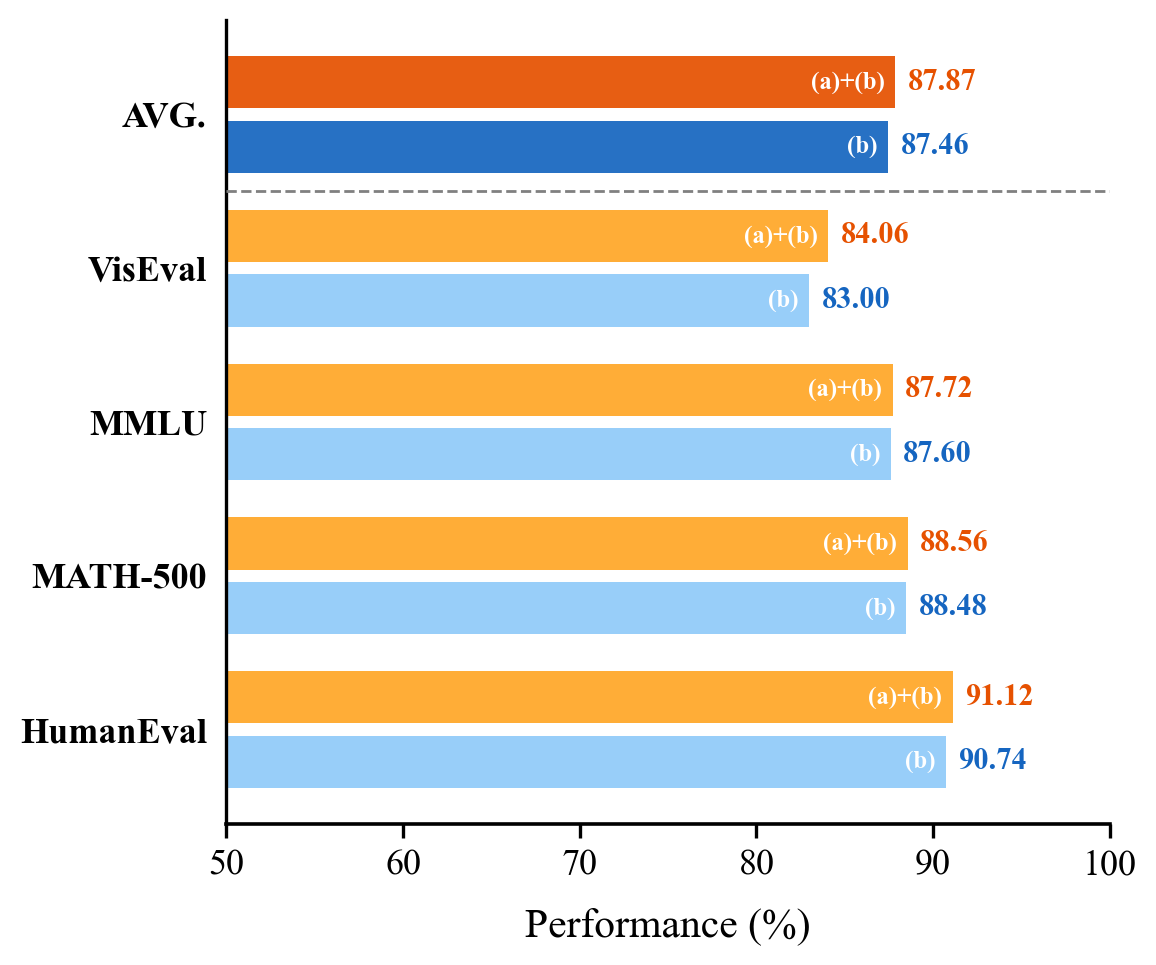}
      \caption{Model Scale: 30B}
      \label{fig:ablation-bottleneck-30b}
  \end{subfigure}
  \label{fig:ablation-bottleneck}
  
\end{figure*}

To empirically evaluate this idea, we compare two cold-start configurations for training the LinUCB model to learn model-bottleneck fitness: one using only type (a) queries, and the other using both type (a) and type (b) queries. Figure~\ref{fig:ablation-bottleneck} reports the performance of WILC under these two configurations.
The results show that incorporating bottleneck-contextualized queries into the cold-start phase yields modest but generally consistent improvements in final performance across both model scales. Taken together, these findings provide empirical support for the value of learning model-bottleneck fitness. Training only on initial queries yields a basic understanding of each model's general strengths, but this is insufficient for a system designed for iterative refinement. In relay-style collaboration, the system repeatedly encounters intermediate answers with specific bottlenecks that require targeted correction. By incorporating bottleneck-contextualized queries into the cold-start phase, the LinUCB-based selection mechanism develops a more nuanced and generalizable understanding of worker capabilities. It learns not merely which model performs better on a broad task type, but more importantly which model is best suited to resolve the bottleneck identified in a preceding worker's failure. This shift from coarse query routing to selection guided by learned model-bottleneck fitness appears to be an important enabler of the deep capability complementarity achieved by WILC.

\subsubsection{Cost-Effectiveness Analysis}\label{sec:cost-analysis}

A natural concern is whether the additional overhead from multi-round, multi-model collaboration is justified by the performance gains it delivers. To address this, we estimate WILC's per-query cost based on its API call structure, empirical round counts, and per-token pricing, then benchmark the cost-performance tradeoff against frontier closed-source models.

Taking the case of $K=3$ workers as an example, Table~\ref{tab:api_wilc} breaks down the API calls per round. Although the theoretical maximum over $r_{\max}=6$ rounds is 21 calls, the PCG gate adapts collaboration depth to problem difficulty: easy queries are resolved after the burn-in round alone, effectively reducing WILC to a one-shot model selection, while difficult queries proceed through additional refinement rounds. The empirical averages are 1.78 rounds (14B) and 1.46 rounds (30B) across our benchmarks, corresponding to roughly 8.3 and 7.4 API calls per query, respectively (see Appendix~\ref{apx:cost-analysis} for details).
This adaptive depth control means that for simpler queries where the PCG gate terminates after the burn-in round, WILC effectively operates as a model selection mechanism. The additional API calls are invested only in genuinely difficult queries where iterative refinement yields meaningful gains, making the multi-round overhead both targeted and well-justified.

\begin{table}[t]
\centering
\caption{API Call Breakdown for WILC Framework ($K{=}3$ Workers, $r_{\max}{=}6$ Rounds)}
\label{tab:api_wilc}
\begin{threeparttable}
\begin{tabular*}{0.8\textwidth}{@{\extracolsep{\fill}}cccccc@{}}
\toprule
 & \multicolumn{3}{c}{\textbf{Component}} & \multicolumn{2}{c}{\textbf{Total}} \\
\cmidrule(lr){2-4} \cmidrule(lr){5-6}
\textbf{Round} & \textbf{Burn-in} & \textbf{OSFS$^\dagger$} & \textbf{Relay} & \textbf{Round} & \textbf{Cumulative} \\
\midrule
1 & 6 & 0 & 0 & 6 & 6 \\
2 & 0 & 1 & 2 & 3 & 9 \\
3 & 0 & 1 & 2 & 3 & 12 \\
4 & 0 & 1 & 2 & 3 & 15 \\
5 & 0 & 1 & 2 & 3 & 18 \\
6 & 0 & 1 & 2 & 3 & 21 \\
\bottomrule
\end{tabular*}
\tabnote{Each component count includes both a worker call and a coordinator call. $^\dagger$OSFS is triggered only when the selected worker differs from the previous round. Assuming a 50\% switching probability, the expected number of OSFS calls per round is 1 (0 when no switch occurs, or 2 when a switch occurs).}
\end{threeparttable}
\end{table}

To quantify cost effectiveness, we evaluate GPT-4o, GPT-5.2, and GPT-5.4 on the same four benchmarks used in our main experiments and compare their average performance and estimated per-query cost against WILC (see Table~\ref{tab:cost_comparison} in the Appendix for the full comparison). Using public API pricing as a common yardstick, a single GPT-5.2 query costs \$0.0088, whereas WILC at the 30B scale costs only $\sim$\$0.0013 per query, about 7$\times$ cheaper, while achieving a comparable average accuracy (87.86\% vs.\ 88.09\%). At the 14B scale the gap is similarly striking: WILC-14B achieves 87.78\%, nearly matching GPT-5.2, at roughly 15\% of the per-query cost. In practice, the cost advantage can be even more pronounced: our experiments use locally served models via \texttt{llama.cpp} with 4-bit quantization, where inference cost is governed by fixed hardware expenditure rather than per-token charges. Under self-hosted deployment, the per-token API charge is eliminated entirely, and the amortized per-query cost decreases with throughput, though total cost of ownership depends on hardware utilization and infrastructure overhead. Meanwhile, self-hosted deployment ensures that no data leaves the organization, satisfying the privacy and compliance requirements that motivate private LLM deployments in the first place.

\section{Conclusion}\label{sec:Conclusion}

This research addresses a fundamental challenge in leveraging the wisdom of LLM crowds: how to enable heterogeneous LLMs to collaboratively solve complex problems through strategic model coordination that achieves deep capability complementarity. Following the computational design science paradigm~\citep{hevner2004design,abbasi2024pathways,fang2025computational}, we designed and evaluated WILC, a novel IT artifact that models multi-LLM collaboration as a dynamic relay-style sequential process driven by complementarity among heterogeneous models. Our evaluation focuses on an enterprise setting in which organizations deploy multiple medium-sized open-source models from different developers under constraints related to computational budgets, data sovereignty, and customization flexibility, while the framework itself is not restricted to a particular parameter scale.

This research contributes to the IS knowledge base in three ways. \textit{First}, we extend the wisdom of crowds paradigm from static aggregation of parallel, independent judgments~\citep{surowiecki2005wisdom} to dynamic, relay-style sequential complementarity. Classic wisdom-of-crowds theory posits that aggregating diverse opinions yields superior collective judgment; our work reconceptualizes collective AI intelligence as an iterative coordination process where each participant builds upon predecessors' identified bottleneck. This shift from ``aggregate diverse opinions'' to ``orchestrate complementary capabilities through state-dependent coordination'' provides a new theoretical lens for IS scholars studying collective intelligence in AI systems. \textit{Second}, to operationalize this relay-style complementarity, we contribute prescriptive design knowledge~\citep{abbasi2024pathways} in the form of two transferable design principles for multi-AI coordination. DP1 (Iterative Reflection-and-Refinement) ensures the relay is feasible: effective multi-model collaboration requires a state-preserving workflow that enables successive models to reflect on and refine prior outputs. DP2 (Complementarity-Driven Model Selection) ensures the relay is productive: model transitions must be governed by verified complementarity rather than static assignment or random selection. These principles are not confined to the specific WILC instantiation but generalize to any multi-AI collaboration scenario requiring heterogeneous capabilities---such as collaborative content creation, multi-agent diagnosis, or iterative software development---offering actionable design guidance for practitioners building adaptive AI coordination systems. \textit{Third}, we make a methodological contribution by introducing model-bottleneck fitness as the basis for an adaptive selection mechanism. Instantiated through online learning (LinUCB), this mechanism updates capability beliefs from real-time feedback, learning which model resolves which type of bottleneck through accumulated experience. This provides a methodological template for building AI coordination systems that can self-improve as task distributions evolve.

Our findings yield several practical implications for enterprise deployment of LLM collaboration systems. First, the cost-performance tradeoff is highly favorable: WILC achieves performance comparable to GPT-5.2 at roughly 7$\times$ lower per-query cost, and the multi-round overhead is targeted rather than uniform---simple queries terminate after a single burn-in round while additional rounds are invested only in genuinely difficult queries. Second, because all models are self-hosted, data remains within the organization, supporting the privacy and compliance requirements that motivate private LLM deployments. Third, unlike dedicated routing methods whose effectiveness is bounded by training data coverage (Appendix~\ref{sec:routing-comparison}), WILC's evaluate-then-select mechanism and bottleneck-driven successor selection remain effective on task types not covered by the cold-start data, an important property given that real-world query distributions are rarely fully predictable. More broadly, our results suggest that capability heterogeneity across an organization's deployed models is not a liability but an asset: by strategically coordinating complementary strengths through the design principles articulated in this study, organizations can unlock collective intelligence that surpasses any individual model. The underlying principles---iterative complementarity verification and adaptive capability learning---may inform the design of multi-AI collaboration systems beyond the LLM domain, such as customer service pipelines, content moderation workflows, and collaborative analytics platforms.

This study has several limitations that point to future research opportunities. First, our evaluation covers four benchmarks with crowds of three workers; validating the framework on larger worker pools and more diverse task domains would strengthen generalizability claims. Relatedly, our empirical evidence is drawn from open-source crowds at the 14B and 30B scales. Because relay-style complementarity rests on capability heterogeneity rather than on a particular parameter scale, the mechanism is in principle applicable to crowds composed of larger or frontier models; whether the gains persist in that regime remains to be tested. Second, the current proxy reward mechanism is general-purpose and effective across our benchmarks, but designing task-adaptive reward signals that better align with domain-specific evaluation criteria could further sharpen model selection and progress assessment. Third, extending the framework to handle highly complex, structured tasks that require multi-step decomposition or cyclic problem-solving processes presents an exciting opportunity. Incorporating automatic task decomposition and more advanced collaboration strategies along such problem-solving chains could broaden WILC's applicability to a wider range of enterprise scenarios.

\bibliographystyle{unsrtnat}
\bibliography{references}

\begin{thebibliography}{61}
\providecommand{\natexlab}[1]{#1}
\providecommand{\url}[1]{\texttt{#1}}
\expandafter\ifx\csname urlstyle\endcsname\relax
  \providecommand{\doi}[1]{doi: #1}\else
  \providecommand{\doi}{doi: \begingroup \urlstyle{rm}\Url}\fi

\bibitem[Haki et~al.(2025)Haki, Safaei, Magan, and
  Griffiths]{haki2025integrating}
Kazem Haki, Dorsa Safaei, Adolfo Magan, and Martin Griffiths.
\newblock Integrating generative {AI} into enterprise platforms: {Insights}
  from salesforce.
\newblock \emph{Information Systems Journal}, 2025.

\bibitem[Raza et~al.(2025)Raza, Jahangir, Riaz, Saeed, and
  Sattar]{raza2025industrial}
Mubashar Raza, Zarmina Jahangir, Muhammad~Bilal Riaz, Muhammad~Jasim Saeed, and
  Muhammad~Awais Sattar.
\newblock Industrial applications of large language models.
\newblock \emph{Scientific Reports}, 15\penalty0 (1):\penalty0 13755, 2025.

\bibitem[Shnitzer et~al.(2023)Shnitzer, Ou, Silva, Soule, Sun, Solomon,
  Thompson, and Yurochkin]{shnitzer2023large}
Tal Shnitzer, Anthony Ou, Mirian Silva, Kate Soule, Yuekai Sun, Justin Solomon,
  Neil Thompson, and Mikhail Yurochkin.
\newblock Large language model routing with benchmark datasets.
\newblock In \emph{Annual Conference on Neural Information Processing Systems},
  2023.

\bibitem[Lu et~al.(2024)Lu, Yuan, Lin, Lin, Yuan, Zhou, and
  Zhou]{lu2024routing}
Keming Lu, Hongyi Yuan, Runji Lin, Junyang Lin, Zheng Yuan, Chang Zhou, and
  Jingren Zhou.
\newblock Routing to the expert: {Efficient} reward-guided ensemble of large
  language models.
\newblock In \emph{Proceedings of the 2024 Conference of the North American
  Chapter of the Association for Computational Linguistics: Human Language
  Technologies (Volume 1: Long Papers)}, pages 1964--1974, 2024.

\bibitem[AIMultiple(2024)]{aimultiple_enterprise_genai_2024}
AIMultiple.
\newblock Enterprise generative {AI}: {10+} use cases \& best practices, 2024.
\newblock URL \url{https://research.aimultiple.com/enterprise-generative-ai/}.
\newblock Accessed: 2025-11-29.

\bibitem[Surowiecki(2005)]{surowiecki2005wisdom}
James Surowiecki.
\newblock \emph{The Wisdom of Crowds}.
\newblock Anchor, 2005.

\bibitem[Wei et~al.(2022{\natexlab{a}})Wei, Zhang, Zhang, Chen, and
  Zeng]{wei2022combining}
Xuan Wei, Zhu Zhang, Mingyue Zhang, Weiyun Chen, and Daniel~Dajun Zeng.
\newblock Combining crowd and machine intelligence to detect false news on
  social media.
\newblock \emph{MIS Quarterly}, 46\penalty0 (2):\penalty0 977--1008,
  2022{\natexlab{a}}.

\bibitem[Atanasov et~al.(2017)Atanasov, Rescober, Stone, Swift,
  Servan-Schreiber, Tetlock, Ungar, and Mellers]{atanasov2017distilling}
Pavel Atanasov, Phillip Rescober, Eric Stone, Samuel~A Swift, Emile
  Servan-Schreiber, Philip Tetlock, Lyle Ungar, and Barbara Mellers.
\newblock Distilling the wisdom of crowds: {Prediction} markets vs. prediction
  polls.
\newblock \emph{Management Science}, 63\penalty0 (3):\penalty0 691--706, 2017.

\bibitem[Wei et~al.(2025)Wei, Zhang, Zhang, Li, and Zeng]{wei2025human}
Xuan Wei, Mingyue Zhang, Qingpeng Zhang, Zhi Li, and Daniel~Dajun Zeng.
\newblock Human-algorithm collaborative truth inference in crowdsourcing.
\newblock \emph{INFORMS Journal on Computing}, 2025.

\bibitem[Becker et~al.(2022)Becker, Guilbeault, and Smith]{becker2022crowd}
Joshua~Aaron Becker, Douglas Guilbeault, and Edward~Bishop Smith.
\newblock The crowd classification problem: {Social} dynamics of binary-choice
  accuracy.
\newblock \emph{Management Science}, 68\penalty0 (5):\penalty0 3949--3965,
  2022.

\bibitem[Thomas et~al.(2021)Thomas, Coon, Westfall, and Lee]{thomas2021model}
Bobby Thomas, Jeff Coon, Holly~A Westfall, and Michael~D Lee.
\newblock Model-based wisdom of the crowd for sequential decision-making tasks.
\newblock \emph{Cognitive Science}, 45\penalty0 (7):\penalty0 e13011, 2021.

\bibitem[Wei et~al.(2022{\natexlab{b}})Wei, Wang, Schuurmans, Bosma, Xia, Chi,
  Le, Zhou, et~al.]{wei2022chain}
Jason Wei, Xuezhi Wang, Dale Schuurmans, Maarten Bosma, Fei Xia, Ed~Chi, Quoc~V
  Le, Denny Zhou, et~al.
\newblock Chain-of-thought prompting elicits reasoning in large language
  models.
\newblock \emph{Advances in Neural Information Processing Systems},
  35:\penalty0 24824--24837, 2022{\natexlab{b}}.

\bibitem[Madaan et~al.(2023)Madaan, Tandon, Gupta, Hallinan, Gao, Wiegreffe,
  Alon, Dziri, Prabhumoye, Yang, et~al.]{madaan2023self}
Aman Madaan, Niket Tandon, Prakhar Gupta, Skyler Hallinan, Luyu Gao, Sarah
  Wiegreffe, Uri Alon, Nouha Dziri, Shrimai Prabhumoye, Yiming Yang, et~al.
\newblock {Self-Refine}: {Iterative} refinement with self-feedback.
\newblock \emph{Advances in Neural Information Processing Systems},
  36:\penalty0 46534--46594, 2023.

\bibitem[Shinn et~al.(2023)Shinn, Cassano, Gopinath, Narasimhan, and
  Yao]{shinn2023reflexion}
Noah Shinn, Federico Cassano, Ashwin Gopinath, Karthik Narasimhan, and Shunyu
  Yao.
\newblock Reflexion: {Language} agents with verbal reinforcement learning.
\newblock \emph{Advances in Neural Information Processing Systems},
  36:\penalty0 8634--8652, 2023.

\bibitem[Wang et~al.(2022)Wang, Wei, Schuurmans, Le, Chi, Narang, Chowdhery,
  and Zhou]{wang2022self}
Xuezhi Wang, Jason Wei, Dale Schuurmans, Quoc~V Le, Ed~H Chi, Sharan Narang,
  Aakanksha Chowdhery, and Denny Zhou.
\newblock Self-consistency improves chain of thought reasoning in language
  models.
\newblock In \emph{The Eleventh International Conference on Learning
  Representations}, 2022.

\bibitem[Jiang et~al.(2023)Jiang, Ren, and Lin]{jiang2023llm}
Dongfu Jiang, Xiang Ren, and Bill~Yuchen Lin.
\newblock {LLM-Blender}: {Ensembling} large language models with pairwise
  ranking and generative fusion.
\newblock In \emph{The 61st Annual Meeting Of The Association For Computational
  Linguistics}, 2023.

\bibitem[Li et~al.(2024)Li, Wang, Zeng, Wu, and Yang]{li2024survey}
Xinyi Li, Sai Wang, Siqi Zeng, Yu~Wu, and Yi~Yang.
\newblock A survey on {LLM}-based multi-agent systems: {Workflow},
  infrastructure, and challenges.
\newblock \emph{Vicinagearth}, 1\penalty0 (1):\penalty0 9, 2024.

\bibitem[Guo et~al.(2024)Guo, Chen, Wang, Chang, Pei, Chawla, Wiest, and
  Zhang]{guo2024large}
T~Guo, X~Chen, Y~Wang, R~Chang, S~Pei, NV~Chawla, O~Wiest, and X~Zhang.
\newblock Large language model based multi-agents: {A} survey of progress and
  challenges.
\newblock In \emph{33rd International Joint Conference on Artificial
  Intelligence (IJCAI 2024)}. IJCAI; Cornell arxiv, 2024.

\bibitem[{Cambridge Dictionary}(2025)]{cambridge2025complementarity}
{Cambridge Dictionary}.
\newblock English dictionary, 2025.
\newblock URL
  \url{https://dictionary.cambridge.org/dictionary/english/complementarity}.
\newblock Retrieved November 24, 2025.

\bibitem[Yin et~al.(2021)Yin, Luo, and Brown]{yin2021learning}
Junming Yin, Jerry Luo, and Susan~A Brown.
\newblock Learning from crowdsourced multi-labeling: {A} variational bayesian
  approach.
\newblock \emph{Information Systems Research}, 32\penalty0 (3):\penalty0
  752--773, 2021.

\bibitem[Hemmer et~al.(2025)Hemmer, Schemmer, K{\"u}hl, V{\"o}ssing, and
  Satzger]{hemmer2025complementarity}
Patrick Hemmer, Max Schemmer, Niklas K{\"u}hl, Michael V{\"o}ssing, and Gerhard
  Satzger.
\newblock Complementarity in human-{AI} collaboration: {Concept}, sources, and
  evidence.
\newblock \emph{European Journal of Information Systems}, pages 1--24, 2025.

\bibitem[Bansal et~al.(2021)Bansal, Wu, Zhou, Fok, Nushi, Kamar, Ribeiro, and
  Weld]{bansal2021does}
Gagan Bansal, Tongshuang Wu, Joyce Zhou, Raymond Fok, Besmira Nushi, Ece Kamar,
  Marco~Tulio Ribeiro, and Daniel Weld.
\newblock Does the whole exceed its parts? {The} effect of {AI} explanations on
  complementary team performance.
\newblock In \emph{Proceedings of the 2021 CHI Conference on Human Factors in
  Computing Systems}, pages 1--16, 2021.

\bibitem[Hevner et~al.(2004)Hevner, March, Park, and Ram]{hevner2004design}
Alan~R Hevner, Salvatore~T March, Jinsoo Park, and Sudha Ram.
\newblock Design science in information systems research.
\newblock \emph{MIS quarterly}, pages 75--105, 2004.

\bibitem[Abbasi et~al.(2024)Abbasi, Parsons, Pant, Sheng, and
  Sarker]{abbasi2024pathways}
Ahmed Abbasi, Jeffrey Parsons, Gautam Pant, Olivia R~Liu Sheng, and Suprateek
  Sarker.
\newblock Pathways for design research on artificial intelligence.
\newblock \emph{Information Systems Research}, 35\penalty0 (2):\penalty0
  441--459, 2024.

\bibitem[Fang et~al.(2025)Fang, Hu, Chau, and Chen]{fang2025computational}
Xiao Fang, Paul~J Hu, Michael Chau, and Hsinchun Chen.
\newblock Computational design science: {A} critical information systems
  research area contributing to artificial intelligence and data science.
\newblock \emph{Available at SSRN 5455094}, 2025.

\bibitem[Kamoi et~al.(2024)Kamoi, Zhang, Zhang, Han, and Zhang]{kamoi2024can}
Ryo Kamoi, Yusen Zhang, Nan Zhang, Jiawei Han, and Rui Zhang.
\newblock When can {LLMs} actually correct their own mistakes? {A} critical
  survey of self-correction of {LLMs}.
\newblock \emph{Transactions of the Association for Computational Linguistics},
  12:\penalty0 1417--1440, 2024.

\bibitem[He et~al.(2025)He, Jiang, Wang, Yang, Peng, Yan, Shen, and
  Xu]{he2025makes}
Chen He, Xun Jiang, Lei Wang, Hao Yang, Chong Peng, Peng Yan, Fumin Shen, and
  Xing Xu.
\newblock What makes reasoning invalid: {Echo} reflection mitigation for large
  language models.
\newblock \emph{arXiv preprint arXiv:2511.06380}, 2025.

\bibitem[Li et~al.(2010)Li, Chu, Langford, and Schapire]{li2010contextual}
Lihong Li, Wei Chu, John Langford, and Robert~E Schapire.
\newblock A contextual-bandit approach to personalized news article
  recommendation.
\newblock In \emph{Proceedings of the 19th International Conference on World
  Wide Web}, pages 661--670, 2010.

\bibitem[F{\"u}gener et~al.(2022)F{\"u}gener, Grahl, Gupta, and
  Ketter]{fugener2022cognitive}
Andreas F{\"u}gener, J{\"o}rn Grahl, Alok Gupta, and Wolfgang Ketter.
\newblock Cognitive challenges in human--artificial intelligence collaboration:
  {Investigating} the path toward productive delegation.
\newblock \emph{Information Systems Research}, 33\penalty0 (2):\penalty0
  678--696, 2022.

\bibitem[Donahue et~al.(2022)Donahue, Chouldechova, and
  Kenthapadi]{donahue2022human}
Kate Donahue, Alexandra Chouldechova, and Krishnaram Kenthapadi.
\newblock Human-algorithm collaboration: {Achieving} complementarity and
  avoiding unfairness.
\newblock In \emph{Proceedings of the 2022 ACM Conference on Fairness,
  Accountability, and Transparency}, pages 1639--1656, 2022.

\bibitem[Steyvers et~al.(2022)Steyvers, Tejeda, Kerrigan, and
  Smyth]{steyvers2022bayesian}
Mark Steyvers, Heliodoro Tejeda, Gavin Kerrigan, and Padhraic Smyth.
\newblock Bayesian modeling of human--{AI} complementarity.
\newblock \emph{Proceedings of the National Academy of Sciences}, 119\penalty0
  (11):\penalty0 e2111547119, 2022.

\bibitem[Dietterich(2000)]{dietterich2000ensemble}
Thomas~G Dietterich.
\newblock Ensemble methods in machine learning.
\newblock In \emph{International workshop on multiple classifier systems},
  pages 1--15. Springer, 2000.

\bibitem[Sagi and Rokach(2018)]{sagi2018ensemble}
Omer Sagi and Lior Rokach.
\newblock Ensemble learning: {A} survey.
\newblock \emph{Wiley interdisciplinary reviews: data mining and knowledge
  discovery}, 8\penalty0 (4):\penalty0 e1249, 2018.

\bibitem[Fu et~al.(2023)Fu, Peng, Sabharwal, Clark, and Khot]{fu2023complexity}
Yao Fu, Hao Peng, Ashish Sabharwal, Peter Clark, and Tushar Khot.
\newblock Complexity-based prompting for multi-step reasoning.
\newblock In \emph{11th International Conference on Learning Representations,
  ICLR 2023}, 2023.

\bibitem[Li et~al.(2023)Li, Lin, Zhang, Fu, Chen, Lou, and Chen]{li2023making}
Yifei Li, Zeqi Lin, Shizhuo Zhang, Qiang Fu, Bei Chen, Jian-Guang Lou, and
  Weizhu Chen.
\newblock Making language models better reasoners with step-aware verifier.
\newblock In \emph{Proceedings of the 61st Annual Meeting of the Association
  for Computational Linguistics (Volume 1: Long Papers)}, pages 5315--5333,
  2023.

\bibitem[Lin et~al.(2024{\natexlab{a}})Lin, Fu, Liu, Li, Gong, Wan, Zhang,
  Wang, Zhang, and Gai]{lin2024just}
Lei Lin, Jiayi Fu, Pengli Liu, Qingyang Li, Yan Gong, Junchen Wan, Fuzheng
  Zhang, Zhongyuan Wang, Di~Zhang, and Kun Gai.
\newblock Just ask one more time! {Self}-agreement improves reasoning of
  language models in (almost) all scenarios.
\newblock In \emph{Findings of the Association for Computational Linguistics:
  ACL 2024}, pages 3829--3852, 2024{\natexlab{a}}.

\bibitem[Xiong et~al.(2023)Xiong, Ding, Cao, Liu, and Qin]{xiong2023examining}
Kai Xiong, Xiao Ding, Yixin Cao, Ting Liu, and Bing Qin.
\newblock Examining inter-consistency of large language models collaboration:
  {An} in-depth analysis via debate.
\newblock In \emph{Findings of the Association for Computational Linguistics:
  EMNLP 2023}, pages 7572--7590, 2023.

\bibitem[Hong et~al.(2024)Hong, Zhuge, Chen, Zheng, Cheng, Zhang, Wang, Wang,
  Yau, Lin, et~al.]{hong2024metagpt}
Sirui Hong, Mingchen Zhuge, Jonathan Chen, Xiawu Zheng, Yuheng Cheng, Ceyao
  Zhang, Jinlin Wang, Zili Wang, Steven Ka~Shing Yau, Zijuan Lin, et~al.
\newblock {MetaGPT}: {Meta} programming for a multi-agent collaborative
  framework.
\newblock In \emph{12th International Conference on Learning Representations,
  ICLR 2024}, 2024.

\bibitem[Qian et~al.(2024)Qian, Liu, Liu, Chen, Dang, Li, Yang, Chen, Su, Cong,
  et~al.]{qian2024chatdev}
Chen Qian, Wei Liu, Hongzhang Liu, Nuo Chen, Yufan Dang, Jiahao Li, Cheng Yang,
  Weize Chen, Yusheng Su, Xin Cong, et~al.
\newblock Chatdev: {Communicative} agents for software development.
\newblock In \emph{Proceedings of the 62nd Annual Meeting of the Association
  for Computational Linguistics (Volume 1: Long Papers)}, pages 15174--15186,
  2024.

\bibitem[Zheng et~al.(2023)Zheng, Zhang, Nguyen, Rampal, Alawadhi, Rong,
  Head-Gordon, Borgs, Chayes, and Yaghi]{zheng2023chatgpt}
Zhiling Zheng, Oufan Zhang, Ha~L Nguyen, Nakul Rampal, Ali~H Alawadhi, Zichao
  Rong, Teresa Head-Gordon, Christian Borgs, Jennifer~T Chayes, and Omar~M
  Yaghi.
\newblock {ChatGPT} research group for optimizing the crystallinity of {MOFs}
  and {COFs}.
\newblock \emph{ACS Central Science}, 9\penalty0 (11):\penalty0 2161--2170,
  2023.

\bibitem[Tang et~al.(2024)Tang, Zou, Zhang, Li, Zhao, Zhang, Cohan, and
  Gerstein]{tang2024medagents}
Xiangru Tang, Anni Zou, Zhuosheng Zhang, Ziming Li, Yilun Zhao, Xingyao Zhang,
  Arman Cohan, and Mark Gerstein.
\newblock {MedAgents}: {Large} language models as collaborators for zero-shot
  medical reasoning.
\newblock In \emph{{ICLR} 2024 Workshop on Large Language Model ({LLM})
  Agents}, 2024.

\bibitem[Du et~al.(2023)Du, Li, Torralba, Tenenbaum, and
  Mordatch]{du2023improving}
Yilun Du, Shuang Li, Antonio Torralba, Joshua~B Tenenbaum, and Igor Mordatch.
\newblock Improving factuality and reasoning in language models through
  multiagent debate.
\newblock In \emph{Forty-first International Conference on Machine Learning},
  2023.

\bibitem[Park et~al.(2022)Park, Popowski, Cai, Morris, Liang, and
  Bernstein]{park2022social}
Joon~Sung Park, Lindsay Popowski, Carrie Cai, Meredith~Ringel Morris, Percy
  Liang, and Michael~S Bernstein.
\newblock Social simulacra: {Creating} populated prototypes for social
  computing systems.
\newblock In \emph{Proceedings of the 35th Annual ACM Symposium on User
  Interface Software and Technology}, pages 1--18, 2022.

\bibitem[Chen et~al.(2023)Chen, Dong, Shu, Zhang, Sesay, Karlsson, Fu, and
  Shi]{chen2023autoagents}
Guangyao Chen, Siwei Dong, Yu~Shu, Ge~Zhang, Jaward Sesay, B{\"o}rje~F
  Karlsson, Jie Fu, and Yemin Shi.
\newblock Autoagents: {A} framework for automatic agent generation.
\newblock \emph{arXiv preprint arXiv:2309.17288}, 2023.

\bibitem[Robbins(1952)]{robbins1952some}
Herbert Robbins.
\newblock Some aspects of the sequential design of experiments.
\newblock \emph{Bulletin of the American Mathematical Society}, 58\penalty0
  (5):\penalty0 527--535, 1952.

\bibitem[Sutton et~al.(1998)Sutton, Barto, et~al.]{sutton1998reinforcement}
Richard~S Sutton, Andrew~G Barto, et~al.
\newblock \emph{Reinforcement learning: {An} introduction}.
\newblock MIT press Cambridge, 1998.

\bibitem[Auer et~al.(2002)Auer, Cesa-Bianchi, and Fischer]{auer2002finite}
Peter Auer, Nicolo Cesa-Bianchi, and Paul Fischer.
\newblock Finite-time analysis of the multiarmed bandit problem.
\newblock \emph{Machine Learning}, 47\penalty0 (2):\penalty0 235--256, 2002.

\bibitem[Sedgwick(2014)]{sedgwick2014spearman}
Philip Sedgwick.
\newblock Spearman’s rank correlation coefficient.
\newblock \emph{BMJ}, 349, 2014.

\bibitem[Chen(2021)]{chen2021evaluating}
Mark Chen.
\newblock Evaluating large language models trained on code.
\newblock \emph{arXiv preprint arXiv:2107.03374}, 2021.

\bibitem[Hendrycks et~al.(2021{\natexlab{a}})Hendrycks, Burns, Kadavath, Arora,
  Basart, Tang, Song, and Steinhardt]{hendrycks2021measuring}
Dan Hendrycks, Collin Burns, Saurav Kadavath, Akul Arora, Steven Basart, Eric
  Tang, Dawn Song, and Jacob Steinhardt.
\newblock Measuring mathematical problem solving with the math dataset.
\newblock \emph{arXiv preprint arXiv:2103.03874}, 2021{\natexlab{a}}.

\bibitem[Hendrycks et~al.(2021{\natexlab{b}})Hendrycks, Burns, Basart, Zou,
  Mazeika, Song, and Steinhardt]{hendryckstest2021}
Dan Hendrycks, Collin Burns, Steven Basart, Andy Zou, Mantas Mazeika, Dawn
  Song, and Jacob Steinhardt.
\newblock Measuring massive multitask language understanding.
\newblock \emph{Proceedings of the International Conference on Learning
  Representations (ICLR)}, 2021{\natexlab{b}}.

\bibitem[Chen et~al.(2024{\natexlab{a}})Chen, Zhang, Xu, Ren, and
  Yang]{chen2024viseval}
Nan Chen, Yuge Zhang, Jiahang Xu, Kan Ren, and Yuqing Yang.
\newblock {VisEval}: {A} benchmark for data visualization in the era of large
  language models.
\newblock \emph{IEEE Transactions on Visualization and Computer Graphics},
  2024{\natexlab{a}}.

\bibitem[Yao et~al.(2022)Yao, Zhao, Yu, Du, Shafran, Narasimhan, and
  Cao]{yao2022react}
Shunyu Yao, Jeffrey Zhao, Dian Yu, Nan Du, Izhak Shafran, Karthik~R Narasimhan,
  and Yuan Cao.
\newblock {ReAct}: {Synergizing} reasoning and acting in language models.
\newblock In \emph{The Eleventh International Conference on Learning
  Representations}, 2022.

\bibitem[Chen et~al.(2024{\natexlab{b}})Chen, Aksitov, Alon, Ren, Xiao, Yin,
  Prakash, Sutton, Wang, and Zhou]{chen2024universal}
Xinyun Chen, Renat Aksitov, Uri Alon, Jie Ren, Kefan Xiao, Pengcheng Yin,
  Sushant Prakash, Charles Sutton, Xuezhi Wang, and Denny Zhou.
\newblock Universal self-consistency for large language models.
\newblock In \emph{ICML Workshop on In-Context Learning}, 2024{\natexlab{b}}.

\bibitem[Lin et~al.(2024{\natexlab{b}})Lin, Tang, Tang, Yang, Chen, Wang, Xiao,
  Dang, Gan, and Han]{lin2024awq}
Ji~Lin, Jiaming Tang, Haotian Tang, Shang Yang, Wei-Ming Chen, Wei-Chen Wang,
  Guangxuan Xiao, Xingyu Dang, Chuang Gan, and Song Han.
\newblock {AWQ}: Activation-aware weight quantization for on-device {LLM}
  compression and acceleration.
\newblock \emph{Proceedings of machine learning and systems}, 6:\penalty0
  87--100, 2024{\natexlab{b}}.

\bibitem[White et~al.(2024)White, Dooley, Roberts, Pal, Feuer, Jain,
  Shwartz-Ziv, Jain, Saifullah, Naidu, et~al.]{white2024livebench}
Colin White, Samuel Dooley, Manley Roberts, Arka Pal, Ben Feuer, Siddhartha
  Jain, Ravid Shwartz-Ziv, Neel Jain, Khalid Saifullah, Siddartha Naidu, et~al.
\newblock {LiveBench}: {A} challenging, contamination-free {LLM} benchmark.
\newblock \emph{arXiv preprint arXiv:2406.19314}, 2024.

\bibitem[Nussbaum et~al.(2025)Nussbaum, Morris, Mulyar, and
  Duderstadt]{nussbaum2025nomic}
Zach Nussbaum, John~Xavier Morris, Andriy Mulyar, and Brandon Duderstadt.
\newblock Nomic embed: Training a reproducible long context text embedder.
\newblock \emph{Transactions on Machine Learning Research}, 2025.

\bibitem[Feng et~al.(2024)Feng, Zhang, and You]{feng2024llmrouter}
Tao Feng, Haozhen Zhang, and Jiaxuan You.
\newblock {LLMRouter}: An open-source library for {LLM} routing.
\newblock \url{https://github.com/ulab-uiuc/LLMRouter}, 2024.

\bibitem[Zhang et~al.(2025)Zhang, Feng, and You]{zhang2025routerr1}
Haozhen Zhang, Tao Feng, and Jiaxuan You.
\newblock {Router-R1}: Teaching {LLMs} multi-round routing and aggregation via
  reinforcement learning.
\newblock \emph{arXiv preprint arXiv:2506.09033}, 2025.

\bibitem[Ong et~al.(2025)Ong, Almahairi, Wu, Chiang, Wu, Gonzalez, Kadous, and
  Stoica]{ong2025routellm}
Isaac Ong, Amjad Almahairi, Vincent Wu, Wei-Lin Chiang, Tianhao Wu, Joseph~E
  Gonzalez, M~Waleed Kadous, and Ion Stoica.
\newblock {RouteLLM}: Learning to route {LLMs} from preference data.
\newblock In \emph{The Thirteenth International Conference on Learning
  Representations}, 2025.

\bibitem[Feng et~al.(2025)Feng, Shen, and You]{feng2025graphrouter}
Tao Feng, Yanzhen Shen, and Jiaxuan You.
\newblock {GraphRouter}: A graph-based router for {LLM} selections.
\newblock In \emph{International Conference on Learning Representations},
  volume 2025, pages 26186--26203, 2025.

\end{thebibliography}

\clearpage
\appendix

\begin{center}
  {\Large\bfseries\scshape Appendix}
\end{center}
\vspace{0.5\baselineskip}

\makeatletter
\setcounter{figure}{0}
\setcounter{table}{0}
\@addtoreset{figure}{section}
\@addtoreset{table}{section}
\renewcommand{\thefigure}{\thesection\arabic{figure}}
\renewcommand{\thetable}{\thesection\arabic{table}}
\makeatother

\section{Notation}
\label{sec:notation}
Table~\ref{tab:notation} summarizes the major notation used in the paper.

\begin{table}[!ht]
\centering
\caption{Major Notation Used in the Paper}
\label{tab:notation}
\small
\setlength{\tabcolsep}{6pt}
\renewcommand{\arraystretch}{1.2}  
\begin{tabular}{p{0.24\linewidth}p{0.68\linewidth}}
\toprule
\textbf{Notation} & \textbf{Description} \\
\midrule
\multicolumn{2}{l}{\textbf{Setting}} \\
$K$ & Number of workers in the LLM crowd. \\
$\mathcal{W}=\{w_1, w_2, \dots, w_K\}$ & Worker pool, where $w_k$ denotes the $k$-th worker model. \\
$d$ & Dimensionality of the context vector (embedding dimension). \\
\midrule
\multicolumn{2}{l}{\textbf{Iterative Problem-Solving Process}} \\
$r$ & Index of the current round in the iterative reflection-and-refinement process. \\
$r_{\max}$ & Maximum number of collaboration rounds (set to 6 in our experiments). \\
$q_0$ & Initial query. \\
$q_r$ & Query at round $r$. \\
$a_r$ & Answer generated at round $r$. \\
$f_r$ & Coordinator reflection generated at round $r$. \\
$w_r$ & Worker selected at round $r$. \\
\midrule
\multicolumn{2}{l}{\textbf{Complementarity Verification Mechanism}} \\
$\mathbf{x}_{q_r}$ & The $d$-dimensional context vector obtained by embedding the query $q_r$. \\
$\boldsymbol{\theta}_{w_k}$ & The $d$-dimensional capability vector of worker $w_k$. \\
$\boldsymbol{\hat{\theta}}_{w_k}$ & The $d$-dimensional estimated capability vector of worker $w_k$. \\
$\boldsymbol{\hat{\theta}}_{w_k}^{\text{cold}}$ & The $d$-dimensional estimated capability vector of worker $w_k$ in cold-start phase. \\
$\mathcal{Q}$ & Auxiliary query set used in the cold-start phase. \\
$\boldsymbol{D}_{w_k}$ & The $m \times d$ historical context matrix for worker $w_k$ across $m$ historical trials. \\
$\boldsymbol{c}_{w_k}$ & The $m$-dimensional actual reward vector corresponding to $\boldsymbol{D}_{w_k}$. \\
$\boldsymbol{A}_{w_k}$ & The $d \times d$ LinUCB matrix for worker $w_k$, where $\boldsymbol{A}_{w_k}=\boldsymbol{D}_{w_k}^\top \boldsymbol{D}_{w_k}+\boldsymbol{I}$. \\
$\boldsymbol{b}_{w_k}$ & The $d$-dimensional LinUCB vector for worker $w_k$, where $\boldsymbol{b}_{w_k}=\boldsymbol{D}_{w_k}^\top \boldsymbol{c}_{w_k}$. \\
$\alpha$ & Exploration parameter in the LinUCB algorithm. \\
$\mathrm{UCB}_{w_k}(r)$ & Upper confidence bound score of worker $w_k$ at round $r$. \\
$R_{r,w_k}$ & Proxy reward of worker $w_k$ at round $r$. \\
$s$ & Number of issues identified in the reflection. \\
$\gamma$ & Penalty assigned to each identified issue in the proxy reward. \\
$\beta$ & Weight balancing the cold-start estimate and online update in capability update. \\
$\rho$ & Spearman's rank correlation coefficient used to set $\beta$ dynamically. \\
\midrule
\multicolumn{2}{l}{\textbf{One-Step Forward Search}} \\
$a'_r$ & Simulated answer generated by the previous worker at round $r$. \\
$f'_r$ & Simulated reflection associated with $a'_r$. \\
$R'_{r,w_{r-1}}$ & Simulated reward of the previous worker at round $r$. \\
\bottomrule
\end{tabular}
\end{table}

\section{The LinUCB Algorithm}\label{sec:linucb}
This section provides a self-contained derivation of the LinUCB algorithm~\citep{li2010contextual} that underlies the worker selection mechanism of WILC. Section~\ref{sec:MAB} presents the components used in the main text; here we supply the estimation and confidence-interval arguments behind them.

\paragraph{Problem setting.} A contextual bandit proceeds over discrete rounds. At round $t$, the learner observes a context vector $\mathbf{x}_{t,a} \in \mathbb{R}^{d}$ for each arm $a$ in the arm set $\mathcal{A}$, selects one arm $a_t$, and receives only the reward $R_{t,a_t}$ of the selected arm. Rewards of unselected arms remain unobserved, so the learner must balance exploiting arms with high estimated reward against exploring arms whose estimates remain uncertain. LinUCB assumes the expected reward is linear in the context,
\begin{equation}
\mathbb{E}[R_{t,a} \mid \mathbf{x}_{t,a}] = \mathbf{x}_{t,a}^\top \boldsymbol{\theta}_a ,
\label{eq:linucb_linear}
\end{equation}
where $\boldsymbol{\theta}_a \in \mathbb{R}^{d}$ is an unknown parameter vector that characterizes arm $a$.

\paragraph{Ridge estimation of $\boldsymbol{\theta}_a$.} Let $\boldsymbol{D}_a \in \mathbb{R}^{m \times d}$ denote the design matrix whose rows are the $m$ contexts previously observed for arm $a$, and let $\boldsymbol{c}_a \in \mathbb{R}^{m}$ collect the corresponding observed rewards. LinUCB estimates $\boldsymbol{\theta}_a$ by ridge regression,
\begin{equation}
\boldsymbol{\hat{\theta}}_a = \arg\min_{\boldsymbol{\theta}} \; \lVert \boldsymbol{D}_a \boldsymbol{\theta} - \boldsymbol{c}_a \rVert_2^2 + \lVert \boldsymbol{\theta} \rVert_2^2 = \boldsymbol{A}_a^{-1} \boldsymbol{b}_a ,
\label{eq:linucb_ridge}
\end{equation}
where $\boldsymbol{A}_a = \boldsymbol{D}_a^\top \boldsymbol{D}_a + \boldsymbol{I}$ and $\boldsymbol{b}_a = \boldsymbol{D}_a^\top \boldsymbol{c}_a$. The identity matrix serves as an $\ell_2$ regularizer and guarantees that $\boldsymbol{A}_a$ is invertible even when few observations are available, which matters in early rounds when $m < d$.

\paragraph{From estimation uncertainty to the exploration bonus.} Treating the rewards as Gaussian observations, the posterior over $\boldsymbol{\theta}_a$ is $\boldsymbol{\theta}_a \sim \mathcal{N}(\boldsymbol{A}_a^{-1}\boldsymbol{b}_a, \boldsymbol{A}_a^{-1})$. For a new context $\mathbf{x}_{t,a}$, the predicted reward $\mathbf{x}_{t,a}^\top \boldsymbol{\theta}_a$ therefore has variance $\mathbf{x}_{t,a}^\top \boldsymbol{A}_a^{-1} \mathbf{x}_{t,a}$, and its standard deviation is $\sqrt{\mathbf{x}_{t,a}^\top \boldsymbol{A}_a^{-1} \mathbf{x}_{t,a}}$. Adding a multiple of this standard deviation to the point estimate yields the upper confidence bound used for arm selection:
\begin{equation}
UCB_a(t) = \mathbf{x}_{t,a}^\top \boldsymbol{\hat{\theta}}_a + \alpha \sqrt{\mathbf{x}_{t,a}^\top \boldsymbol{A}_a^{-1} \mathbf{x}_{t,a}}, \qquad a_t = \arg\max_{a \in \mathcal{A}} UCB_a(t) .
\label{eq:linucb_ucb_appendix}
\end{equation}
This clarifies the role of the second term: it is not an arbitrary penalty but the estimated uncertainty of the reward prediction for the specific context at hand, scaled by the exploration coefficient $\alpha$. Directions in the context space that have been observed frequently for arm $a$ contribute large eigenvalues to $\boldsymbol{A}_a$, shrinking the bonus; directions that remain unexplored keep it large.

\paragraph{Online update and computational cost.} After observing $(a_t, \mathbf{x}_{t,a_t}, R_{t,a_t})$, the sufficient statistics of the selected arm admit a rank-one update, and the parameter estimate follows directly:
\begin{equation}
\boldsymbol{A}_{a_t} \leftarrow \boldsymbol{A}_{a_t} + \mathbf{x}_{t,a_t} \mathbf{x}_{t,a_t}^\top, \quad
\boldsymbol{b}_{a_t} \leftarrow \boldsymbol{b}_{a_t} + R_{t,a_t} \mathbf{x}_{t,a_t}, \quad
\boldsymbol{\hat{\theta}}_{a_t} \leftarrow \boldsymbol{A}_{a_t}^{-1} \boldsymbol{b}_{a_t} .
\label{eq:linucb_update_appendix}
\end{equation}
Equation~(\ref{eq:linucb_update_appendix}) is identical to Equation~(\ref{eq:update_A_b_standard}) in the main text and is repeated here so that this section can be read independently. Because only $\boldsymbol{A}_a$ and $\boldsymbol{b}_a$ need to be retained, the algorithm requires $O(d^2)$ memory per arm and $O(Kd^2)$ time per round for $K$ arms, without storing the full interaction history. This property is what allows WILC to revise its estimate of each worker's capability profile online, as new bottlenecks are encountered within a single problem-solving process.

\paragraph{Adaptation in WILC.} WILC departs from standard LinUCB in two respects, both detailed in Section~\ref{sec:PCF}: worker capability vectors are initialized from an auxiliary dataset during a cold-start phase rather than from scratch, and subsequent updates blend this prior estimate with query-specific proxy feedback through a weighting parameter $\beta$.

\section{Baseline Configuration}
This section describes the configuration of the baseline methods used in our experiments.

\subsection{Single Execution}\label{sec:single-execution}
For MATH-500 and MMLU, which are reasoning tasks, we employ the chain-of-thought (CoT) prompting to guide the model's reasoning process for better performance; for HumanEval, we employ the common prompts used for this benchmark, focusing on function completion; for VisEval, we use the recommended system prompt (with one-shot example) in the original benchmark. Figures \ref{fig:single-humaneval} to \ref{fig:single-viseval} show the single execution prompts for the four tasks, respectively.

\begin{tcolorbox}[breakable, enhanced, base, title={Single Execution Prompt for HumanEval}]
    \begin{lstlisting}[basicstyle=\scriptsize\ttfamily, breaklines=true, breakatwhitespace=false]
You are an AI that only responds with Python code, NOT ENGLISH. You will be given a function signature and its docstring by the user. Write your full implementation (restate the function signature). Use a Python code block to write your response. For example:\n```python\nprint('Hello world!')\n```
\end{lstlisting}
\end{tcolorbox}
\vspace{-1em}
\captionsetup{hypcap=false}
\captionof{figure}{Single Execution Prompt for HumanEval}
\label{fig:single-humaneval}

\begin{tcolorbox}[breakable, enhanced, base, title={Single Execution Prompt for MATH-500}]
    \begin{lstlisting}[basicstyle=\scriptsize\ttfamily, breaklines=true, breakatwhitespace=false]
You are a math expert. Solve the given math question step by step and output the answer in the format of \\boxed{...}, like this: \\boxed{123}.
\end{lstlisting}
\end{tcolorbox}
\vspace{-1em}
\captionsetup{hypcap=false}
\captionof{figure}{Single Execution Prompt for MATH-500}
\label{fig:single-math-500}

\begin{tcolorbox}[breakable, enhanced, base, title={Single Execution Prompt for MMLU}]
    \begin{lstlisting}[basicstyle=\scriptsize\ttfamily, breaklines=true, breakatwhitespace=false]
You are a knowledgeable academic expert. Solve the given multiple-choice question step by step and output the answer in the format of \\boxed{...}, like this: Answer: \\boxed{A}.
\end{lstlisting}
\end{tcolorbox}
\vspace{-1em}
\captionsetup{hypcap=false}
\captionof{figure}{Single Execution Prompt for MMLU}
\label{fig:single-mmlu}

\begin{tcolorbox}[breakable, enhanced, base, title={Single Execution Prompt for VisEval}]
    \begin{lstlisting}[basicstyle=\scriptsize\ttfamily, breaklines=true, breakatwhitespace=false]
You're a helpful assistant proficient in writing Python code for data visualization. Upon receiving relevant context, such as available variables and any pre-executed code, your goal is to complete the Python code to generate a visualization that meets the user's request.

# Example
## Question
### Variables:

employee_data: pandas.DataFrame(shape=(10, 4), columns=["Employee_ID", "Name", "Department", "Salary"])
        Employee_ID     Name      Department  Salary
    0          101  Alice       Engineering   12000
    1          102  Bob         Engineering   13000
    2          103  Charlie     Sales         14000
    3          104  Dana        Sales         15000
    4          105  Ethan       Marketing     16000
    5          106  Fiona       Marketing     17000
    6          107  George      Engineering   18000
    7          108  Hannah      Sales         19000
    8          109  Isaac       Marketing     20000
    9          110  Jade        Engineering   21000

### Executed Code:

import pandas as pd

### Request:

Create a scatter plot showing the relationship between the salary and the department for each employee.

## Answer
```python
import pandas as pd
import matplotlib.pyplot as plt

# Group the data by department
department_salary = employee_data.groupby(['Department'])['Salary'].mean()

# Plot the scatter plot
plt.figure(figsize=(10, 5))
plt.scatter(department_salary.index, department_salary.values)
plt.title('Relationship between Salary and Department for Each Employee')
plt.xlabel('Department')
plt.ylabel('Salary')
plt.xticks(rotation=45)
plt.show()
```
\end{lstlisting}
\end{tcolorbox}
\vspace{-1em}
\captionsetup{hypcap=false}
\captionof{figure}{Single Execution Prompt for VisEval}
\label{fig:single-viseval}

\subsection{ReAct}\label{sec:react}
Figures \ref{fig:react-humaneval} to \ref{fig:react-viseval} show the ReAct prompts used in our experiments for the four tasks, respectively. They adhere to the ``Thought-Action-Observation'' cycle inherent to the ReAct method, thereby enabling the adaptive implementation of ReAct in LLMs.

\begin{tcolorbox}[breakable, enhanced, base, title={ReAct Prompt for HumanEval}]
    \begin{lstlisting}[basicstyle=\scriptsize\ttfamily, breaklines=true, breakatwhitespace=false]
You are an AI that responds with Python code. You will be given a function signature and its docstring. Write the full implementation (restate the function signature) to complete the function. Use a Python code block to write your response. For example:\n```python\nprint('Hello world!')\n```

Please use the following format to solve the problem:

Question: the function signature and docstring you must implement
Thought: Analyze the problem, understand what the function should do, consider edge cases and implementation approach.
Action: python_code
Action Input: Describe the implementation step (e.g., initialize variables, handle edge cases, implement main logic, etc.).
Observation: The code snippet for this step.
... (this Thought/Action/Action Input/Observation can repeat N times)
Thought: I now know the final answer
Final Answer: the complete Python code wrapped in triple backticks (```), like this: ```python\nprint('Hello world!')\n```

Example:

Question: 
```python
def has_close_elements(numbers: List[float], threshold: float) -> bool:
    \"\"\" Check if in given list of numbers, are any two numbers closer to each other than
    given threshold.
    >>> has_close_elements([1.0, 2.0, 3.0], 0.5)
    False
    >>> has_close_elements([1.0, 2.8, 3.0, 4.0, 5.0, 2.0], 0.3)
    True
    \"\"\"
```
Thought: I need to check if any two numbers in the list are closer than the threshold. I should iterate through all pairs of numbers and compare their distances.
Action: python_code
Action Input: Iterate through all pairs and check if their distance is less than the threshold.
Observation: The code to implement this is:
```python
from typing import List

def has_close_elements(numbers: List[float], threshold: float) -> bool:
    for i in range(len(numbers)):
        for j in range(i + 1, len(numbers)):
            if abs(numbers[i] - numbers[j]) < threshold:
                return True
    return False
```
Thought: I now know the final answer
Final Answer:
```python
from typing import List

def has_close_elements(numbers: List[float], threshold: float) -> bool:
    \"\"\" Check if in given list of numbers, are any two numbers closer to each other than
    given threshold.
    >>> has_close_elements([1.0, 2.0, 3.0], 0.5)
    False
    >>> has_close_elements([1.0, 2.8, 3.0, 4.0, 5.0, 2.0], 0.3)
    True
    \"\"\"
    for i in range(len(numbers)):
        for j in range(i + 1, len(numbers)):
            if abs(numbers[i] - numbers[j]) < threshold:
                return True
    return False
```
\end{lstlisting}
\end{tcolorbox}
\vspace{-1em}
\captionsetup{hypcap=false}
\captionof{figure}{ReAct Prompt for HumanEval}
\label{fig:react-humaneval}

\begin{tcolorbox}[breakable, enhanced, base, title={ReAct Prompt for MATH-500}]
    \begin{lstlisting}[basicstyle=\scriptsize\ttfamily, breaklines=true, breakatwhitespace=false]
You are a math expert. Solve the given math question step by step and output the answer in the format of \\boxed{...}, like this: \\boxed{123}.

Please use the following format to solve the problem:

Question: the input question you must answer
Thought: Analyze the problem, consider possible solution strategies and approaches.
Action: Determine the next mathematical operation (such as applying a specific formula, transforming an equation, computing an expression, etc.).
Action Input: The input to the action (e.g., the formula to apply, the equation to transform, the expression to compute, etc.).
Observation: The result of the action (e.g., the result of the formula application, the transformed equation, the computed expression, etc.).
... (this Thought/Action/Action Input/Observation can repeat N times)
Thought: I now know the final answer
Final Answer: the final answer to the original input question in the format of \\boxed{...}, like this: \\boxed{123}.

Example:

Question: Find the area of a triangle with base 8 cm and height 6 cm.
Thought: I need to use the formula for the area of a triangle: Area = (1/2) * base * height.
Action: Apply the formula.
Action Input: The base is 8 cm and the height is 6 cm.
Observation: The area of the triangle is (1/2) * 8 cm * 6 cm = 24 cm^2.
Thought: I now know the final answer
Final Answer: \\boxed{24}.
\end{lstlisting}
\end{tcolorbox}
\vspace{-1em}
\captionsetup{hypcap=false}
\captionof{figure}{ReAct Prompt for MATH-500}
\label{fig:react-math-500}

\begin{tcolorbox}[breakable, enhanced, base, title={ReAct Prompt for MMLU}]
    \begin{lstlisting}[basicstyle=\scriptsize\ttfamily, breaklines=true, breakatwhitespace=false]
You are a knowledgeable academic expert. Solve the given multiple-choice question step by step and output the answer in the format of \\boxed{...}, like this: Answer: \\boxed{A}.

Please use the following format to solve the problem:

Question: the multiple-choice question you must answer
Thought: Analyze the question, understand what is being asked, recall relevant knowledge.
Action: reasoning
Action Input: Analyze each option or work through the problem step by step.
Observation: The result of analyzing this option or reasoning step.
... (this Thought/Action/Action Input/Observation can repeat N times)
Thought: I now know the final answer
Final Answer: Answer: \\boxed{X} where X is A, B, C, or D

Example:

Question: What is the capital of France?
A. London
B. Berlin
C. Paris
D. Madrid

Thought: I need to recall which city is the capital of France.
Action: reasoning
Action Input: Analyze each option - London is the capital of UK, Berlin is the capital of Germany, Paris is known to be the capital of France, Madrid is the capital of Spain.
Observation: Paris is the capital city of France.
Thought: I now know the final answer
Final Answer: Answer: \\boxed{C}.
\end{lstlisting}
\end{tcolorbox}
\vspace{-1em}
\captionsetup{hypcap=false}
\captionof{figure}{ReAct Prompt for MMLU}
\label{fig:react-mmlu}

\begin{tcolorbox}[breakable, enhanced, base, title={ReAct Prompt for VisEval}]
    \begin{lstlisting}[basicstyle=\scriptsize\ttfamily, breaklines=true, breakatwhitespace=false]
You're a helpful assistant proficient in writing Python code for data visualization. Upon receiving relevant context, such as available variables and any pre-executed code, your goal is to complete the Python code to generate a visualization that meets the user's request.

Please use the following format to solve the problem:

Question: the input question you must answer
Thought: you should always think about what to do (e.g., I need to group the data by which column, I need to plot a scatter plot, etc.)
Action: the action to take (use `python_code` only)
Action Input: the input to the action (e.g., the column to group by, the chart type to plot, etc.)
Observation: the result of the action (e.g., the Python code to implement the action, etc.)
... (this Thought/Action/Action Input/Observation can repeat N times)
Thought: I now know the final answer
Final Answer: the final answer to the original input question (the complete Python code wrapped in triple backticks (```))

Example:

Thought: I need to group the data by the 'Opening_year' and sum the 'capacity' for each year. Then I will plot a line chart
Action: python_code
Action Input: the column to group by: 'Opening_year', the column to sum: 'capacity', the chart type to plot: 'line'
Observation: The python code to generate a line chart displaying the sum of capacity of cinemas open for each year is as follows:
```python
import pandas as pd
import matplotlib.pyplot as plt

# Assuming cinema_dataset is already loaded as a pandas DataFrame

grouped_data = cinema_dataset.groupby('Opening_year')['Capacity'].sum()

plt.plot(grouped_data.index, grouped_data.values)
plt.xlabel('Year')
plt.ylabel('Total Capacity')
plt.title('Sum of Capacity per Year')
plt.show()
```
Thought: I now know the final answer
Final Answer: 
```python
import pandas as pd
import matplotlib.pyplot as plt

# Assuming cinema_dataset is already loaded as a pandas DataFrame

grouped_data = cinema_dataset.groupby('Opening_year')['Capacity'].sum()

plt.plot(grouped_data.index, grouped_data.values)
plt.xlabel('Year')
plt.ylabel('Total Capacity')
plt.title('Sum of Capacity per Year')
plt.show()
```
\end{lstlisting}
\end{tcolorbox}
\vspace{-1em}
\captionsetup{hypcap=false}
\captionof{figure}{ReAct Prompt for VisEval}
\label{fig:react-viseval}

\subsection{Self-Ensemble and Heterogeneous Ensemble}\label{sec:self-ensemble-and-heterogeneous-ensemble}
The self-ensemble and heterogeneous ensemble share the same prompt for response synthesis. Guided by this prompt, in actual implementations of model ensemble, we provide the original question along with multiple responses as context to the model. The model then analyzes all provided responses and generates a synthesized or selected final answer based on its evaluation.

\begin{tcolorbox}[breakable, enhanced, base, title={Response Synthesis Prompt}]
    \begin{lstlisting}[basicstyle=\scriptsize\ttfamily, breaklines=true, breakatwhitespace=false]
You are an expert evaluator. You will be given multiple solutions/answers to the same problem. Your goal is to analyze these solutions and select or synthesize the most consistent and correct answer.

Please review all the provided solutions carefully and consider:
1. Which solution(s) appear most frequently (consensus)?
2. Which solution(s) have the most sound reasoning or implementation?
3. Which solution(s) best address the problem requirements?
\end{lstlisting}
\end{tcolorbox}
\vspace{-1em}
\captionsetup{hypcap=false}
\captionof{figure}{Response Synthesis Prompt}
\label{fig:ensemble-prompt}

\section{Comparison with Query Routing}\label{sec:routing-comparison}

Query routing methods assign each incoming query to the single model predicted to be most suitable, typically based on a routing function learned from labeled training data~\citep{shnitzer2023large,feng2024llmrouter}. Since routing represents an alternative paradigm for leveraging multiple models, we provide a dedicated comparison here. 

All routing methods use the same worker pool as WILC (see Table~\ref{tab:llms} in the main text). For a fair comparison, we train all routing methods on the worker performance data collected during WILC's cold-start stage. The cold-start data, LiveBench~\citep{white2024livebench}, covers six domains: reasoning, data analysis, coding, instruction following, math, and language (1,436 queries total). All query embeddings are computed using the same \texttt{nomic-embed-text:v1.5} model~\citep{nussbaum2025nomic} as WILC. We run each method with five random seeds and report the mean performance.

\subsection{Routing Methods}

We evaluate three representative routing methods:

\begin{itemize}
    \item \textbf{LLM Routing}~\citep{shnitzer2023large}: For each worker model, LLM Routing learns a routing function based on query embeddings to predict the probability that the model will provide the correct answer. At inference time, the new query is passed through all per-model routing functions, and the worker model with the highest predicted correctness probability is selected.
  \item \textbf{Elo Router}~\citep{feng2024llmrouter}: Elo Router constructs a symmetric pairwise comparison table among all worker models from the training data and fits Elo ratings via logistic regression (without intercept). At inference time, every query is routed to the worker with the globally highest Elo rating, regardless of query content. This is a query-agnostic baseline.
  \item \textbf{Router-R1}~\citep{zhang2025routerr1}: For each worker model, we train a two-layer MLP (384$\to$256$\to$128$\to$1) with MSE loss (Adam, lr$=$1e-3, 300 epochs) to predict the proxy reward score. At inference time, all per-model MLPs score the query and the model with the highest predicted reward is selected. This is a simplified offline adaptation of the original Router-R1, which uses a vLLM-based agentic router with chain-of-thought reasoning.
\end{itemize}

We additionally include WILC restricted to a single round ($r_{\max}{=}1$), which corresponds to the ``w/o multi-round'' ablation in the main text (Section~\ref{sec:ablation-study}). Unlike routing methods that predict model suitability from query embeddings, this configuration employs an \emph{evaluate-then-select} strategy: all workers produce responses on the target query during the burn-in stage, and the best-performing output is directly returned based on observed proxy rewards. This disentangles the contribution of WILC's model selection from its iterative multi-round collaboration.

Two other prominent routing methods are not included in our comparison. RouteLLM~\citep{ong2025routellm} is designed to route between a strong but expensive model and a weak but cheap model to optimize the cost-quality tradeoff, an objective that does not apply to our setting where all workers are open-source models of comparable scale with no meaningful cost differential. GraphRouter~\citep{feng2025graphrouter} trains a GNN over a heterogeneous query--model graph and requires large-scale interaction records (e.g., 24{,}000 query--model pairs in the original paper) to learn effective node representations, which exceeds the amount of training data available in our cold-start setup.

\subsection{Results}

Table~\ref{tab:routing_comparison} presents the comparison between routing methods, WILC ($r_{\max}{=}1$, i.e., burn-in only), and the full WILC framework.

\begin{table}[!ht]
\centering
\caption{Comparison with Query Routing Methods (\%)}
\label{tab:routing_comparison}
\small
\begin{threeparttable}
\begin{tabular*}{\textwidth}{@{\extracolsep{\fill}}l|lllll}
\toprule
\textbf{Method} & \textbf{HumanEval} & \textbf{MATH-500} & \textbf{MMLU} & \textbf{VisEval} & \textbf{AVG.} \\
\midrule
\multicolumn{6}{c}{\textit{14B Scale}} \\
\midrule
LLM Routing & 86.46\sym{***} & 87.20\sym{***} & 83.93\sym{***} & 54.27\sym{***} & 77.97\sym{***} \\
Elo Router & \textbf{95.61} & \textbf{90.40} & 87.37\sym{**} & 49.73\sym{***} & 80.78\sym{***} \\
Router-R1 & 85.73\sym{***} & 82.24\sym{***} & 81.99\sym{***} & 56.73\sym{***} & 76.67\sym{***} \\
\midrule
WILC ($r_{\max}{=}1$) & 95.12 & 89.46\sym{**} & \underline{88.64} & \underline{72.06}\sym{***} & \underline{86.32}\sym{***} \\
WILC & \underline{95.48} & \underline{89.98} & \textbf{88.92} & \textbf{76.72} & \textbf{87.78} \\
\midrule\midrule
\multicolumn{6}{c}{\textit{30B Scale}} \\
\midrule
LLM Routing & 89.27\sym{**} & 85.08\sym{***} & 85.25\sym{***} & 61.27\sym{***} & 80.22\sym{***} \\
Elo Router & 87.80\sym{***} & 71.96\sym{***} & 84.01\sym{***} & 63.20\sym{***} & 76.74\sym{***} \\
Router-R1 & 85.12\sym{***} & 79.84\sym{***} & 84.22\sym{***} & 61.40\sym{***} & 77.65\sym{***} \\
\midrule
WILC ($r_{\max}{=}1$) & \underline{90.58} & \underline{88.40}\sym{*} & \underline{87.66} & \underline{80.70}\sym{***} & \underline{86.84}\sym{**} \\
WILC & \textbf{91.12} & \textbf{88.56} & \textbf{87.72} & \textbf{84.06} & \textbf{87.87} \\
\bottomrule
\end{tabular*}
\tabnote{Significance levels $^{*}p<0.1$, $^{**}p<0.05$, $^{***}p<0.01$ are from paired $t$-tests across five runs: routing baselines are compared with WILC; WILC ($r_{\max}{=}1$) is compared with full WILC. The best result in each column per scale is shown in \textbf{bold} and the second best is \underline{underlined}. WILC ($r_{\max}{=}1$) uses only the burn-in round (all workers generate responses, the best is selected by proxy evaluation) without subsequent multi-round refinement.}
\end{threeparttable}
\end{table}

\subsection{Analysis}

The results reveal that routing effectiveness is bounded by training data coverage. On HumanEval and MATH-500, which correspond to coding and math domains well-represented in the LiveBench cold-start data, the strongest routing method (Elo Router at 14B; LLM Routing at 30B) achieves performance comparable to WILC on individual benchmarks. However, on VisEval---a task requiring natural-language-to-visualization code generation (NL2VIS), a task type not directly represented in the LiveBench cold-start categories---all routing methods collapse. At the 14B scale, the best routing method achieves only 56.73\% on VisEval versus 76.72\% for WILC; at the 30B scale, the gap remains substantial (63.20\% vs.\ 84.06\%). Moreover, at the 30B scale all three routing methods score below every single-model baseline on VisEval (Table~\ref{tab:main_results_30b}), indicating that routing on uncovered task types may be not merely uninformative but actively harmful: the learned routing functions systematically favor workers that are suboptimal for the unseen domain.

This contrast highlights a fundamental difference in robustness. WILC's single-round variant ($r_{\max}{=}1$) already outperforms all routing baselines on AVG by a wide margin (86.32\% vs.\ $\leq$80.78\% at 14B; 86.84\% vs.\ $\leq$80.22\% at 30B), despite using the same worker pool. The reason is that its evaluate-then-select mechanism (described above) directly observes worker performance on the target query, making it agnostic to training data coverage. The full WILC framework further improves upon this single-round selection through iterative multi-round refinement, where successive relay steps among complementary workers yield additional gains, particularly on challenging tasks like VisEval (+4.66 pp at 14B; +3.36 pp at 30B over $r_{\max}{=}1$).

Moreover, the persistent gap between WILC and its single-round variant on VisEval demonstrates that multi-round iterative collaboration remains effective even on task types not covered by the cold-start data. This is because WILC's successor selection is guided by \textit{model-bottleneck fitness}---which worker best resolves a specific bottleneck type (e.g., logical errors, code bugs, format violations)---rather than model-query fitness that maps surface-level domain features to model preferences (as discussed in Section~\ref{subsubsec:value-of-learning-model-bottleneck-fitness}). Since bottleneck patterns are largely scenario-agnostic (a debugging deficiency manifests similarly whether the task is MATH or NL2VIS), the relay-style complementarity mechanism continues to identify suitable successors even for task types absent from the cold-start categories. In practical deployment, where target queries rarely conform perfectly to training distributions, this property provides a robustness advantage that routing-based approaches inherently lack.

\section{Detailed Cost-Effectiveness Comparison}
\label{apx:cost-analysis}

This section provides a detailed cost-effectiveness comparison between WILC and frontier closed-source models. The analysis requires two preparatory steps: (1) evaluating frontier model performance on the same benchmarks, and (2) measuring the actual number of collaboration rounds consumed by WILC.

\paragraph{Frontier model performance.} We evaluate GPT-4o, GPT-5.2, and GPT-5.4 on the same four benchmarks used in our main experiments---HumanEval, MATH-500, MMLU, and VisEval. Table~\ref{tab:gpt_benchmark} reports the per-benchmark results.

\begin{table}[!ht]
\centering
\caption{Benchmark Performance of Frontier Closed-Source Models}
\label{tab:gpt_benchmark}
\small
\begin{threeparttable}
\begin{tabular*}{0.8\textwidth}{@{\extracolsep{\fill}}lcccc}
\toprule
\textbf{Model} & \textbf{HumanEval} & \textbf{MATH-500} & \textbf{MMLU} & \textbf{VisEval} \\
\midrule
GPT-4o        & 87.80 & 75.20 & 83.96 & 82.00 \\
GPT-5.2       & 93.90 & 86.40 & 89.40 & 82.67 \\
GPT-5.4       & 93.29 & 89.20 & 90.25 & 86.67 \\
\bottomrule
\end{tabular*}
\tabnote{All models are evaluated via the official OpenAI API under a zero-shot setting.}
\end{threeparttable}
\end{table}

\paragraph{Average actual rounds.} In WILC framework, the PCG gate terminates the process as soon as complementarity is exhausted, often after only the burn-in round. Table~\ref{tab:avg_rounds} reports the average actual rounds across benchmarks and model scales. The majority of queries terminate after round~1, yielding overall averages of 1.78 (14B) and 1.46 (30B), well below the theoretical maximum of $r_{\max}=6$.

\begin{table}[!ht]
\centering
\caption{Average Actual Rounds by Benchmark and Scale}
\label{tab:avg_rounds}
\begin{threeparttable}
\begin{tabular*}{0.85\textwidth}{@{\extracolsep{\fill}}l*{4}{>{\centering\arraybackslash}p{1.5cm}}}
\toprule
& \multicolumn{2}{c}{\textbf{Avg. Rounds}} & \multicolumn{2}{c}{\textbf{Round-1 Rate (\%)}} \\
\cmidrule(lr){2-3} \cmidrule(lr){4-5}
\textbf{Benchmark} & \textbf{14B} & \textbf{30B} & \textbf{14B} & \textbf{30B} \\
\midrule
HumanEval & 2.18 & 1.64 & 56.1 & 85.8 \\
MATH-500  & 1.46 & 1.05 & 90.8 & 96.2 \\
MMLU      & 1.05 & 1.01 & 97.7 & 99.6 \\
VisEval   & 2.41 & 2.13 & 71.3 & 77.3 \\
\midrule
\textbf{Unweighted Avg.} & \textbf{1.78} & \textbf{1.46} & \textbf{79.0} & \textbf{89.7} \\
\bottomrule
\end{tabular*}
\tabnote{``Avg.\ Rounds'' is the mean number of actual collaboration rounds per query, where the actual round count is the round at which the PCG gate terminates the process. ``Round-1 Rate'' is the percentage of queries resolved after only the burn-in round ($r=1$). Unweighted averages are computed across the four benchmarks.}
\end{threeparttable}
\end{table}

\paragraph{Cost comparison.} Combining the frontier model performance (Table~\ref{tab:gpt_benchmark}) with the empirical round counts (Table~\ref{tab:avg_rounds}) and public API pricing from OpenRouter\footnote{We use OpenRouter (\url{https://openrouter.ai}) as a unified pricing source because it hosts both the open-source models used by WILC (e.g., Qwen2.5, DeepSeek-R1, Phi-4) and the frontier closed-source models (GPT-4o, GPT-5.2, GPT-5.4) under a single marketplace, enabling an apples-to-apples cost comparison. Official provider pricing (e.g., OpenAI's API) covers only their own model families and may differ from aggregator pricing due to margin structures; using a single platform avoids mixing pricing regimes. All prices were retrieved in May 2026.} (May 2026), Table~\ref{tab:cost_comparison} presents a comprehensive cost-effectiveness comparison.\footnote{We adopt public API pricing as a standardized yardstick because the cost of self-hosted deployment is difficult to quantify in a generalizable manner---it depends on hardware configuration, GPU utilization rate, energy cost, and maintenance overhead, all of which vary across organizations. API pricing provides a reproducible, hardware-agnostic baseline for cross-method comparison. We also omit latency comparisons for similar reasons: end-to-end response time under self-hosted deployment is largely determined by local infrastructure (GPU count, memory bandwidth, batching strategy, network topology) rather than the method itself, making it uninformative as a method-level metric and incomparable with cloud-based API latency.}

\begin{table}[!ht]
\centering
\caption{Cost-Effectiveness Comparison across Four Benchmarks}
\label{tab:cost_comparison}
\small
\begin{threeparttable}
\begin{tabular*}{0.95\textwidth}{l*{6}{>{\centering\arraybackslash}p{2.1cm}}}
\toprule
\textbf{Method} & \textbf{Scale} & \textbf{Avg. (\%)} & \multicolumn{2}{c}{\textbf{Price (/1M tokens)}} & \textbf{Calls} & \textbf{Cost/Query} \\
\cmidrule(lr){4-5}
 & & & \textbf{Input} & \textbf{Output} & \textbf{/Query} & \textbf{(USD)} \\
\midrule
GPT-4o      & ---  & 82.24 & \$2.50  & \$10.00  & 1    & \$0.0075 \\
GPT-5.2     & ---  & 88.09 & \$1.75  & \$14.00  & 1    & \$0.0088 \\
GPT-5.4     & ---  & 89.85 & \$2.50  & \$15.00  & 1    & \$0.0100 \\
\midrule
WILC        & 14B  & 87.78 & \$0.06  & \$0.20   & 8.3  & \$0.0013 \\
WILC        & 30B  & 87.87 & \$0.09  & \$0.18   & 7.4  & \$0.0013 \\
\bottomrule
\end{tabular*}
\tabnote{Average performance is computed over HumanEval, MATH-500, MMLU, and VisEval (see Table~\ref{tab:gpt_benchmark} for per-benchmark GPT results; WILC results are from Tables~\ref{tab:main_results_14b} and~\ref{tab:main_results_30b}). Pricing is from OpenRouter (May 2026). Cost per query assumes 1{,}000 input tokens and 500 output tokens per call. Open-source model prices are averaged across workers in each scale pool. WILC's calls per query are derived from the API call structure in Table~\ref{tab:api_wilc}: 6 calls for the burn-in round plus 3 calls for each additional round, yielding 8.3 calls for 14B (avg.\ 1.78 rounds) and 7.4 calls for 30B (avg.\ 1.46 rounds), based on the empirical round counts in Table~\ref{tab:avg_rounds}. Self-hosted deployment, as used in our experiments, incurs fixed hardware cost rather than per-token charges, potentially lowering the per-query cost even further at scale.}
\end{threeparttable}
\end{table}

Several observations emerge from the comparison. First, WILC at the 14B scale achieves an average accuracy comparable to GPT-5.2 (87.78\% vs.\ 88.09\%) at roughly 15\% of the per-query cost---a 7$\times$ cost reduction---and substantially outperforms GPT-4o (82.24\%) at an even larger cost advantage. Second, WILC at the 30B scale achieves an average accuracy of 87.86\%, nearly matching GPT-5.2 while costing only 15\% as much per query (\$0.0013 vs.\ \$0.0088). Even compared with GPT-5.4, the strongest frontier model considered, WILC-30B trails by only 2.0 percentage points in average accuracy at approximately 13\% of the cost (8$\times$ cheaper).

These estimates use API pricing as a common yardstick, but the actual economics of WILC are even more favorable. In our experiments, all open-source models are served locally with \texttt{llama.cpp} using 4-bit quantization (Q4\_K\_M). Under self-hosted deployment, inference cost is governed by fixed hardware expenditure rather than per-token charges. While the amortized per-query cost depends on query volume---potentially higher than API pricing at very small scale---it decreases rapidly with throughput, as the per-token API charge is eliminated entirely and total cost of ownership is spread over a growing number of queries.

Beyond cost, self-hosted deployment offers a qualitative advantage that closed-source APIs cannot match: enhanced data control. All data remain within the organization's infrastructure, supporting the privacy and compliance requirements that motivate private LLM deployments in sectors such as finance, healthcare, and government~\citep{aimultiple_enterprise_genai_2024}. WILC thus enables organizations to achieve frontier-level performance through the collaborative intelligence of affordable, privately deployed models---without sacrificing data control or incurring prohibitive API costs.

\end{document}